\documentclass{article}
\usepackage{leaplab}
\usepackage[table,x11names]{xcolor}

\usepackage[utf8]{inputenc} 
\usepackage[T1]{fontenc}    
\usepackage{url}            
\usepackage{booktabs}       
\usepackage{amsfonts}       
\usepackage{nicefrac}       
\usepackage{microtype}      
\usepackage{tabularx}   
\usepackage{enumitem}

\usepackage[utf8]{inputenc} 
\usepackage[T1]{fontenc}    
\usepackage{url}            
\usepackage{booktabs}       
\usepackage{amsfonts}       
\usepackage{nicefrac}       
\usepackage{microtype}      
\definecolor{citecolor}{HTML}{0071bc}
\definecolor{shadecolor}{HTML}{EFEFEF}
\definecolor{linkcolor}{HTML}{9A4D92}
\usepackage[colorlinks, linkcolor=linkcolor, colorlinks, anchorcolor=blue, citecolor=citecolor, pagebackref=True]{hyperref}

\usepackage{amsmath}
\usepackage{microtype}
\usepackage{graphicx}
\usepackage{booktabs} 
\usepackage{caption}
\usepackage{wrapfig}
\usepackage{float}

\usepackage{titletoc} 

\usepackage{subcaption}
\usepackage[export]{adjustbox}
\usepackage{footnote}
\usepackage{tablefootnote}
\usepackage{threeparttable}
\usepackage[capitalize,noabbrev]{cleveref}

\usepackage{tabularray}
\UseTblrLibrary{booktabs}
\usepackage{multirow}
\usepackage{multicol}
\usepackage{xspace}
\usepackage{mathrsfs}
\usepackage{pifont}
\usepackage{makecell}
\usepackage{CJK}

\usepackage{marvosym} 
\usepackage[most]{tcolorbox}

\definecolor{brand}{HTML}{000000}
\definecolor{xulabwine}{HTML}{5E3545}
\definecolor{titlebg}{HTML}{F7F2F3}

\hypersetup{citecolor=citecolor,linkcolor=linkcolor,filecolor=linkcolor,urlcolor=magenta}

\usepackage{titletoc}

\newenvironment{leapabstract}{\begin{tcolorbox}[
colback=titlebg,colframe=white,boxrule=0pt,arc=10pt,
left=16pt,right=16pt,top=12pt,bottom=12pt,width=\textwidth,
enlarge left by=0mm,before skip=10pt,after skip=10pt]}{\end{tcolorbox}}

\fancypagestyle{fancytitlepage}{\fancyhead{}\lhead{\color{brand}\Large\bfseries Xu Lab}\rhead{\color{brand}\itshape HKUST}\cfoot{}}

\usepackage{amssymb}
\usepackage{bm}
\usepackage{amsthm}

\newcommand{\SO}{\mathrm{SO}}

\DeclareMathOperator*{\argmin}{arg\,min}
\DeclareMathOperator{\rotsix}{rot6d}
\newtheorem{proposition}{Proposition}

\usepackage{flafter}

\icmltitlerunning{PosePlaner: Denoising Any Feed-Forward Pose Predictor}
\hypersetup{pdfauthor={Yihang Chen, Bin Tan, Yujun Shen, Yinghao Xu, Nan Xue}}

\begin{document}
\raggedbottom

\icmltitle{PosePlaner: Denoising Any Feed-Forward Pose Predictor from Pairwise Planar Geometry}

\noindent\begin{minipage}{\textwidth}
\begin{icmlauthorlist}
\textbf{Yihang Chen}$^{1,2}$ \enspace \textbf{Bin Tan}$^{2,\ddagger}$ \enspace \textbf{Yujun Shen}$^{2}$ \enspace \textbf{Yinghao Xu}$^{1,\ddagger}$ \enspace \textbf{Nan Xue}$^{2}$
\end{icmlauthorlist}
\par\vspace{4pt}
$^{1}$HKUST, Hong Kong, China \quad $^{2}$Robbyant, Ant Group, China
\par\vspace{2pt}
$^{\ddagger}$Corresponding Authors
\end{minipage}
\par\vspace{4pt}

\begin{figure}[H]
\centering
\includegraphics[width=0.85\linewidth, trim={0 120 0 120}, clip]{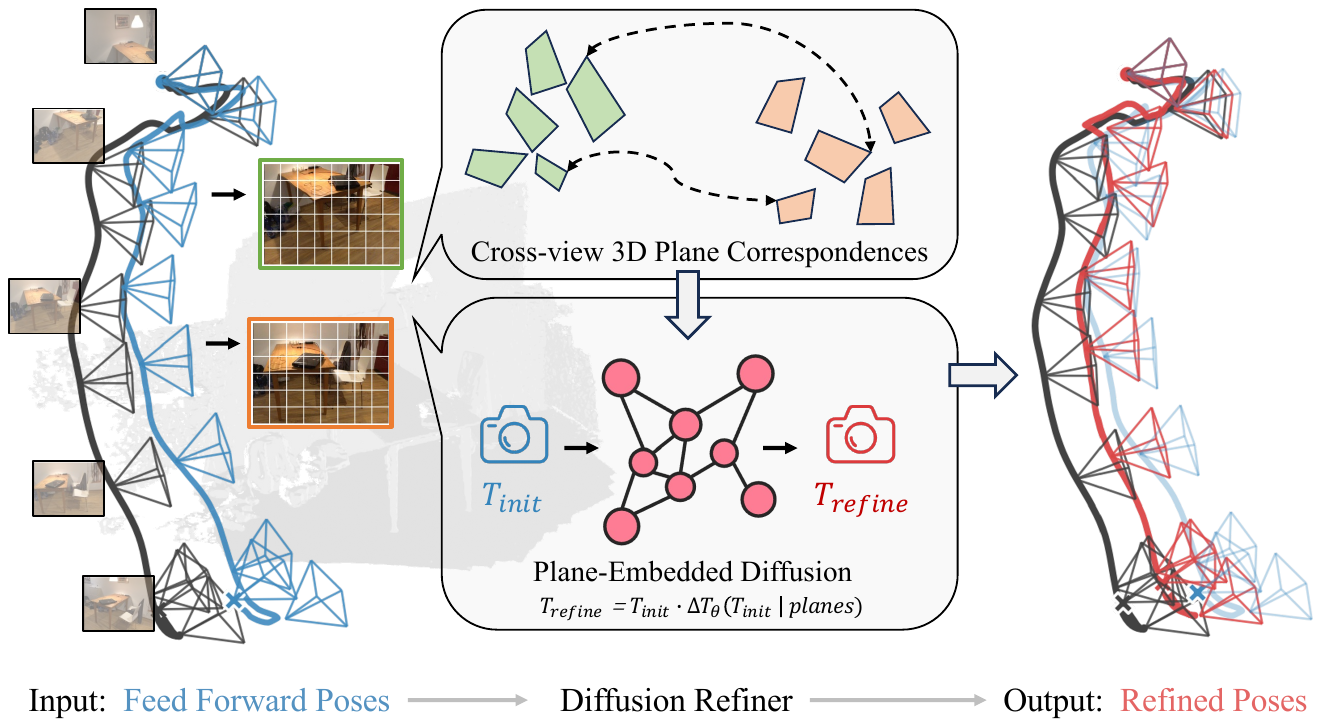}
\caption{Our proposed \textsc{PosePlaner} refines any feed-forward pose in one Euler step of a plane-embedded diffusion adapter, and extends to $n$-view streaming via per-edge view-graph refinement aggregated by rotation averaging and a \textsc{Glomap}-style positioner. \textbf{Left:} $T_{\text{init}}$ (\textcolor{blue}{\textbf{blue}}) drifts from GT (\textbf{black}). \textbf{Middle:} conditioned on per-view planar primitives, the refiner yields $T_{\text{refine}} = T_{\text{init}}\cdot\Delta T_\theta(T_{\text{init}}\mid\text{planes})$ (\textcolor{red}{\textbf{red}}) in one step. \textbf{Right:} refined trajectory (\textcolor{red}{\textbf{red}}), raw feed-forward (\textcolor{blue}{\textbf{blue}}), GT (\textbf{black}).}
\label{fig:teaser}
\end{figure}

\begin{leapabstract}
Recent multi-view Transformers (e.g., DUSt3R and VGGT) have advanced 3D reconstruction by scaling training data and model capacity from two views to many views and long sequences, subsuming the classical SfM/SLAM pipeline of feature extraction, matching, and pose estimation in a single feed-forward pass.
Yet this subsumption hides the interfaces between pipeline stages---correspondences, geometry, and pose are entangled in a single forward pass, \textbf{leaving supervised fine-tuning as the major path} to improvement and offering no mechanism to correct predictions from geometric evidence at test time.
We propose \textsc{PosePlaner} to close this gap by reviving the classical coupling between correspondences and pose via flow matching, treating pose refinement as a conditional generation problem that takes coarse geometric matches and initial pose as input condition and fuses the resulting pose hypotheses into a single refined estimate, providing a continuous, data-driven analog of RANSAC voting.
Our method is trained efficiently on two-view data, yet serves as a periodic refiner for streaming $n$-view pose estimators with no retraining.
Empirically, it corrects failures of feed-forward predictors, substantially improves correspondence-based solvers, and reduces accumulated drift in long streaming sequences.
Project page: \url{https://robbyant-research.github.io/PosePlaner/}.

\end{leapabstract}

\section{Introduction}
\label{sec:intro}
Recovering relative camera pose between views of a scene is a foundational primitive in 3-D vision and the entry point of structure-from-motion~\citep{schonberger2016structure}, multi-view stereo~\citep{yao2018mvsnet}, SLAM~\citep{mur2015orb}, and novel-view synthesis~\citep{nerf,3dgs}. The problem is hard at both ends: in the two-view setting under limited baselines and outlier-prone correspondences, and across long streaming sequences where errors accumulate frame by frame. Every pose failure propagates into the downstream pipeline.

Two prominent paradigms in recent work reach this problem from opposite directions. \emph{Feed-forward pose predictors}~\citep{vggt,dust3r} map an image pair directly to $(R,\bm{t})$ via visual priors, but their prediction is terminal: downstream correspondences cannot refine it. 
\emph{Matcher-plus-solver pipelines}~\citep{superglue,loftr,roma,mast3r} hand rich correspondences to a five-point solver~\citep{nister5pt} wrapped in RANSAC~\citep{ransac}, but the matcher's priors collapse into per-pair inlier / outlier decisions and the RANSAC threshold tuning seldom transfers across scenes. Both paradigms break the match-pose coupling: visual priors and per-match geometric evidence never reach the same differentiable channel.

In this paper, we bridge the gap between these two paradigms by reintroducing visual correspondences into feed-forward 3D reconstruction.
Analogous to the algebraic solvers that seed the classical SfM pipeline~\citep{schonberger2016structure}, we treat feed-forward models as pose initializers and refine their output as a learned geometric correction step.
Our key insight is that 3D surface points paired with their normals already encode a cross-view geometric constraint: under the true relative pose, matched patch-normal pairs must satisfy an algebraic warping identity, and departures from it yield differentiable residuals that grow with pose error. Planar primitives realize this constraint compactly---one $3$-D center and one normal per image patch. We learn to denoise the pose residual via flow matching, conditioned on the initialized pose and matched patch-normal pairs, yielding a refiner that generalizes from two-view training to $N$-view inference without retraining.
%


%


We introduce \textsc{PosePlaner} (Fig.~\ref{fig:teaser}), a pose refiner that accepts any initial pose $T_{\text{init}}$ and returns a refined relative pose given patch-normal matching pairs. Building on \textsc{Plana3r}~\citep{plana3r}, we use planar primitives as compact patch-normal representations, where each patch predicts a 3D center and plane normal in its own camera frame; pairing these patch-level primitives via an external pixel matcher (e.g., \textsc{MASt3R}~\citep{mast3r}, \textsc{LoFTR}~\citep{loftr}) yields $M$ two-view correspondences whose warping residuals serve as the conditioning signal for the refiner. For pose refinement in SE(3), we exploit the asymmetric difficulty of rotation and translation: a flow-matching~\citep{liu2022rectified,flowmatching} multi-query Transformer~\citep{transformer,dit} denoises the rotation residual conditioned on the plane-warping residuals, producing $M$ hypotheses that a learned score head fuses with the initial pose by softmax-weighted voting; translation is then recovered by a geometric solver under the refined rotation.
Trained once on two-view pairs, \textsc{PosePlaner} extends as a periodic tail refiner for streaming $n$-view pose estimators on long sequences with no retraining.
Beyond classical bundle adjustment and pose-graph backends~\citep{schonberger2016structure}, to the best of our knowledge this is the first \emph{learned, diffusion-based} refiner that operates on the pose outputs of streaming $n$-view estimators without sequence-specific retraining.
Experimentally, our \textsc{PosePlaner} consistently improves upstream estimators from both paradigms on two benchmarks, and its gains grow with sequence length in the streaming setting, where drift correction matters most---e.g., on ScanNet at $N{=}2000$, rotation AUC@$5^\circ$ improves by $+20$ points for \textsc{CUT3R}~\citep{cut3r} and $+21$ for \textsc{TTT3R}~\citep{ttt3r}.
\vspace{-4pt}
\paragraph{Our contributions.}
\begin{itemize}[leftmargin=1.2em, itemsep=1pt, topsep=2pt, parsep=0pt]
\item We propose \textsc{PosePlaner}, the first \emph{learned, diffusion-based} pose refiner that revives the classical correspondence--pose coupling for feed-forward 3D reconstruction models, treating pose refinement as a conditional generation problem over plane-warping residuals.
\item We design a hybrid SE(3) refinement scheme: a flow-matching multi-query Transformer for rotation with score-based hypothesis fusion, and a geometric solver for translation, enabling single-step inference.
\item We show that a model trained once on two-view pairs serves as a periodic tail refiner for streaming $n$-view estimators with no retraining. 
\end{itemize}
\section{Related Work}
\label{sec:related}
\paragraph{Two-view and feed-forward pose estimation.}
Classical structure-from-motion~\citep{schonberger2016structure,mur2015orb} recovers relative pose from sparse correspondences via a closed-form essential-matrix solver~\citep{nister5pt} wrapped in RANSAC~\citep{ransac}; modern matchers~\citep{superglue,loftr,roma,mast3r} replace hand-crafted descriptors with learned dense features, but the discrete solver still collapses each match into an inlier / outlier vote, and a single bad RANSAC threshold can erase the entire inlier set. Differentiable RANSAC variants~\citep{dsac,dsacpp,dsacstar} relax this voting into a soft, end-to-end-trainable selection, yet gradients still flow through aggregate inlier counts rather than per-correspondence geometric residuals. A complementary line skips the solver and regresses pose directly via large feed-forward $3$-D backbones~\citep{dust3r,mast3r,vggt} or their online streaming descendants~\citep{cut3r,ttt3r,wint3r,infinitevggt,streamvggt,stream3r,spann3r,gca}; both produce a terminal prediction whose downstream correspondences and accumulated drift never re-enter the prediction loop. \textsc{PosePlaner} sits downstream of either paradigm, correcting any initial pose using the geometric constraint encoded in matched plane primitives and restoring the pose--correspondence coupling that both break.

\paragraph{Planar primitives for $3$-D understanding.}
Indoor scenes are dominated by piecewise-planar structure, motivating reconstruction by sparse $3$-D plane primitives rather than dense points or meshes~\citep{yu2019single,sparseplane,planarrecon,planarsplatting,zhao2024monoplane,liu2025towards,plana3r}. Early feed-forward methods~\citep{planenet,planercnn,planerectr} regress plane masks and per-pixel plane parameters via instance segmentation, but depend on manually annotated plane labels (\textsc{ScanNetV2}~\citep{scannet}, \textsc{Matterport3D}~\citep{chang2017matterport3dlearningrgbddata}), capping both training scale and cross-dataset generalization. \textsc{PlanarSplatting}~\citep{planarsplatting} breaks this bottleneck with a differentiable plane-primitive rasterizer whose rendering loss is annotation-free, and \textsc{Plana3r}~\citep{plana3r} uses it to train a feed-forward stereo plane-primitive backbone at scale, predicting per-patch normals and centers end-to-end. On the pose side, \textsc{NopeSAC}~\citep{nopesac} matches a few scene-level plane instances, feeds each pair through a shared MLP to produce one pose hypothesis per correspondence, and soft-fuses these (with the init) via a learned scoring layer over their warping costs. \textsc{PosePlaner} retains the soft-fusion idea but operates at the much denser \emph{patch level} and replaces the per-plane MLP regressor with a \emph{continuous flow-matching denoiser} jointly conditioned on the plane-warping residuals of all matched patch pairs.

\paragraph{Diffusion models for pose and $3$-D.}
Denoising diffusion~\citep{ddpm} has been adapted to pose by \textsc{PoseDiffusion}~\citep{posediffusion} (camera trajectories conditioned on image features), \textsc{RayDiffusion}~\citep{raydiffusion} (denoising a per-pixel ray bundle), and \textsc{DiffusionSfM}~\citep{diffusionsfm} (jointly denoising per-pixel ray origins and endpoints to recover structure and motion); all three are visual-prior-driven, with no explicit per-correspondence geometric channel. Flow matching~\citep{flowmatching,liu2022rectified} replaces the noise-prediction objective with a deterministic vector-field target, enabling single-step inference. \textsc{PosePlaner} adopts this flow-matching parameterization and conditions on \emph{plane-warping residuals} rather than pixels, so the denoiser sees the algebraic quantity a classical solver would minimize---a geometric channel decoupled from any single matcher's visual statistics.

\section{Method}
\label{sec:method}
\textsc{PosePlaner} takes any initial two-view relative pose $T_{\text{init}}$ of the input image pair $(\mathcal{I}_1, \mathcal{I}_2)$ and refines it into a more accurate estimate. The main idea rests on a classical observation:

\begin{proposition}[Plane-Warping Identity]
\label{prop:warp}
Let $(\bm{n}_1, d_1)$ and $(\bm{n}_2, d_2)$ be the parameters of a 3D plane observed in two camera frames related by a relative pose $T_{\text{rel}} = (R, \bm{t})$. Then
\begin{equation}
\label{eq:warp-prop}
\bm{n}_1 = R\,\bm{n}_2, \qquad d_1 = d_2 - (R\,\bm{n}_2)^{\!\top}\bm{t}.
\end{equation}
Departures from this identity produce the \emph{plane-warping residuals}
\begin{equation}
\label{eq:residual-prop}
\Delta\bm{n}(R) = \left\|\bm{n}_1 - R\,\bm{n}_2\right\|_2, \qquad \Delta d(R,\bm{t}) = |d_1 - d_2 + (R\,\bm{n}_2)^{\!\top}\bm{t}|,
\end{equation}
which vanish at the true pose and grow with pose error.
\end{proposition}

\begin{figure}[t]
\centering
\includegraphics[width=\linewidth, trim=0 125 0 125, clip]{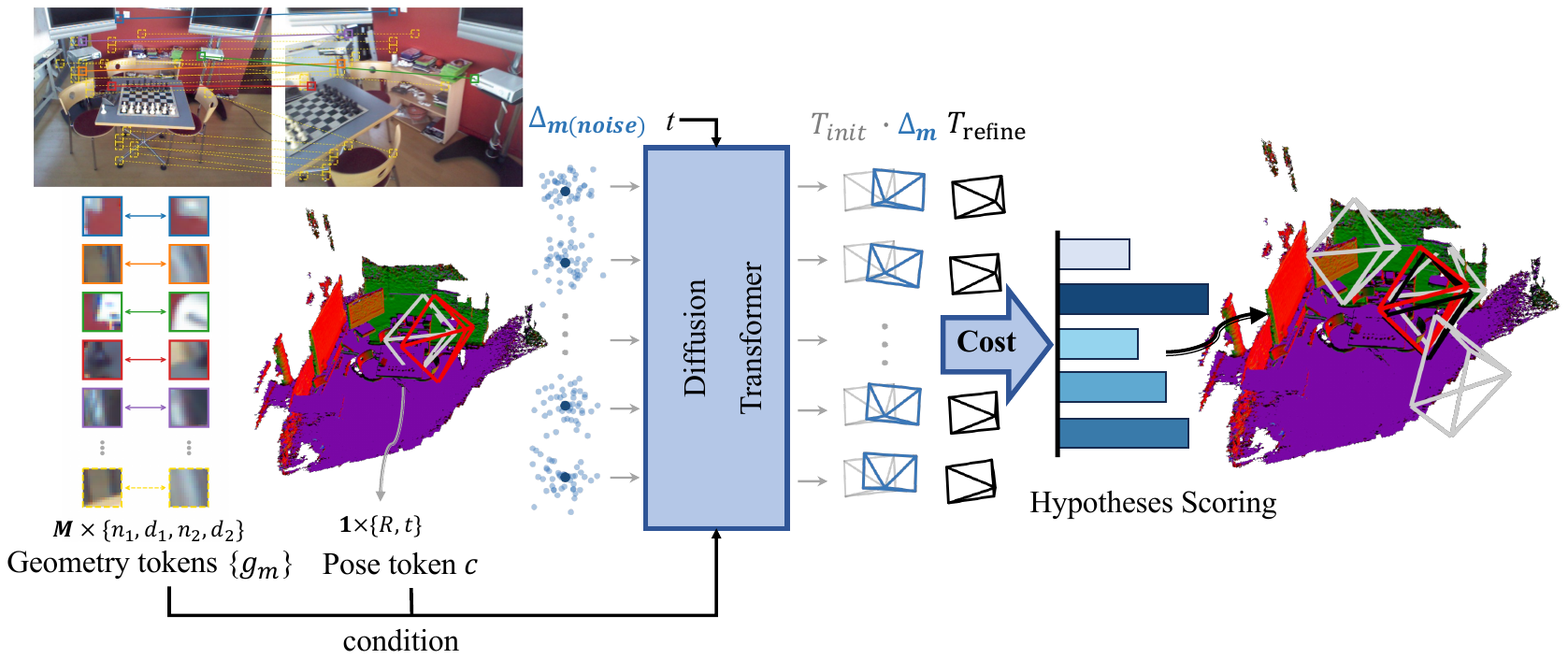} 
\caption{\textbf{\textsc{PosePlaner} pipeline.} A per-view local-frame plane head produces $M$ geometry tokens $\{\bm{g}_{m}\}$ (paired plane primitives across the two views) and a global pose token $\bm{c}$ that encodes the initial relative pose $T_{\text{init}}$. The multi-query diffusion Transformer denoises $T_{\text{init}}$ in a single Euler step under these conditioning tokens, yielding $M$ hypothesis frustums $h_{1},\dots,h_{M}$. A score head softly fuses the hypotheses into the final refined pose $T_{\text{refined}}$, used to align the two-view reconstruction.}
\label{fig:pipeline}
\end{figure}

These residuals provide an estimator-agnostic conditioning signal for any upstream source.
Given a two-view Vision Transformer that reads out patch-level 3D planes in each view's own camera frame (Sec.~\ref{sec:planehead}), we compute the plane-warping residuals between matched pairs and use them as conditioning for a flow-matching Transformer that denoises the rotation residual; translation is then recovered by a geometric solver under the refined rotation (Sec.~\ref{sec:refiner}). See Fig.~\ref{fig:pipeline} for an overview.

\subsection{Patchwise Plane Primitives and Correspondences}
\label{sec:planehead}

Our method follows the convention of Vision Transformers~\citep{vit}, patchifying the input images $(\mathcal{I}_1,\mathcal{I}_2)$ into $S$ patch tokens each. A siamese ViT encoder processes both views independently, and two cross-attention decoders fuse information across views, predicting a 3D planar primitive per patch, specifically a center depth, two radii, and an orientation quaternion from which the plane normal $\bm{n}$ and offset $d$ are recovered.

\paragraph{Plane representation from Siamese Vision Transformers.}
Each patch in a view predicts a local planar primitive parameterized by a 3D center $\bm{P} \in \mathbb{R}^3$ and a unit surface normal $\bm{n} \in S^2$, both expressed in its own camera frame. The plane offset follows as $d = -\bm{n}^{\!\top}\bm{P}$, giving a compact $(\bm{n}, d)$ per patch. We adopt the architecture of \textsc{Plana3r}~\citep{plana3r} with one key change: whereas \textsc{Plana3r} expresses all primitives in a shared global frame (view~$1$), we set each primitive to live in its own source camera's frame. 
The training details of this Siamese Vision Transformer network are in App.~\ref{app:plane-training}.

\paragraph{Two-view plane correspondences.}
After the plane head produces $(\bm{n}, d)$ for every patch in both views, we establish patch-level correspondences from any external pixel matcher (e.g., SuperGlue~\citep{superglue}, LoFTR~\citep{loftr}, ROMA~\citep{roma}, or MASt3R~\citep{mast3r}). Given pixel-level matches, we convert each matched pixel pair to its enclosing patch indices and accumulate a vote matrix $V \in \mathbb{R}^{S \times S}$ counting the number of pixel matches per patch pair. Patch-level correspondences are then extracted by mutual nearest neighbors: patch $i$ in view~$1$ is matched to patch $j$ in view~$2$ if and only if $j = \arg\max_k V_{ik}$ and $i = \arg\max_k V_{kj}$. This yields $M$ matched plane pairs $\{(\bm{n}^1_m, d^1_m),\, (\bm{n}^2_m, d^2_m)\}_{m=1}^{M}$, which we pack into geometry tokens
\begin{equation}
\label{eq:geotoken}
\bm{g}_{m} \;=\; \big[\,\bm{n}^{1}_{m}\;\big\|\;\bar d^{1}_{m}\;\big\|\;\bm{n}^{2}_{m}\;\big\|\;\bar d^{2}_{m}\,\big]\in\mathbb{R}^{8},\qquad m=1,\dots,M,
\end{equation}
where $\bar d^{v}_{m}$ denotes the scene-normalized offset (Sec.~\ref{sec:refiner}). Each token is the minimal sufficient statistic for evaluating the plane-warping residuals of Eq.~\eqref{eq:residual-prop} at any candidate pose. Note that $M$ is not a fixed hyperparameter: it is determined per image pair by the matcher's mutual-NN output and varies with scene texture and matcher density (typically tens to a few hundred per pair for the matchers we use). The $M$ residuals collectively expose the pose error along diverse normal orientations, forming the conditioning signal consumed by the flow-matching refiner below.

\subsection{Flow-Matching Diffusion Refiner}
\label{sec:refiner}

The refiner takes $T_{\text{init}}=(R_{\text{init}},\bm{t}_{\text{init}})$ and the geometry tokens $\{\bm{g}_m\}$ of Sec.~\ref{sec:planehead}, and produces a refined pose via flow matching~\citep{flowmatching,liu2022rectified}. For $t\sim\mathcal{U}[0,1]$ and $\bm{\epsilon}\sim\mathcal{N}(0,I)$, the formulation defines a linear probability path and a constant target velocity
\begin{equation}
\bm{z}_{t} \;=\; (1-t)\,\bm{z}_{0}+t\,\bm{\epsilon},
\qquad
\bm{v}^{\star}(\bm{z}_{t},t) \;=\; \bm{\epsilon}-\bm{z}_{0},
\end{equation}
and the refiner $F_{\theta}$ regresses $\bm{v}^{\star}$ at every $(\bm{z}_t, t)$. The straight-line path admits a \emph{single} deterministic Euler step at inference, so refinement reduces to one forward pass (see App.~\ref{app:denoising} for the formal argument).



\paragraph{Velocity field parameterization.} The velocity field $F_\theta$ is specified by three choices: the coordinate system, the denoising target, and the conditioning signal.

We normalize all plane offsets by a per-pair scalar $\sigma=\tfrac{1}{2}(\overline{|d^{1}|}+\overline{|d^{2}|})$ so that $\bar d^{v}_{m}=d^{v}_{m}/\sigma$. A robust median over the matched pairs recovers the normalized translation scale:
\begin{equation}
\label{eq:sest}
s_{\text{est}}(R,\hat{\bm{t}}) \;=\; \mathrm{median}_{m\in\mathcal{M}^{\star}}\bigg|\frac{\bar d^{2}_{m}-\bar d^{1}_{m}}{(R\bm{n}^{2}_{m})^{\!\top}\hat{\bm{t}}}\bigg|,
\qquad
\bm{\tau} \;=\; s_{\text{est}}(R,\hat{\bm{t}})\,\hat{\bm{t}},
\end{equation}
where $\mathcal{M}^{\star}$ excludes matches with near-zero denominators. This maps any $T_{\text{init}}$ into the range of $\bar d$ regardless of its metric scale; the metric translation is recovered by $\bm{t}=\sigma\,\bm{\tau}$.

The denoising target is the $9$-dimensional residual
\begin{equation}
\label{eq:z0}
\bm{z}_{0} \;=\; \big[\,\rotsix(R_{\text{gt}}R_{\text{init}}^{\!\top})\;\big\|\;\bm{\tau}_{\text{gt}}-\bm{\tau}_{\text{init}}\,\big]\in\mathbb{R}^{9},
\end{equation}
with rotation in the continuous $6$-D representation of~\citet{rot6d} (see App.~\ref{app:geo-rot6d} for why quaternions are suboptimal under flow matching). Although our final output uses only the rotation, the translation channel acts as a contextual signal that tightens the rotation prediction (Sec.~\ref{sec:exp-ablation}).

The refiner conditions on the $M$ geometry tokens $\{\bm{g}_{m}\}$ from Eq.~\eqref{eq:geotoken} and a global pose token $\bm{c}=\psi([\rotsix(R_{\text{init}})\|\bm{\tau}_{\text{init}}])\in\mathbb{R}^{d}$. Linear projections lift both to width $d$ and $L_{\text{cond}}{=}4$ self-attention blocks produce the context $\hat{\bm{C}}$ consumed by the denoiser.

\paragraph{Multi-query denoiser.} The denoiser instantiates $M$ queries, one per correspondence, each solving the same pose-residual denoising problem under a different local geometric condition. Every query shares a latent-time component
\begin{equation}
\label{eq:qshared}
\bm{q}_{0} \;=\; \phi_{z}(\bm{z}_{t}) + \phi_{t}\!\big(\mathrm{SinEmb}(t)\big)\in\mathbb{R}^{d},
\end{equation}
and adds a per-match geometry bias $\bm{q}_{m}=\bm{q}_{0}+W_{b}\,\hat{\bm{g}}_{m}$ that grounds each query in correspondence~$m$. The $M$ queries cross-attend to $\hat{\bm{C}}$ and emit per-match velocity predictions $\bm{v}_{m}\in\mathbb{R}^{9}$. During training, these per-match predictions are aggregated into a single fused velocity by geometric weights
\begin{equation}
\label{eq:omega}
\omega_{m} \;=\; \mathrm{softmax}(-\beta\,g_{m}),\qquad g_{m} \;=\; \|\Delta\bm{n}_{m}\|_{2} + \lambda\,|\Delta d_{m}|,\qquad (\lambda,\beta)=(3,10),
\end{equation}
so the $M$ queries cooperatively regress the shared FM target $\bm{v}^{\star}$ via their weighted average; correspondences with small warping residuals dominate the gradient. Per-query diversity at inference arises from the per-match geometry bias $W_b\hat{\bm{g}}_m$, which biases each $\bm{v}_m$ toward a different local geometric condition. At inference, each query is integrated independently into one of $M$ hypothesis poses, and the $M{+}1$ hypotheses (with $T_{\text{init}}$) are fused by the score head below.

\paragraph{Score-based fusion.} A score head evaluates each of the $M{+}1$ hypotheses (including $T_{\text{init}}$) on plane-warping costs
\begin{equation}
\label{eq:scorecost}
c^{r}_{h,m}=\big\|R_{h}\bm{n}^{2}_{m}-\bm{n}^{1}_{m}\big\|_{2},\qquad
c^{t}_{h,m}=\big\|(\bar d^{2}_{m}-(R_{h}\bm{n}^{2}_{m})^{\!\top}\bm{\tau}_{h})\,R_{h}\bm{n}^{2}_{m}-\bar d^{1}_{m}\bm{n}^{1}_{m}\big\|_{2},
\end{equation}
and produces softmax weights $\bm{w}_{r},\bm{w}_{t}$ over the hypotheses. The fused refined pose reads
\begin{equation}
\label{eq:fusion}
R_{\text{ref}}=\Pi_{\SO(3)}\!\Big(\sum_{h=0}^{M}w_{r,h}\,\rotsix(R_{h})\Big),\qquad
\bm{\tau}_{\text{ref}}=\sum_{h=0}^{M}w_{t,h}\,\bm{\tau}_{h},
\end{equation}
with $\Pi_{\SO(3)}$ Gram--Schmidt orthogonalization.

\paragraph{Training objective.} The refiner and the score head are optimized jointly with
\begin{equation}
\label{eq:loss}
\mathcal{L}_{\text{refine}} \;=\; \underbrace{\Big\|\!\sum_{m}\omega_{m}\,\bm{v}_{m}-\bm{v}^{\star}\Big\|_{1}}_{\mathcal{L}_{\text{FM}}}
\;+\;0.1\cdot\underbrace{\mathbb{E}_{t}\!\Big[(1{-}t)^{2}\!\sum_{m}\omega_{m}\,g_{m}\Big]}_{\mathcal{L}_{\text{geo}}}
\;+\;0.01\cdot\mathcal{L}_{\text{score}},
\end{equation}
where $\mathcal{L}_{\text{score}}$ denotes the hypothesis-selection loss of~\citep{nopesac}, concentrating $\bm{w}_{r}$ (resp.\ $\bm{w}_{t}$) on the hypothesis with the smallest rotation (resp.\ translation) error and penalizing off-diagonal entries of the correspondence-indexed cost matrix. The three terms encode complementary signals: $\mathcal{L}_{\text{FM}}$ drives the velocity regression on the pose residual, $\mathcal{L}_{\text{geo}}$ constrains the predicted trajectory to the $(\Delta\bm{n},\Delta d)\!\to\!\bm{0}$ manifold of Eq.~\eqref{eq:residual-prop}, and $\mathcal{L}_{\text{score}}$ supervises the soft hypothesis selection used in Eq.~\eqref{eq:fusion}. The temporal weight $(1{-}t)^{2}$ in $\mathcal{L}_{\text{geo}}$ concentrates geometric supervision on iterates near $\bm{z}_{0}$, where the warping residuals become linearly informative of the remaining pose error.

\subsection{Inference}
At test time we sample $\bm{z}_{1}\!\sim\!\mathcal{N}(0,I)$, broadcast it to the $M$ queries, and take a \emph{single} forward Euler step \emph{per query} along the learned velocity field,
\begin{equation}
\label{eq:inf}
\bm{z}_{0,m} \;=\; \bm{z}_{1} - \bm{v}_{m}(\bm{z}_{1},\,t{=}1), \qquad m=1,\dots,M,
\end{equation}
whose unit step covers the full $t\!\in\![0,1]$ interval. Decoding each $\bm{z}_{0,m}$ and composing with $T_{\text{init}}$ yields $M$ pose hypotheses $\{T_{m}\}_{m=1}^{M}$ in one evaluation of $F_{\theta}$; the score head then fuses them with $T_{\text{init}}$ through Eq.~\eqref{eq:fusion} into the refined pose $(R_{\text{ref}},\bm{\tau}_{\text{ref}})$. We take $R^{\star}\!=\!R_{\text{ref}}$ as the final rotation and recover the metric translation by the geometric step below, warm-started from $\bm{t}_{\text{init}}$.

\paragraph{Two-view translation refinement.} With $R$ fixed to $R^{\star}$, translation reduces to a near-linear problem. We lock the magnitude $s=\|\bm{t}_{\text{init}}\|$ at the scale from Eq.~\eqref{eq:sest} and optimize only the unit direction $\hat{\bm{t}}\in S^{2}$ via a Huber-robust objective combining the plane-offset residual $\Delta d_{m}$ and a $3$-D point-alignment residual $\Delta\bm{P}_{m}=\bm{P}^{1}_{m}-R^{\star}\bm{P}^{2}_{m}-s\hat{\bm{t}}$, where $\bm{P}^{v}_{m}$ is the back-projection of matched pixel $m$ in view $v$ through the dense depth that \textsc{PlanarSplatting} rasterizes from the plane-head primitives. The objective is regularized by an init-anchored Tikhonov prior on the direction:
\begin{equation}
\label{eq:trans-refine}
\hat{\bm{t}}^{\star} \;=\; \argmin_{\hat{\bm{t}}\in S^{2}}\;\sum_{m\in\mathcal{M}}\!\Big[\rho_{d}\!\big(\Delta d_{m}\big)+\rho_{p}\!\big(\|\Delta\bm{P}_{m}\|_{2}\big)\Big] \;+\; \lambda_{\text{reg}}\,\|\hat{\bm{t}}-\hat{\bm{t}}_{\text{init}}\|_{2}^{2}.
\end{equation}
$\rho_{d}$ and $\rho_{p}$ are Huber loss functions and $\lambda_{\text{reg}}$ is set adaptively from the init's self-consistency with the predicted plane field, so that an already-accurate init is protected (large $\lambda_{\text{reg}}$, solver returns $\approx\hat{\bm{t}}_{\text{init}}$) while a poor init lets the data terms dominate. The closed-form schedule and the construction of the dense $\bm{P}^{v}_{m}$ are detailed in App.~\ref{app:trans-solver}.

\paragraph{Multi-view aggregation.} For streaming $n$-view inputs, we run the diffusion refiner on every edge of the refinement window to obtain refined pairwise rotations $\{R^{\star}_{ij}\}$ while keeping the upstream's raw pairwise translation directions $\{\bm{t}^{\text{raw}}_{ij}\}$ unchanged. Per-frame absolute rotations are then recovered by rotation averaging~\citep{rotaveraging,rotaveraging_iccv} on $\{R^{\star}_{ij}\}$, and per-frame camera centers by a \textsc{Glomap}-style fix-rotation positioner~\citep{glomap} initialised from the raw translations; the explicit objectives are given in App.~\ref{app:trans-solver}. The two-view LM translation solver of Eq.~\eqref{eq:trans-refine} is bypassed at $n$-view scale---per-frame translations are determined globally by RA + the positioner from the dense edge set---so the same learned refiner (whose only role at $n$-view is to refine each $R_{ij}$) serves both regimes without retraining.

\section{Experiment}
\label{sec:experiment}
We evaluate \textsc{PosePlaner} as a plug-in refiner for both two-view and streaming $n$-view pose estimators across two benchmarks, ScanNet1500~\citep{scannet} and 7Scenes~\citep{7scenes}, in Sec.~\ref{sec:exp-twoview-new} and Sec.~\ref{sec:exp-nview-new}; an additional zero-shot two-view evaluation on Matterport3D~\citep{chang2017matterport3dlearningrgbddata} (\textsc{SparsePlanes}~\citep{sparseplane} test split) is reported in App.~\ref{app:eval-data} (Tab.~\ref{tab:matterport-twoview}).
Implementation and training details are in App.~\ref{app:impl}.
 
\subsection{Two-View Camera Pose Refinement}
\label{sec:exp-twoview-new}

\paragraph{Evaluation protocol.} We evaluate on ScanNet1500~\citep{scannet} ($1500$ standard test pairs from~\citep{superglue}) and 7Scenes~\citep{7scenes} ($4070$ pairs formed at frame gap $50$ across all $46$ sequences; zero-shot, not in our training mixture). We attach \textsc{PosePlaner} downstream of seven upstream pose sources: four correspondence-based solvers (\textsc{SuperGlue}~\citep{superglue}, \textsc{LoFTR}~\citep{loftr}, \textsc{ROMA}~\citep{roma}, \textsc{MASt3R}~\citep{mast3r}) and three diffusion-based pose estimators (\textsc{PoseDiffusion}~\citep{posediffusion}, \textsc{RayDiffusion}~\citep{raydiffusion}, \textsc{DiffusionSfM}~\citep{diffusionsfm}). Each correspondence-based upstream uses its own native pose solver followed by a two-view bundle adjustment to produce the init pose. We report pairwise rotation and translation AUC at $5^\circ$ and $10^\circ$ (higher is better).

\begin{table}[!t]
\centering
\small
\setlength{\tabcolsep}{2.5pt}
\caption{\textbf{Two-view pose refinement on ScanNet1500~\citep{scannet} and 7Scenes~\citep{7scenes}.} R/t: rotation/translation-direction AUC at $5^\circ$ and $10^\circ$ (\%, higher is better). \emph{Valid} gives the common evaluation subset within each block. \textsc{PosePlaner} uses \textsc{MASt3R} conditioning throughout (Tab.~\ref{tab:match-ablation}). Best per block in \textbf{bold}. \textsuperscript{$\dagger$}~Tested zero-shot.}
\label{tab:twoview}
\resizebox{\linewidth}{!}{%
\begin{tabular}{l|ccccc|ccccc}
\toprule
\multirow{2}{*}{Method} & \multicolumn{5}{c|}{ScanNet1500} & \multicolumn{5}{c}{7Scenes\textsuperscript{$\dagger$}} \\
                        & R@5 & R@10 & t@5 & t@10 & Valid & R@5 & R@10 & t@5 & t@10 & Valid \\
\midrule
\multicolumn{11}{l}{\textit{Correspondence-based solvers}} \\
\textsc{SuperGlue}~\citep{superglue}                    & 43.9 & 62.4 & 21.3 & 39.6 & $1427/1500$ & 42.3 & 65.9 & 18.9 & 37.6 & $4014/4070$ \\
\textsc{SuperGlue}\,+\,\textsc{NopeSAC} head & 34.9 & 58.7 & 7.4 & 19.9 & & 34.3 & 61.5 & 7.1 & 19.9 & \\
\textsc{SuperGlue}\,+\,\textsc{PosePlaner}              & \textbf{48.9} & \textbf{68.9} & \textbf{21.5} & \textbf{40.4} & & \textbf{45.8} & \textbf{69.7} & \textbf{19.0} & \textbf{37.6} & \\
\midrule
\textsc{LoFTR}~\citep{loftr}                            & 11.5 & 25.0 & 2.1 & 5.6 & $1028/1500$ & 17.7 & 34.2 & 3.7 & 10.1 & $3585/4070$ \\
\textsc{LoFTR}\,+\,\textsc{NopeSAC} head & 26.7 & 49.9 & 1.6 & 4.6 & & 27.1 & 52.8 & 2.6 & 8.3 & \\
\textsc{LoFTR}\,+\,\textsc{PosePlaner}                  & \textbf{29.4} & \textbf{52.9} & \textbf{3.5} & \textbf{9.0} & & \textbf{31.2} & \textbf{56.8} & \textbf{4.4} & \textbf{12.7} & \\
\midrule
\textsc{ROMA}~\citep{roma}                              & \textbf{66.7} & \textbf{81.2} & \textbf{39.1} & \textbf{58.8} & $1500/1500$ & 50.7 & 73.3 & \textbf{25.6} & \textbf{46.7} & $4068/4070$ \\
\textsc{ROMA}\,+\,\textsc{NopeSAC} head & 38.6 & 61.8 & 9.8 & 25.6 & & 35.4 & 62.6 & 8.4 & 23.2 & \\
\textsc{ROMA}\,+\,\textsc{PosePlaner}                   & 64.1 & 80.1 & 37.0 & 57.3 & & \textbf{52.0} & \textbf{74.1} & 25.4 & 46.1 & \\
\midrule
\textsc{MASt3R}~\citep{mast3r}                          & \textbf{66.5} & \textbf{82.4} & \textbf{41.3} & \textbf{61.6} & $1495/1500$ & 50.6 & 73.3 & \textbf{26.7} & \textbf{47.9} & $4065/4070$ \\
\textsc{MASt3R}\,+\,\textsc{NopeSAC} head & 38.2 & 61.6 & 10.2 & 26.0 & & 35.4 & 62.7 & 8.8 & 23.2 & \\
\textsc{MASt3R}\,+\,\textsc{PosePlaner}                 & 65.0 & 81.7 & 38.6 & 59.8 & & \textbf{51.5} & \textbf{74.0} & 26.0 & 47.0 & \\
\midrule
\multicolumn{11}{l}{\textit{Diffusion-based pose estimators (visual conditioning)}} \\
\textsc{PoseDiff.}~\citep{posediffusion}                & 2.2 & 9.0 & 0.5 & 1.2 & $1500/1500$ & 3.2 & 12.2 & 0.4 & 1.5 & $4070/4070$ \\
\textsc{PoseDiff.}\,+\,\textsc{NopeSAC} head           & 18.2 & \textbf{41.3} & 0.8 & 3.0 & & 18.0 & 42.4 & 0.4 & 1.5 & \\
\textsc{PoseDiff.}\,+\,\textsc{PosePlaner}              & \textbf{18.9} & 38.8 & \textbf{2.2} & \textbf{6.0} & & \textbf{20.6} & \textbf{44.6} & \textbf{1.0} & \textbf{3.8} & \\
\midrule
\textsc{RayDiff.}~\citep{raydiffusion}                  & 11.7 & 33.4 & 1.7 & 6.9 & $1500/1500$ & 10.3 & 32.7 & 1.0 & 4.4 & $4070/4070$ \\
\textsc{RayDiff.}\,+\,\textsc{NopeSAC} head            & 6.5 & 19.2 & 0.3 & 1.3 & & 7.4 & 20.5 & 0.2 & 0.7 & \\
\textsc{RayDiff.}\,+\,\textsc{PosePlaner}               & \textbf{25.3} & \textbf{50.5} & \textbf{5.4} & \textbf{15.5} & & \textbf{23.3} & \textbf{51.4} & \textbf{2.2} & \textbf{7.6} & \\
\midrule
\textsc{DiffusionSfM}~\citep{diffusionsfm}              & 37.8 & 64.3 & 3.0 & 16.0 & $1500/1500$ & 39.6 & 66.6 & 3.9 & 16.5 & $4070/4070$ \\
\textsc{DiffusionSfM}\,+\,\textsc{NopeSAC} head        & 35.0 & 59.5 & 4.1 & 13.8 & & 33.5 & 61.1 & 4.3 & 14.9 & \\
\textsc{DiffusionSfM}\,+\,\textsc{PosePlaner}           & \textbf{43.6} & \textbf{68.6} & \textbf{12.8} & \textbf{30.5} & & \textbf{43.6} & \textbf{69.1} & \textbf{6.2} & \textbf{20.3} & \\
\bottomrule
\end{tabular}%
}
\end{table}

\paragraph{NopeSAC configuration.} We evaluate the frozen \textsc{NopeSAC}~\citep{nopesac} refinement head using \textsc{MASt3R} patch matches and our predicted plane pairs. Initial poses are post-BA estimates for correspondence-based upstreams and independently rerun predictions for diffusion-based upstreams on the same test pairs. We bypass its PlaneTR frontend, native matching, and pixel camera prediction, retaining the official plane warp, AIM, 50-slot head, and soft fusion. The released ScanNet checkpoint is used without retraining on either benchmark. We report its directly predicted rotation and translation; \textsc{PosePlaner} retains its geometric translation refinement (Eq.~\eqref{eq:trans-refine}). This evaluates the released head under shared frontend inputs.

\paragraph{Quantitative results.} In Tab.~\ref{tab:twoview}, \textsc{PosePlaner} outperforms the \textsc{NopeSAC} head at both reported rotation and translation thresholds for all four correspondence-based upstreams on both benchmarks. The visual-conditioning outputs of \textsc{PoseDiffusion}, \textsc{RayDiffusion}, and \textsc{DiffusionSfM} also improve substantially after refinement, indicating that the plane-warping residuals carry information complementary to visual-only priors.

\begin{table}[t]
\centering
\small
\setlength{\tabcolsep}{5pt}
\caption{\textbf{Controlled comparison with bundle adjustment.} BA and \textsc{PosePlaner} use the same initial poses and \textsc{MASt3R} correspondences. R@5, R@10, and R@20 are rotation AUC at $5^\circ$, $10^\circ$, and $20^\circ$ (\%, higher is better). Best results in each initialization group are \textbf{bold}. Table~\ref{tab:twoview}'s correspondence-based initializations have already undergone BA.}
\label{tab:ba-control}
\begin{tabular}{l|ccc|ccc}
\toprule
\multirow{2}{*}{Initialization / refinement} & \multicolumn{3}{c|}{ScanNet1500} & \multicolumn{3}{c}{7Scenes} \\
 & R@5 & R@10 & R@20 & R@5 & R@10 & R@20 \\
\midrule
\textsc{LoFTR} (pre-BA) & 11.5 & 25.2 & 44.1 & 17.8 & 34.3 & 53.4 \\
\quad + BA & 13.4 & 26.5 & 44.9 & 24.4 & 39.6 & 56.7 \\
\quad + \textsc{PosePlaner} & \textbf{29.5} & \textbf{53.1} & \textbf{70.7} & \textbf{31.2} & \textbf{56.8} & \textbf{75.4} \\
\midrule
\textsc{MASt3R} (pre-BA) & 66.0 & 82.1 & 90.9 & 49.6 & 72.7 & 86.1 \\
\quad + BA & \textbf{66.5} & \textbf{82.4} & \textbf{91.1} & 50.6 & 73.3 & 86.4 \\
\quad + \textsc{PosePlaner} & 64.7 & 81.5 & 90.6 & \textbf{50.9} & \textbf{73.6} & \textbf{86.5} \\
\bottomrule
\end{tabular}
\end{table}

\paragraph{Comparison with classical refinement.} Both methods start from the same \textsc{LoFTR}- or \textsc{MASt3R}-derived pre-BA poses and share the same \textsc{MASt3R} matching output. BA uses \textsc{MASt3R}'s point correspondences, while \textsc{PosePlaner} uses patch-level correspondences. Overall, \textsc{PosePlaner} substantially outperforms BA with \textsc{LoFTR} initialization on both datasets and achieves higher rotation AUC with \textsc{MASt3R} initialization on 7Scenes (Tab.~\ref{tab:ba-control}).

\begin{figure*}[!t]
\centering
\vspace{-4pt}
\begin{minipage}[c]{0.03\linewidth}
  \centering
  \rotatebox{90}{\footnotesize \textsc{LoFTR}}
\end{minipage}%
\begin{minipage}[c]{0.31\linewidth}
  \centering
  \includegraphics[width=\linewidth]{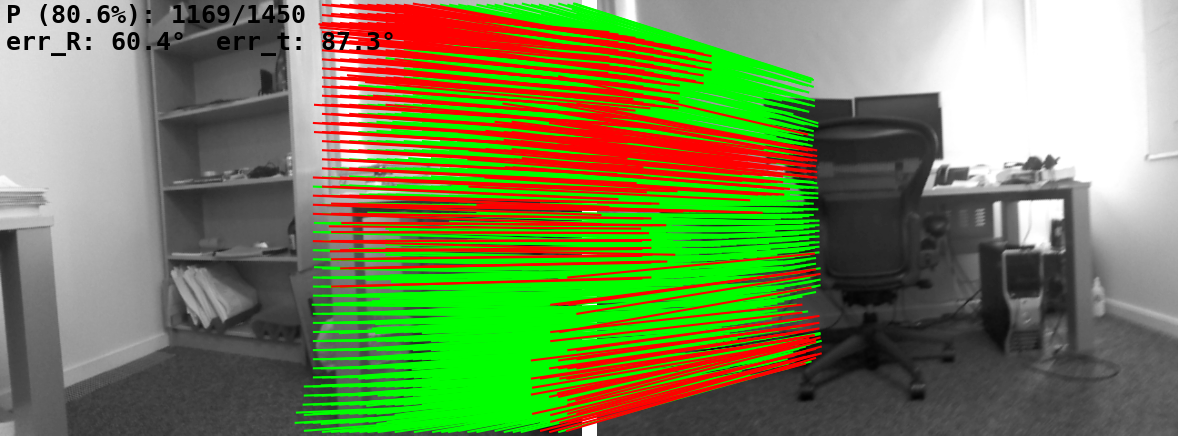}
\end{minipage}\hfill
\begin{minipage}[c]{0.31\linewidth}
  \centering
  \includegraphics[width=\linewidth]{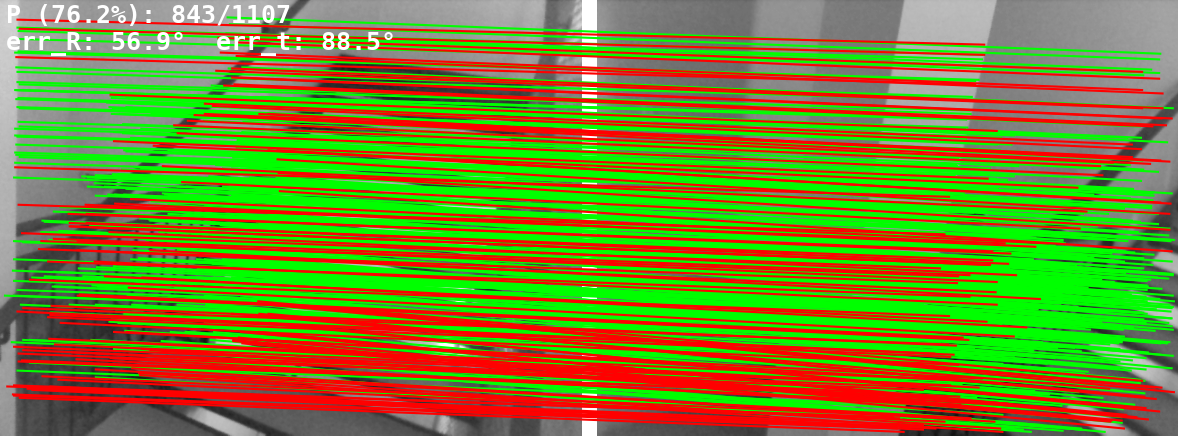}
\end{minipage}\hfill
\begin{minipage}[c]{0.31\linewidth}
  \centering
  \includegraphics[width=\linewidth]{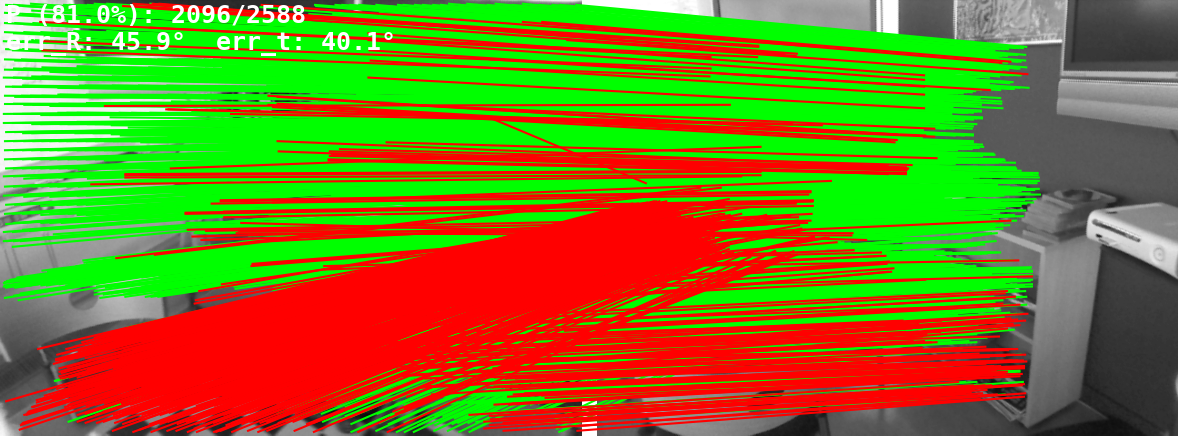}
\end{minipage}\\[2pt]
\begin{minipage}[c]{0.03\linewidth}
  \centering
  \rotatebox{90}{\footnotesize + Ours}
\end{minipage}%
\begin{minipage}[c]{0.31\linewidth}
  \centering
  \includegraphics[width=\linewidth]{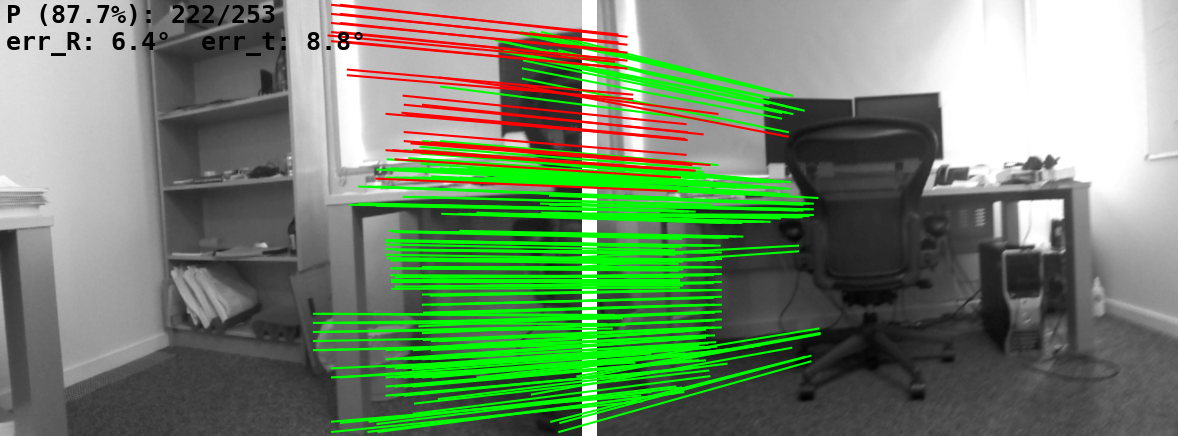}
\end{minipage}\hfill
\begin{minipage}[c]{0.31\linewidth}
  \centering
  \includegraphics[width=\linewidth]{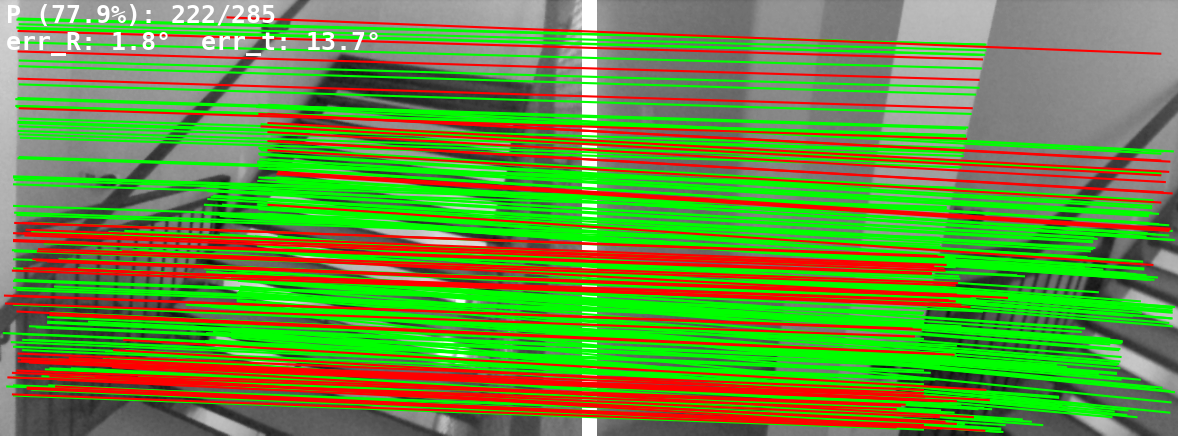}
\end{minipage}\hfill
\begin{minipage}[c]{0.31\linewidth}
  \centering
  \includegraphics[width=\linewidth]{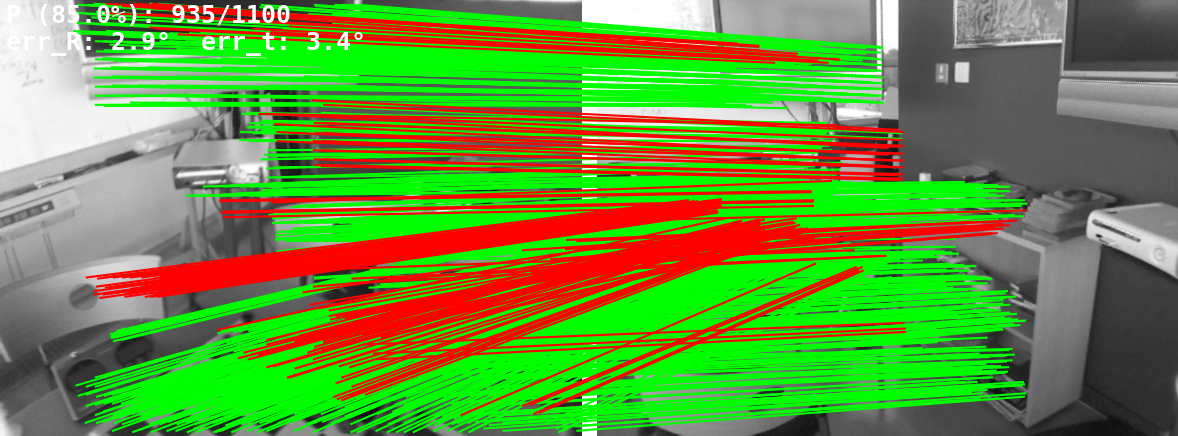}
\end{minipage}
\caption{\textbf{Planar primitives as a stand-alone match filter.} Three \textsc{7Scenes} pairs, \textsc{LoFTR}~\citep{loftr} upstream. \textbf{Top:} raw \textsc{LoFTR} matches. \textbf{Bottom:} matches surviving the cut $\lVert\Delta\bm{n}\rVert+\lvert\Delta d\rvert\!\le\!1$. \textcolor{green}{\textbf{Green}}/\textcolor{red}{\textbf{red}} mark RANSAC inliers/outliers; top-left numbers are inlier ratio and pose error.}
\label{fig:outlier-robust}
\vspace{-6pt}
\end{figure*}

\paragraph{Plane primitives also work as a stand-alone match filter.} The per-patch plane primitives also help \emph{without} invoking the refiner: applying the cut $\|\Delta\bm{n}\|+|\Delta d|\!\le\!1$ to \textsc{LoFTR} matches removes a large fraction of RANSAC outliers while preserving inliers, and re-solving pose on the filtered subset substantially reduces rotation and translation error (Fig.~\ref{fig:outlier-robust}).

\subsection{Multi-View Refinement on Streaming Pose Estimators}
\label{sec:exp-nview-new}

\paragraph{Evaluation protocol.} On ScanNet~\citep{scannet} and 7Scenes~\citep{7scenes} (zero-shot), we form $N$-frame trajectories at $N\!\in\!\{200,500,2000\}$ at frame gap $2$; the streaming baseline ingests the first $N$ frames unrefined and \textsc{PosePlaner} refines the last $100$ pairwise. Both raw and refined per-pair $(R,\bm{t})$ pass through the \emph{same} multi-view backend (rotation averaging + fix-rotation \textsc{Glomap}-style positioner~\citep{glomap}; full details in App.~\ref{app:training-recipe}), so the only difference is whether each input pair is refined. We evaluate four streaming reconstructors (\textsc{CUT3R}~\citep{cut3r}, \textsc{TTT3R}~\citep{ttt3r}, \textsc{WinT3R}~\citep{wint3r}, \textsc{Spann3R}~\citep{spann3r}) and report pairwise rotation and translation AUC at $5^\circ$ and $20^\circ$ over the $\binom{100}{2}$ tail pairs.

\begin{table*}[t]
\centering
\caption{\textbf{$n$-view pose accuracy on streaming ScanNet~\citep{scannet} and 7Scenes~\citep{7scenes}.} Pairwise R and t AUC at $5^\circ$ and $20^\circ$ over the $\binom{100}{2}$ pairs in the refined tail. Higher is better. {\textsuperscript{$\dagger$}}~Tested zero-shot.}
\label{tab:nview}
\resizebox{\linewidth}{!}{%
\begin{tabular}{l|cccc|cccc|cccc}
\toprule
\multirow{2}{*}{Method}
  & \multicolumn{4}{c|}{$N{=}200$}
  & \multicolumn{4}{c|}{$N{=}500$}
  & \multicolumn{4}{c}{$N{=}2000$} \\
  & R@5 & R@20 & t@5 & t@20 & R@5 & R@20 & t@5 & t@20 & R@5 & R@20 & t@5 & t@20 \\
\midrule
\multicolumn{13}{l}{\emph{ScanNet}} \\
\textsc{CUT3R}                                & 23.3 & 65.0 & 4.1 & 24.0 & 16.5 & 55.7 & 2.2 & 19.0 & 17.3 & 53.4 & 1.0 & 13.2 \\
\textsc{CUT3R}\,+\,\textsc{PosePlaner}        & \textbf{40.5} & \textbf{80.3} & \textbf{7.7} & \textbf{38.0} & \textbf{37.3} & \textbf{78.9} & \textbf{6.3} & \textbf{38.4} & \textbf{37.3} & \textbf{79.8} & \textbf{3.4} & \textbf{31.2} \\
\textsc{TTT3R}                                & 46.2 & 83.1 & \textbf{11.1} & 42.7 & 36.1 & 79.3 & 8.0 & 38.7 & 27.5 & 72.5 & 3.4 & 28.6 \\
\textsc{TTT3R}\,+\,\textsc{PosePlaner}        & \textbf{50.7} & \textbf{85.6} & 10.8 & \textbf{45.9} & \textbf{47.1} & \textbf{84.3} & \textbf{10.8} & \textbf{47.3} & \textbf{48.2} & \textbf{85.0} & \textbf{10.4} & \textbf{41.8} \\
\textsc{WinT3R}                               & 37.7 & 77.1 & 8.4 & 37.4 & 28.3 & 66.3 & 5.5 & 28.8 & 19.2 & 48.6 & 3.1 & 16.8 \\
\textsc{WinT3R}\,+\,\textsc{PosePlaner}       & \textbf{45.9} & \textbf{83.7} & \textbf{9.5} & \textbf{42.7} & \textbf{42.5} & \textbf{80.7} & \textbf{7.8} & \textbf{41.3} & \textbf{27.9} & \textbf{65.0} & \textbf{3.9} & \textbf{20.8} \\
\textsc{Spann3R}                              & 32.4 & 69.2 & 6.9 & 31.9 & 20.0 & 55.6 & 3.5 & 21.4 & 19.4 & 52.8 & 1.8 & 16.0 \\
\textsc{Spann3R}\,+\,\textsc{PosePlaner}      & \textbf{45.6} & \textbf{82.2} & \textbf{9.3} & \textbf{42.2} & \textbf{38.8} & \textbf{76.4} & \textbf{7.0} & \textbf{37.1} & \textbf{37.1} & \textbf{77.4} & \textbf{4.2} & \textbf{30.9} \\
\midrule
\multicolumn{13}{l}{\emph{7Scenes\textsuperscript{$\dagger$}}} \\
\textsc{CUT3R}                                & 22.0 & 65.4 & 3.2 & 24.8 & 16.2 & 54.4 & 2.0 & 17.4 & 12.0 & 47.3 & 0.4 & 7.7 \\
\textsc{CUT3R}\,+\,\textsc{PosePlaner}        & \textbf{35.1} & \textbf{77.7} & \textbf{5.9} & \textbf{34.8} & \textbf{35.1} & \textbf{78.0} & \textbf{4.7} & \textbf{35.7} & \textbf{28.1} & \textbf{71.0} & \textbf{2.2} & \textbf{26.0} \\
\textsc{TTT3R}                                & 44.1 & \textbf{83.9} & \textbf{11.2} & \textbf{48.8} & 36.0 & 79.4 & 7.9 & 40.5 & 38.4 & 82.1 & 3.4 & 27.1 \\
\textsc{TTT3R}\,+\,\textsc{PosePlaner}        & \textbf{44.5} & 83.3 & 11.0 & 47.8 & \textbf{41.9} & \textbf{83.7} & \textbf{9.7} & \textbf{47.3} & \textbf{41.7} & \textbf{83.6} & \textbf{4.0} & \textbf{31.2} \\
\textsc{WinT3R}                               & 40.0 & 81.7 & 7.4 & 37.8 & 44.2 & 82.9 & 12.7 & 45.7 & 41.6 & 78.7 & 12.7 & 35.5 \\
\textsc{WinT3R}\,+\,\textsc{PosePlaner}       & \textbf{43.9} & \textbf{83.4} & \textbf{7.6} & \textbf{42.4} & \textbf{46.3} & \textbf{85.0} & \textbf{12.8} & \textbf{49.4} & \textbf{42.9} & \textbf{81.2} & \textbf{15.2} & \textbf{38.4} \\
\textsc{Spann3R}                              & 34.5 & 77.2 & 7.7 & 37.3 & 35.7 & 76.5 & 8.1 & 38.9 & 39.6 & 77.0 & \textbf{9.2} & 32.5 \\
\textsc{Spann3R}\,+\,\textsc{PosePlaner}      & \textbf{40.8} & \textbf{81.8} & \textbf{8.9} & \textbf{42.9} & \textbf{42.1} & \textbf{82.6} & \textbf{9.4} & \textbf{46.8} & \textbf{43.7} & \textbf{79.7} & 8.2 & \textbf{33.9} \\
\bottomrule
\end{tabular}%
}
\vspace{-6pt}
\end{table*}

\begin{figure*}[!t]
\centering
\vspace{-4pt}
\begin{minipage}[c]{0.258\linewidth}
  \centering
  \begin{tabular}[b]{@{}c@{\hspace{1pt}}c@{}}
    \parbox[b][0.95in][c]{0.13in}{\centering\rotatebox{90}{\small Scene0757\_00}} &
    \includegraphics[height=0.95in]{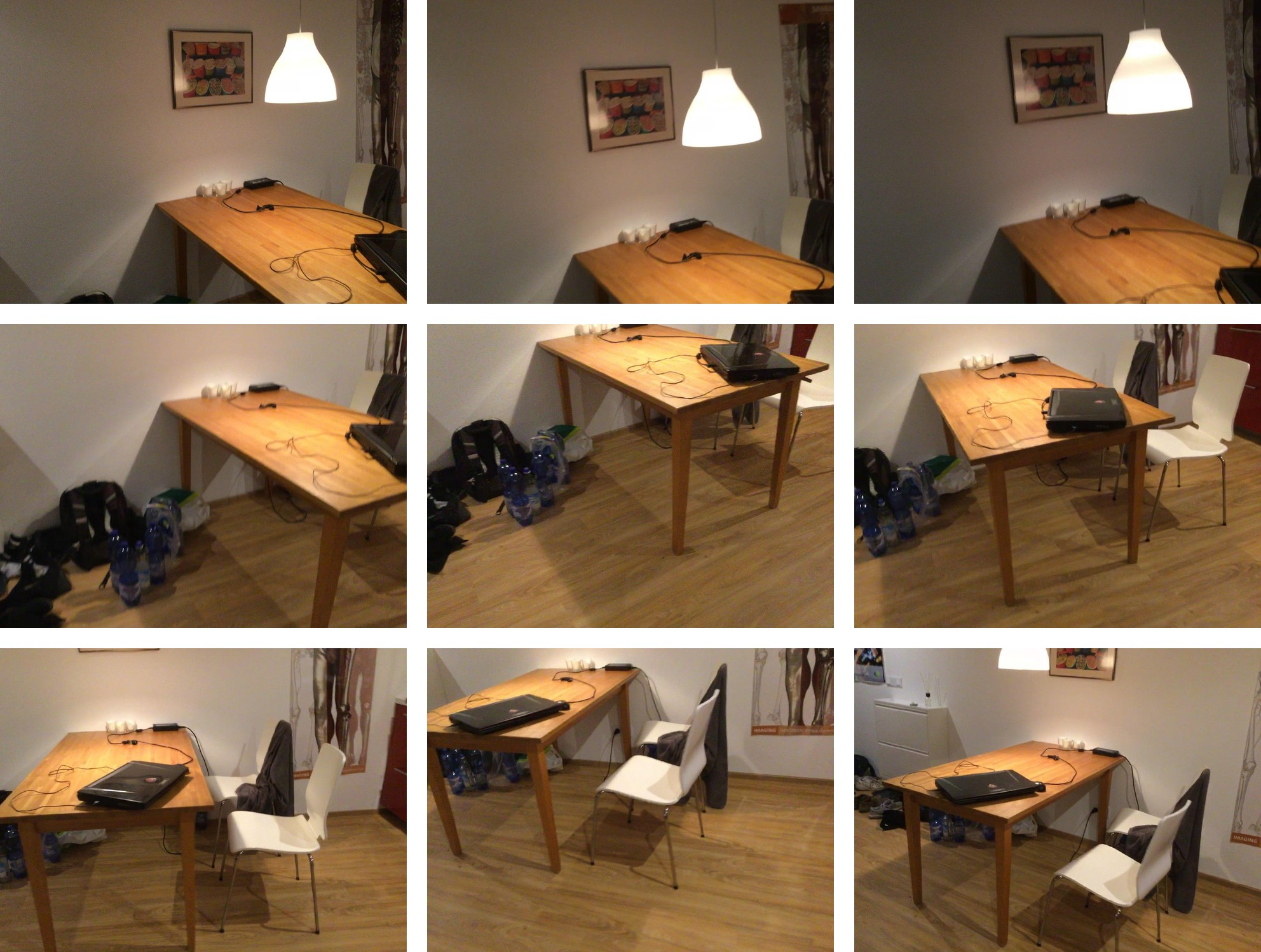}
  \end{tabular}\\[-1pt]
  {\small \phantom{\textsc{CUT3R} $\pm$\,ours}}%
\end{minipage}%
\begin{minipage}[c]{0.245\linewidth}
  \centering
  \includegraphics[width=\linewidth,height=1.35in,keepaspectratio]{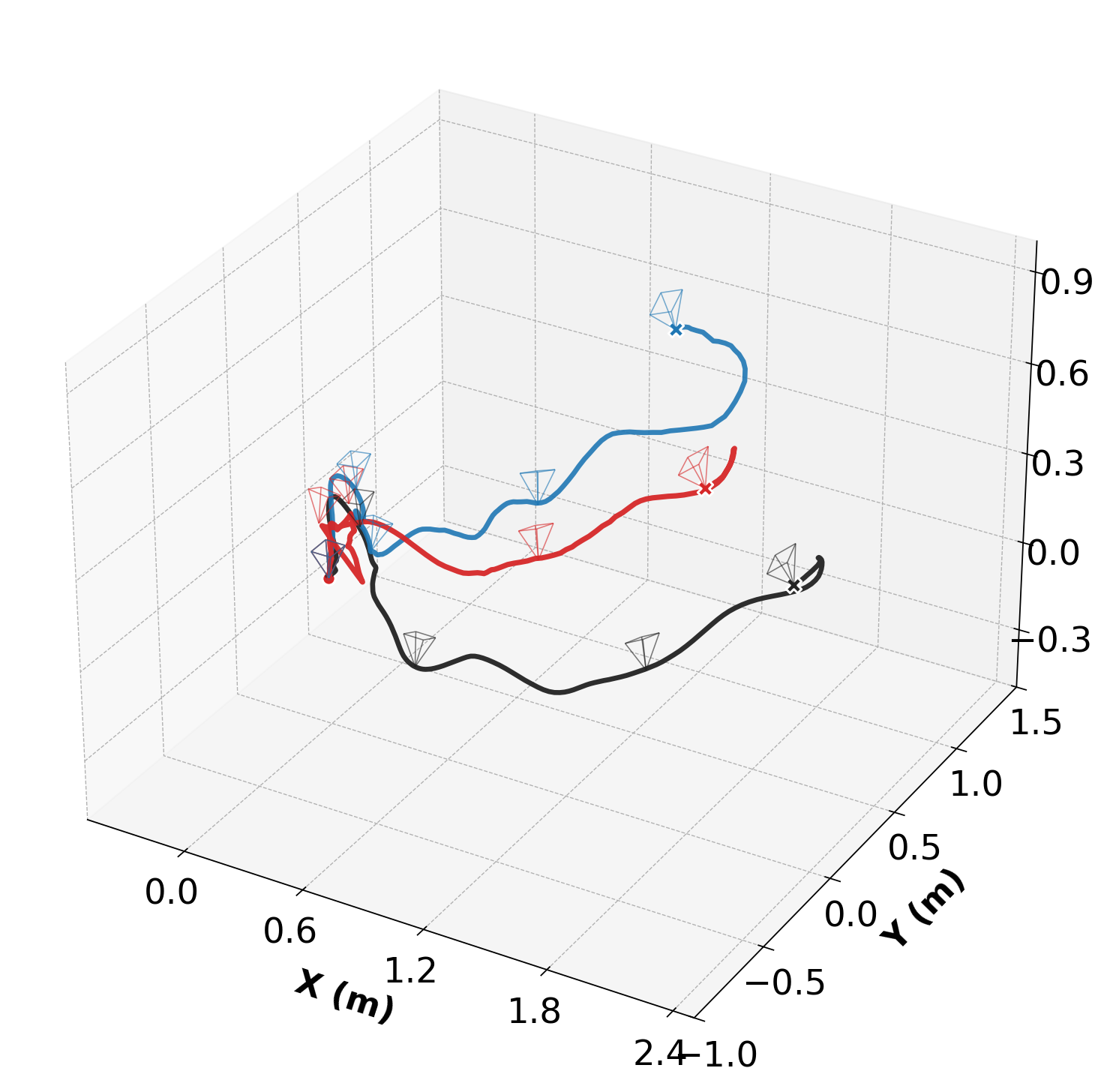}\\[-1pt]
  {\small \textsc{CUT3R} $\pm$\,ours}
\end{minipage}%
\begin{minipage}[c]{0.245\linewidth}
  \centering
  \includegraphics[width=\linewidth,height=1.35in,keepaspectratio]{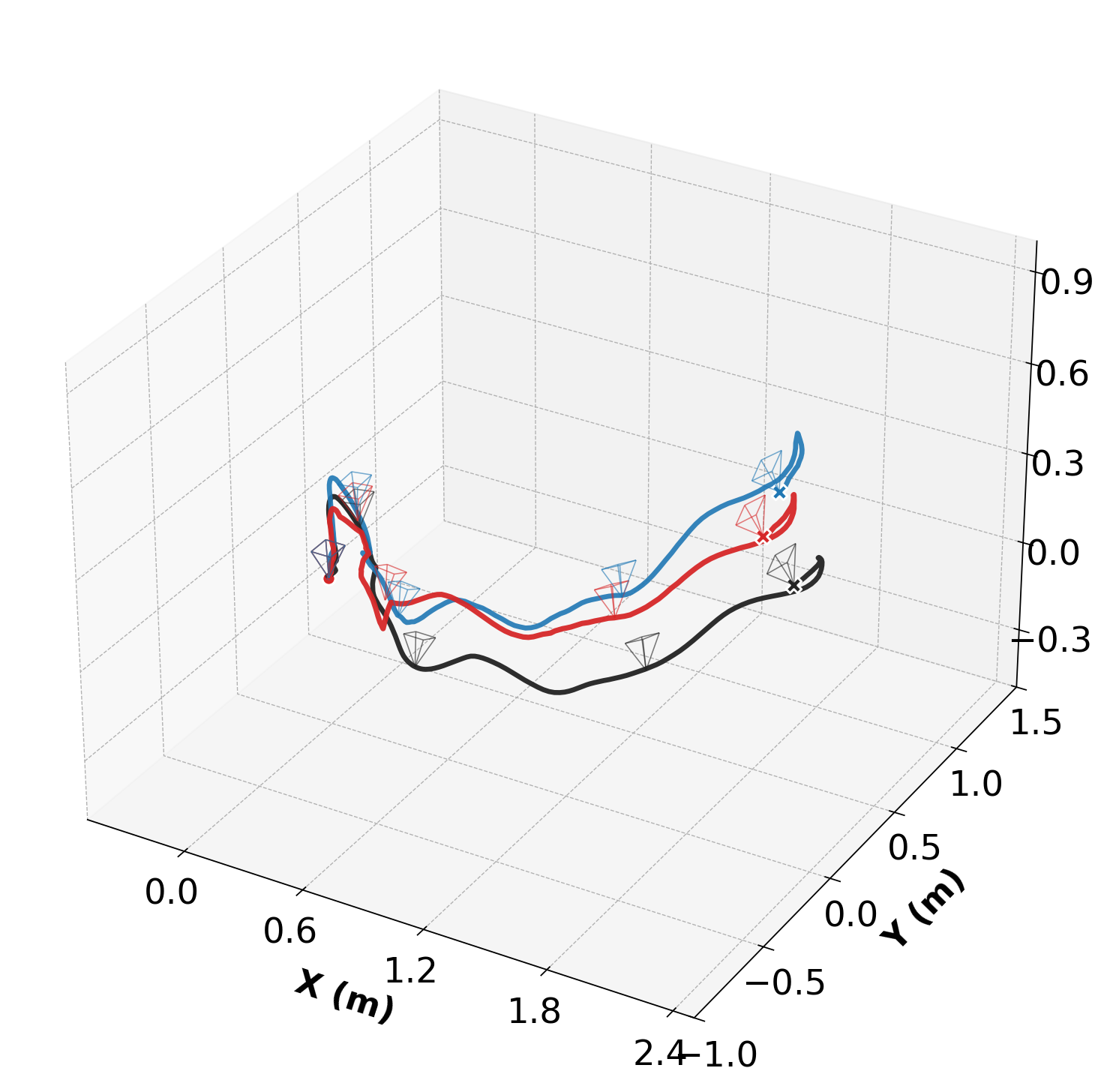}\\[-1pt]
  {\small \textsc{TTT3R} $\pm$\,ours}
\end{minipage}%
\begin{minipage}[c]{0.245\linewidth}
  \centering
  \includegraphics[width=\linewidth,height=1.35in,keepaspectratio]{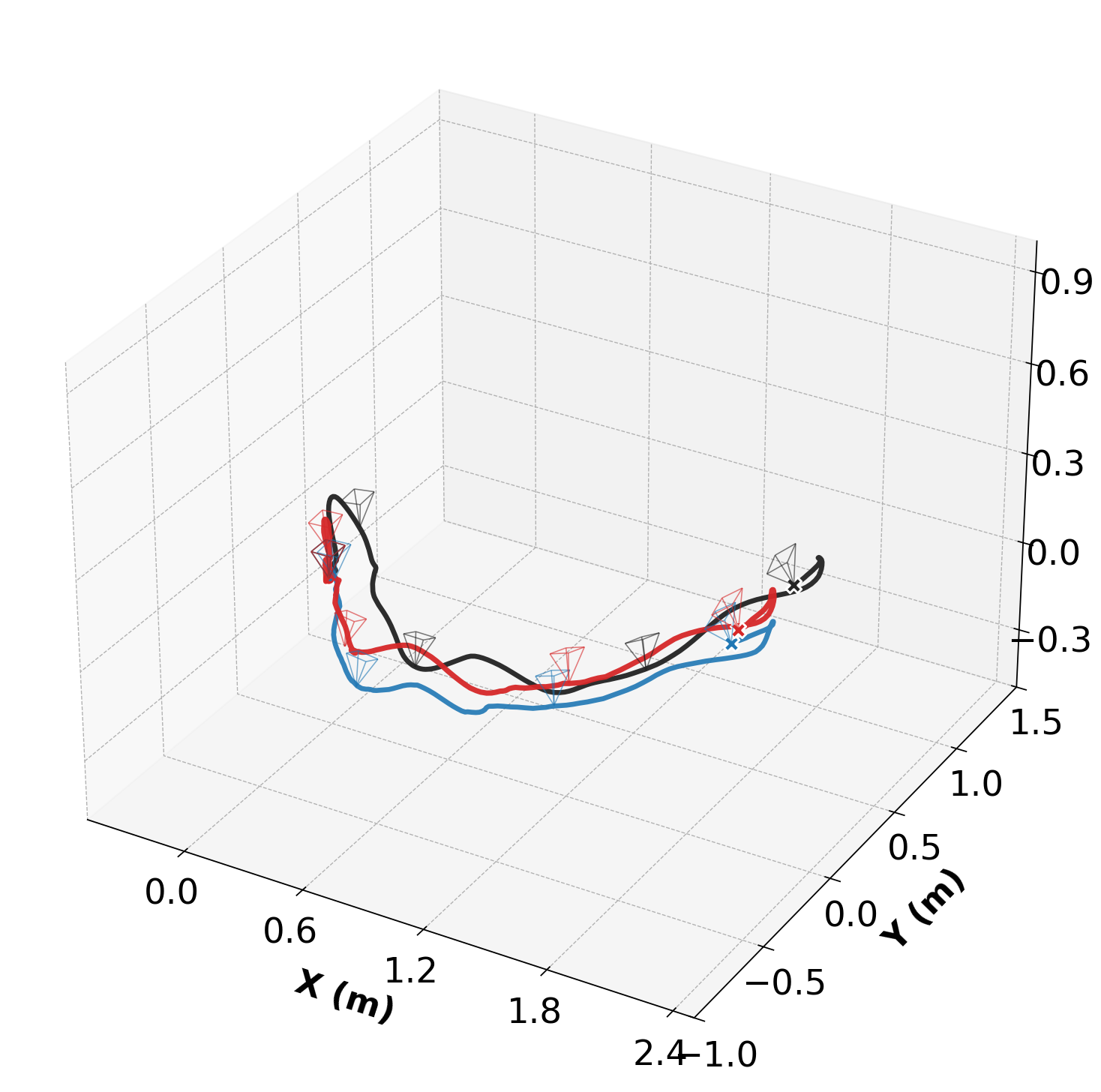}\\[-1pt]
  {\small \textsc{WinT3R} $\pm$\,ours}
\end{minipage}
\caption{\textbf{Streaming pose refinement on a representative ScanNet~\citep{scannet} sequence (Scene0757\_00).} Left: sample frames. Right three panels: estimated camera trajectories under \textsc{CUT3R}~\citep{cut3r}, \textsc{TTT3R}~\citep{ttt3r}, and \textsc{WinT3R}~\citep{wint3r}. \textbf{Black}: ground-truth; \textcolor{blue}{\textbf{blue}}: raw baseline; \textcolor{red}{\textbf{red}}: after \textsc{PosePlaner} refinement. $\bullet$~start, $\times$~end.}
\label{fig:streaming-curves}
\vspace{-6pt}
\end{figure*}

\paragraph{Quantitative results.} Tab.~\ref{tab:nview} sweeps $N$ to vary the drift entering refinement. At $N{=}200$ the baseline has accumulated little drift and refinement is roughly at parity, while at $N{=}500$ and especially $N{=}2000$ both rotation and translation AUC improve clearly across all four baselines on both datasets, with the gap widening monotonically with $N$. This converts streaming pose estimation from an unbounded-error regime into a bounded-error one at the cost of a single refinement pass over the last $100$ frames.

\paragraph{Qualitative comparisons.} Fig.~\ref{fig:streaming-curves} visualizes a representative ScanNet trajectory under our sliding-window refinement protocol (full details in App.~\ref{app:streaming-extra}). The raw streaming baselines (blue) drift progressively from ground truth (black); after refinement (red) the refined tail snaps back into close agreement across all three baselines (\textsc{CUT3R}, \textsc{TTT3R}, \textsc{WinT3R}), mirroring the tail-$100$-frame AUC improvements that Tab.~\ref{tab:nview} reports. Additional sequences are in App.~\ref{app:streaming-extra}.

\paragraph{Zero-shot outdoor generalization.}
We evaluate zero-shot multi-view refinement on Oxford Spires~\citep{DBLP:journals/ijrr/TaoMZWFF26} and MatrixCity~\citep{DBLP:conf/iccv/0002JXX0L023} with CUT3R and TTT3R at $N\in\{200,500,2000\}$ (Tab.~\ref{tab:outdoor-nview}). Both the plane predictor and pose refiner are trained only on indoor data and kept frozen. Rotation AUC improves at $N=500$ and $N=2000$, while gains at $N=200$ are less consistent and some translation metrics decline. Additional tests on CrossLoc~\citep{DBLP:conf/cvpr/YanZRLD22} drone sequences at $N=200$ and $N=500$, together with trajectory construction details, are provided in App.~\ref{app:eval-data}.

\begin{table*}[t]
\centering
\small
\setlength{\tabcolsep}{4pt}
\renewcommand{\arraystretch}{1.0}
\caption{\textbf{Zero-shot outdoor multi-view refinement.} R@5/R@20 and t@5/t@20 are rotation and translation-direction AUC at $5^\circ$/$20^\circ$ (\%, $\uparrow$). Better results within each pair are \textbf{bold}. \textsuperscript{$\dagger$}~Tested zero-shot.}
\label{tab:outdoor-nview}
\resizebox{\textwidth}{!}{%
\begin{tabular}{l|cccc|cccc|cccc}
\toprule
\multirow{2}{*}{Method} & \multicolumn{4}{c|}{$N=200$} & \multicolumn{4}{c|}{$N=500$} & \multicolumn{4}{c}{$N=2000$} \\
 & R@5 & R@20 & t@5 & t@20 & R@5 & R@20 & t@5 & t@20 & R@5 & R@20 & t@5 & t@20 \\
\midrule
\multicolumn{13}{l}{\textit{Oxford Spires}\textsuperscript{$\dagger$}} \\
\textsc{CUT3R} & 59.1 & 85.1 & 28.9 & 66.2 & 35.5 & 75.7 & 8.1 & 52.5 & 34.0 & 70.1 & 11.6 & 46.5 \\
\textsc{CUT3R}\,+\,\textsc{PosePlaner} & \textbf{65.8} & \textbf{90.8} & \textbf{33.5} & \textbf{70.5} & \textbf{50.4} & \textbf{82.8} & \textbf{22.4} & \textbf{64.1} & \textbf{53.3} & \textbf{85.8} & \textbf{22.7} & \textbf{67.4} \\
\textsc{TTT3R} & \textbf{73.6} & \textbf{92.7} & \textbf{45.8} & \textbf{77.3} & 59.7 & 88.5 & 27.5 & \textbf{74.9} & 37.2 & 76.5 & 11.4 & 53.2 \\
\textsc{TTT3R}\,+\,\textsc{PosePlaner} & 72.3 & 92.5 & 42.0 & 74.9 & \textbf{62.9} & \textbf{89.2} & \textbf{29.8} & 74.0 & \textbf{51.3} & \textbf{85.6} & \textbf{19.9} & \textbf{67.9} \\
\midrule
\multicolumn{13}{l}{\textit{MatrixCity}\textsuperscript{$\dagger$}} \\
\textsc{CUT3R} & 13.2 & 48.5 & 4.8 & 20.1 & 12.4 & 49.3 & 3.1 & 15.9 & 9.9 & 40.4 & 3.7 & 15.2 \\
\textsc{CUT3R}\,+\,\textsc{PosePlaner} & \textbf{16.4} & \textbf{55.6} & \textbf{5.6} & \textbf{22.8} & \textbf{15.1} & \textbf{53.5} & \textbf{3.5} & \textbf{17.8} & \textbf{15.5} & \textbf{54.0} & \textbf{8.2} & \textbf{22.2} \\
\textsc{TTT3R} & 18.1 & 52.7 & 5.2 & 18.8 & 10.5 & 45.3 & 4.3 & 18.7 & 12.7 & 47.2 & 3.8 & 15.8 \\
\textsc{TTT3R}\,+\,\textsc{PosePlaner} & \textbf{20.6} & \textbf{56.9} & \textbf{6.6} & \textbf{23.0} & \textbf{14.4} & \textbf{49.7} & \textbf{5.2} & \textbf{19.4} & \textbf{17.2} & \textbf{54.5} & \textbf{7.3} & \textbf{20.2} \\
\bottomrule
\end{tabular}%
}
\end{table*}

\subsection{Ablation Study}
\label{sec:exp-ablation}

We ablate \textsc{PosePlaner} along three axes within Tab.~\ref{tab:ablation-components}: (i) what conditioning the refiner needs (\emph{w/o} $\{\bm{g}_m\}$, $\bm{c}$), (ii) how the $M{+}1$ pose hypotheses are aggregated (\emph{w/o score head}, i.e., the softmax-weighted score-head fusion is replaced by a uniform average), and (iii) how the pose residual is parameterized (\emph{w/o translation} drops the translation channel from $\bm{z}_0$, leaving a rot-only $\bm{z}_0\!\in\!\mathbb{R}^{6}$; \emph{w/o rot6d} replaces the rot6d rotation latent with a unit quaternion in $\mathbb{R}^{4}$).

\begin{table}[t]
\centering
\small
\setlength{\tabcolsep}{3pt}
\caption{\textbf{Ablation study on ScanNet1500} with \textsc{SuperGlue} and \textsc{LoFTR} upstream. R and t AUC at $\tau\!\in\!\{5,10,20\}^\circ$; higher is better, best per column in \textbf{bold}. Removing the pose token $\bm{c}$ forces the model to estimate pose without the upstream init, so the \textsc{SuperGlue} and \textsc{LoFTR} blocks of that row are identical by construction.}
\label{tab:ablation-components}
\resizebox{\linewidth}{!}{%
\begin{tabular}{l|cccccc|cccccc}
\toprule
\multirow{2}{*}{Variant} & \multicolumn{6}{c|}{Init: \textsc{SuperGlue}} & \multicolumn{6}{c}{Init: \textsc{LoFTR}} \\
\cmidrule(lr){2-7} \cmidrule(lr){8-13}
 & R@5 & R@10 & R@20 & t@5 & t@10 & t@20 & R@5 & R@10 & R@20 & t@5 & t@10 & t@20 \\
\midrule
Full (ours)             & \textbf{48.9} & \textbf{68.9} & \textbf{82.1} & \textbf{21.5} & \textbf{40.4} & \textbf{59.5} & \textbf{29.4} & \textbf{52.9} & \textbf{70.6} & \textbf{3.5}  & \textbf{9.0}  & \textbf{18.4} \\
w/o $\{\bm{g}_{m}\}$    & 43.0          & 61.7          & 76.0          & 21.3          & 39.6          & 58.0          & 11.5          & 25.1          & 44.2          & 2.1           & 5.6           & 13.0          \\
w/o $\bm{c}$            & 17.4          & 36.2          & 54.5          & 2.1           & 5.0           & 10.9          & 17.4          & 36.2          & 54.5          & 2.1           & 5.0           & 10.9          \\
w/o score head          & 48.0            & 68.1            & 81.6            & 21.4            & 40.4            & 59.4            & 26.5            & 50.4            & 69.0            & 3.2            & 8.7            & 17.9            \\
w/o translation         & 40.5          & 61.1          & 77.3          & 21.0          & 39.7          & 58.7          & 16.4          & 36.1          & 58.4          & 2.3           & 6.7           & 15.9          \\
w/o rot6d               & 43.7            & 65.0            & 79.7            & 21.0            & 40.1            & 58.9            & 20.4          & 43.3          & 64.0          & 2.3            & 7.5            & 16.9            \\
\bottomrule
\end{tabular}%
}
\end{table}

\paragraph{Conditioning ablation.} Removing the geometry tokens $\{\bm{g}_{m}\}$ collapses performance to the init level: without per-match evidence the model can only reproduce the prior mean. Removing the pose token $\bm{c}$ (equivalently, replacing $T_{\text{init}}$ with the identity) forces the model to estimate pose from scratch; the SuperGlue and LoFTR columns become identical since the upstream pose no longer enters the model. The model still recovers a reasonable pose from warping residuals alone, but consistently below the accuracy obtained with a real init.

\paragraph{Score-head fusion vs.\ uniform averaging.} The \emph{w/o score head} variant uniform-averages the $M{+}1$ pose hypotheses instead of fusing them by the score head's softmax weights. Performance drops: outlier hypotheses receive the same weight as well-aligned ones, whereas the score head learns to discount them via plane-warping cost (Eq.~\eqref{eq:scorecost}).

\paragraph{Pose-residual parameterization matters.} The \emph{w/o translation} variant drops the translation channel from the diffusion residual (rot-only $\bm{z}_0\!\in\!\mathbb{R}^{6}$); rotation accuracy degrades because the bilinear coupling between $R$ and $\bm{t}$ in $\Delta d$ provides a contextual signal that tightens the rotation prediction. The \emph{w/o rot6d} variant replaces the continuous $6$-D rotation representation~\citep{rot6d} with a unit quaternion in $\mathbb{R}^{4}$ and reduces rotation accuracy. Quaternions form a double cover of $\SO(3)$: $q$ and $-q$ represent the same rotation. Choosing a unique quaternion for each rotation introduces discontinuities that can produce abrupt changes in Euclidean flow-matching targets. The continuous $6$-D representation avoids this ambiguity. Appendix~\ref{app:geo-rot6d} provides the mathematical explanation.

\paragraph{Training objective.} Flow matching with a single Euler step outperforms direct regression and 100-step DDPM in rotation and translation-direction AUC on ScanNet1500 with LoFTR initialization. Appendix~\ref{app:objective-comparison} presents the quantitative results and experimental setup.

\subsection{Limitations}
\label{sec:limitations}

\textsc{PosePlaner} relies on reliable planar evidence, so scenes with limited planar structure or inaccurate plane predictions can reduce refinement gains. The plane source is interchangeable: Appendix~\ref{app:alternative-planes} demonstrates refinement with MoGe-derived planes~\citep{DBLP:conf/cvpr/WangXDX00Y25}, suggesting that improved geometry predictors could further benefit our method. For already-accurate initial poses, plane-prediction noise can outweigh the remaining opportunity for correction and cause slight regressions, as observed for \textsc{ROMA} and \textsc{MASt3R} on ScanNet1500 (Tab.~\ref{tab:twoview}). Refinement is therefore most useful when the initialization leaves substantial room for correction; it is not guaranteed to improve an already-near-optimal estimate.

\section{Conclusion}
\label{sec:conclusion}

We presented \textsc{PosePlaner}, a flow-matching pose refiner that restores correspondence--pose coupling in feed-forward $3$-D reconstruction. It combines single-step rotation refinement conditioned on matched planar primitives with geometric translation recovery for two-view inputs. Trained on two-view pairs, the same refiner improves streaming multi-view poses without retraining. Experiments demonstrate gains across correspondence-based and diffusion-based pose estimators and reduced drift in streaming reconstruction. Extending the conditioning to other geometric primitives is a promising direction for future work.

\clearpage
\bibliography{references}

@inproceedings{vggt,
  author       = {Jianyuan Wang and
                  Minghao Chen and
                  Nikita Karaev and
                  Andrea Vedaldi and
                  Christian Rupprecht and
                  David Novotn{\'{y}}},
  title        = {{VGGT:} Visual Geometry Grounded Transformer},
  booktitle    = {{IEEE/CVF} Conference on Computer Vision and Pattern Recognition,
                  {CVPR} 2025, Nashville, TN, USA, June 11-15, 2025},
  pages        = {5294--5306},
  publisher    = {Computer Vision Foundation / {IEEE}},
  year         = {2025},
  url          = {https://openaccess.thecvf.com/content/CVPR2025/html/Wang\_VGGT\_Visual\_Geometry\_Grounded\_Transformer\_CVPR\_2025\_paper.html},
  doi          = {10.1109/CVPR52734.2025.00499},
  bibsource    = {dblp computer science bibliography, https://dblp.org}
}

@inproceedings{dust3r,
  author       = {Shuzhe Wang and
                  Vincent Leroy and
                  Yohann Cabon and
                  Boris Chidlovskii and
                  J{\'{e}}r{\^{o}}me Revaud},
  title        = {DUSt3R: Geometric 3D Vision Made Easy},
  booktitle    = {{IEEE/CVF} Conference on Computer Vision and Pattern Recognition,
                  {CVPR} 2024, Seattle, WA, USA, June 16-22, 2024},
  pages        = {20697--20709},
  publisher    = {{IEEE}},
  year         = {2024},
  url          = {https://doi.org/10.1109/CVPR52733.2024.01956},
  doi          = {10.1109/CVPR52733.2024.01956},
  bibsource    = {dblp computer science bibliography, https://dblp.org}
}

@inproceedings{raydiffusion,
  author       = {Jason Y. Zhang and
                  Amy Lin and
                  Moneish Kumar and
                  Tzu{-}Hsuan Yang and
                  Deva Ramanan and
                  Shubham Tulsiani},
  title        = {Cameras as Rays: Pose Estimation via Ray Diffusion},
  booktitle    = {The Twelfth International Conference on Learning Representations,
                  {ICLR} 2024, Vienna, Austria, May 7-11, 2024},
  publisher    = {OpenReview.net},
  year         = {2024},
  url          = {https://openreview.net/forum?id=EanCFCwAjM},
  bibsource    = {dblp computer science bibliography, https://dblp.org}
}

@inproceedings{posediffusion,
  author       = {Jianyuan Wang and
                  Christian Rupprecht and
                  David Novotn{\'{y}}},
  title        = {PoseDiffusion: Solving Pose Estimation via Diffusion-aided Bundle
                  Adjustment},
  booktitle    = {{IEEE/CVF} International Conference on Computer Vision, {ICCV} 2023,
                  Paris, France, October 1-6, 2023},
  pages        = {9739--9749},
  publisher    = {{IEEE}},
  year         = {2023},
  url          = {https://doi.org/10.1109/ICCV51070.2023.00896},
  doi          = {10.1109/ICCV51070.2023.00896},
  bibsource    = {dblp computer science bibliography, https://dblp.org}
}

@article{nerf,
  author       = {Ben Mildenhall and
                  Pratul P. Srinivasan and
                  Matthew Tancik and
                  Jonathan T. Barron and
                  Ravi Ramamoorthi and
                  Ren Ng},
  title        = {NeRF: representing scenes as neural radiance fields for view synthesis},
  journal      = {Commun. {ACM}},
  volume       = {65},
  number       = {1},
  pages        = {99--106},
  year         = {2022},
  url          = {https://doi.org/10.1145/3503250},
  doi          = {10.1145/3503250},
  bibsource    = {dblp computer science bibliography, https://dblp.org}
}

@article{3dgs,
  author       = {Bernhard Kerbl and
                  Georgios Kopanas and
                  Thomas Leimk{\"{u}}hler and
                  George Drettakis},
  title        = {3D Gaussian Splatting for Real-Time Radiance Field Rendering},
  journal      = {{ACM} Trans. Graph.},
  volume       = {42},
  number       = {4},
  pages        = {139:1--139:14},
  year         = {2023},
  url          = {https://doi.org/10.1145/3592433},
  doi          = {10.1145/3592433},
  bibsource    = {dblp computer science bibliography, https://dblp.org}
}

@inproceedings{mast3r,
  author       = {Vincent Leroy and
                  Yohann Cabon and
                  J{\'{e}}r{\^{o}}me Revaud},
  editor       = {Ales Leonardis and
                  Elisa Ricci and
                  Stefan Roth and
                  Olga Russakovsky and
                  Torsten Sattler and
                  G{\"{u}}l Varol},
  title        = {Grounding Image Matching in 3D with MASt3R},
  booktitle    = {Computer Vision - {ECCV} 2024 - 18th European Conference, Milan, Italy,
                  September 29-October 4, 2024, Proceedings, Part {LXXII}},
  series       = {Lecture Notes in Computer Science},
  pages        = {71--91},
  publisher    = {Springer},
  year         = {2024},
  url          = {https://doi.org/10.1007/978-3-031-73220-1\_5},
  doi          = {10.1007/978-3-031-73220-1\_5},
  bibsource    = {dblp computer science bibliography, https://dblp.org}
}

@inproceedings{superglue,
  author       = {Paul{-}Edouard Sarlin and
                  Daniel DeTone and
                  Tomasz Malisiewicz and
                  Andrew Rabinovich},
  title        = {SuperGlue: Learning Feature Matching With Graph Neural Networks},
  booktitle    = {2020 {IEEE/CVF} Conference on Computer Vision and Pattern Recognition,
                  {CVPR} 2020, Seattle, WA, USA, June 13-19, 2020},
  pages        = {4937--4946},
  publisher    = {Computer Vision Foundation / {IEEE}},
  year         = {2020},
  url          = {https://openaccess.thecvf.com/content\_CVPR\_2020/html/Sarlin\_SuperGlue\_Learning\_Feature\_Matching\_With\_Graph\_Neural\_Networks\_CVPR\_2020\_paper.html},
  doi          = {10.1109/CVPR42600.2020.00499},
  bibsource    = {dblp computer science bibliography, https://dblp.org}
}

@inproceedings{roma,
  author       = {Johan Edstedt and
                  Qiyu Sun and
                  Georg B{\"{o}}kman and
                  M{\aa}rten Wadenb{\"{a}}ck and
                  Michael Felsberg},
  title        = {RoMa: Robust Dense Feature Matching},
  booktitle    = {{IEEE/CVF} Conference on Computer Vision and Pattern Recognition,
                  {CVPR} 2024, Seattle, WA, USA, June 16-22, 2024},
  pages        = {19790--19800},
  publisher    = {{IEEE}},
  year         = {2024},
  url          = {https://doi.org/10.1109/CVPR52733.2024.01871},
  doi          = {10.1109/CVPR52733.2024.01871},
  bibsource    = {dblp computer science bibliography, https://dblp.org}
}

@inproceedings{loftr,
  author       = {Jiaming Sun and
                  Zehong Shen and
                  Yuang Wang and
                  Hujun Bao and
                  Xiaowei Zhou},
  title        = {LoFTR: Detector-Free Local Feature Matching With Transformers},
  booktitle    = {{IEEE} Conference on Computer Vision and Pattern Recognition, {CVPR}
                  2021, virtual, June 19-25, 2021},
  pages        = {8922--8931},
  publisher    = {Computer Vision Foundation / {IEEE}},
  year         = {2021},
  url          = {https://openaccess.thecvf.com/content/CVPR2021/html/Sun\_LoFTR\_Detector-Free\_Local\_Feature\_Matching\_With\_Transformers\_CVPR\_2021\_paper.html},
  doi          = {10.1109/CVPR46437.2021.00881},
  bibsource    = {dblp computer science bibliography, https://dblp.org}
}

@inproceedings{cut3r,
  author       = {Qianqian Wang and
                  Yifei Zhang and
                  Aleksander Holynski and
                  Alexei A. Efros and
                  Angjoo Kanazawa},
  title        = {Continuous 3D Perception Model with Persistent State},
  booktitle    = {{IEEE/CVF} Conference on Computer Vision and Pattern Recognition,
                  {CVPR} 2025, Nashville, TN, USA, June 11-15, 2025},
  pages        = {10510--10522},
  publisher    = {Computer Vision Foundation / {IEEE}},
  year         = {2025},
  url          = {https://openaccess.thecvf.com/content/CVPR2025/html/Wang\_Continuous\_3D\_Perception\_Model\_with\_Persistent\_State\_CVPR\_2025\_paper.html},
  doi          = {10.1109/CVPR52734.2025.00983},
  bibsource    = {dblp computer science bibliography, https://dblp.org}
}

@article{wint3r,
  author       = {Zizun Li and
                  Jianjun Zhou and
                  Yifan Wang and
                  Haoyu Guo and
                  Wenzheng Chang and
                  Yang Zhou and
                  Haoyi Zhu and
                  Junyi Chen and
                  Chunhua Shen and
                  Tong He},
  title        = {WinT3R: Window-Based Streaming Reconstruction with Camera Token Pool},
  journal      = {CoRR},
  volume       = {abs/2509.05296},
  year         = {2025},
  url          = {https://doi.org/10.48550/arXiv.2509.05296},
  doi          = {10.48550/ARXIV.2509.05296},
  eprinttype   = {arXiv},
  eprint       = {2509.05296},
  bibsource    = {dblp computer science bibliography, https://dblp.org}
}

@article{ttt3r,
  author       = {Xingyu Chen and
                  Yue Chen and
                  Yuliang Xiu and
                  Andreas Geiger and
                  Anpei Chen},
  title        = {{TTT3R:} 3D Reconstruction as Test-Time Training},
  journal      = {CoRR},
  volume       = {abs/2509.26645},
  year         = {2025},
  url          = {https://doi.org/10.48550/arXiv.2509.26645},
  doi          = {10.48550/ARXIV.2509.26645},
  eprinttype   = {arXiv},
  eprint       = {2509.26645},
  bibsource    = {dblp computer science bibliography, https://dblp.org}
}

@inproceedings{spann3r,
  author       = {Hengyi Wang and
                  Lourdes Agapito},
  title        = {3D Reconstruction with Spatial Memory},
  booktitle    = {International Conference on 3D Vision, 3DV 2025, Singapore, March
                  25-28, 2025},
  pages        = {78--89},
  publisher    = {{IEEE}},
  year         = {2025},
  url          = {https://doi.org/10.1109/3DV66043.2025.00013},
  doi          = {10.1109/3DV66043.2025.00013},
  bibsource    = {dblp computer science bibliography, https://dblp.org}
}

@inproceedings{glomap,
  author       = {Linfei Pan and
                  D{\'{a}}niel Bar{\'{a}}th and
                  Marc Pollefeys and
                  Johannes L. Sch{\"{o}}nberger},
  editor       = {Ales Leonardis and
                  Elisa Ricci and
                  Stefan Roth and
                  Olga Russakovsky and
                  Torsten Sattler and
                  G{\"{u}}l Varol},
  title        = {Global Structure-from-Motion Revisited},
  booktitle    = {Computer Vision - {ECCV} 2024 - 18th European Conference, Milan, Italy,
                  September 29-October 4, 2024, Proceedings, Part {XL}},
  series       = {Lecture Notes in Computer Science},
  pages        = {58--77},
  publisher    = {Springer},
  year         = {2024},
  url          = {https://doi.org/10.1007/978-3-031-73661-2\_4},
  doi          = {10.1007/978-3-031-73661-2\_4},
  bibsource    = {dblp computer science bibliography, https://dblp.org}
}

@inproceedings{yao2018mvsnet,
  author       = {Yao Yao and
                  Zixin Luo and
                  Shiwei Li and
                  Tian Fang and
                  Long Quan},
  editor       = {Vittorio Ferrari and
                  Martial Hebert and
                  Cristian Sminchisescu and
                  Yair Weiss},
  title        = {MVSNet: Depth Inference for Unstructured Multi-view Stereo},
  booktitle    = {Computer Vision - {ECCV} 2018 - 15th European Conference, Munich,
                  Germany, September 8-14, 2018, Proceedings, Part {VIII}},
  series       = {Lecture Notes in Computer Science},
  pages        = {785--801},
  publisher    = {Springer},
  year         = {2018},
  url          = {https://doi.org/10.1007/978-3-030-01237-3\_47},
  doi          = {10.1007/978-3-030-01237-3\_47},
  bibsource    = {dblp computer science bibliography, https://dblp.org}
}

@article{mur2015orb,
  author       = {Raul Mur{-}Artal and
                  J. M. M. Montiel and
                  Juan D. Tard{\'{o}}s},
  title        = {{ORB-SLAM:} {A} Versatile and Accurate Monocular {SLAM} System},
  journal      = {{IEEE} Trans. Robotics},
  volume       = {31},
  number       = {5},
  pages        = {1147--1163},
  year         = {2015},
  url          = {https://doi.org/10.1109/TRO.2015.2463671},
  doi          = {10.1109/TRO.2015.2463671},
  bibsource    = {dblp computer science bibliography, https://dblp.org}
}

@inproceedings{schonberger2016structure,
  author       = {Johannes L. Sch{\"{o}}nberger and
                  Jan{-}Michael Frahm},
  title        = {Structure-from-Motion Revisited},
  booktitle    = {2016 {IEEE} Conference on Computer Vision and Pattern Recognition,
                  {CVPR} 2016, Las Vegas, NV, USA, June 27-30, 2016},
  pages        = {4104--4113},
  publisher    = {{IEEE} Computer Society},
  year         = {2016},
  url          = {https://doi.org/10.1109/CVPR.2016.445},
  doi          = {10.1109/CVPR.2016.445},
  bibsource    = {dblp computer science bibliography, https://dblp.org}
}

@article{plana3r,
  author       = {Changkun Liu and
                  Bin Tan and
                  Zeran Ke and
                  Shangzhan Zhang and
                  Jiachen Liu and
                  Ming Qian and
                  Nan Xue and
                  Yujun Shen and
                  Tristan Braud},
  title        = {{PLANA3R:} Zero-shot Metric Planar 3D Reconstruction via Feed-Forward
                  Planar Splatting},
  journal      = {CoRR},
  volume       = {abs/2510.18714},
  year         = {2025},
  url          = {https://doi.org/10.48550/arXiv.2510.18714},
  doi          = {10.48550/ARXIV.2510.18714},
  eprinttype   = {arXiv},
  eprint       = {2510.18714},
  bibsource    = {dblp computer science bibliography, https://dblp.org}
}

@inproceedings{planarsplatting,
  author       = {Bin Tan and
                  Rui Yu and
                  Yujun Shen and
                  Nan Xue},
  title        = {PlanarSplatting: Accurate Planar Surface Reconstruction in 3 Minutes},
  booktitle    = {{IEEE/CVF} Conference on Computer Vision and Pattern Recognition,
                  {CVPR} 2025, Nashville, TN, USA, June 11-15, 2025},
  pages        = {1190--1199},
  publisher    = {Computer Vision Foundation / {IEEE}},
  year         = {2025},
  url          = {https://openaccess.thecvf.com/content/CVPR2025/html/Tan\_PlanarSplatting\_Accurate\_Planar\_Surface\_Reconstruction\_in\_3\_Minutes\_CVPR\_2025\_paper.html},
  doi          = {10.1109/CVPR52734.2025.00119},
  bibsource    = {dblp computer science bibliography, https://dblp.org}
}

@article{nopesac,
  author       = {Bin Tan and
                  Nan Xue and
                  Tianfu Wu and
                  Gui{-}Song Xia},
  title        = {{NOPE-SAC:} Neural One-Plane {RANSAC} for Sparse-View Planar 3D Reconstruction},
  journal      = {{IEEE} Trans. Pattern Anal. Mach. Intell.},
  volume       = {45},
  number       = {12},
  pages        = {15233--15248},
  year         = {2023},
  url          = {https://doi.org/10.1109/TPAMI.2023.3314745},
  doi          = {10.1109/TPAMI.2023.3314745},
  bibsource    = {dblp computer science bibliography, https://dblp.org}
}

@inproceedings{rot6d,
  author       = {Yi Zhou and
                  Connelly Barnes and
                  Jingwan Lu and
                  Jimei Yang and
                  Hao Li},
  title        = {On the Continuity of Rotation Representations in Neural Networks},
  booktitle    = {{IEEE} Conference on Computer Vision and Pattern Recognition, {CVPR}
                  2019, Long Beach, CA, USA, June 16-20, 2019},
  pages        = {5745--5753},
  publisher    = {Computer Vision Foundation / {IEEE}},
  year         = {2019},
  url          = {http://openaccess.thecvf.com/content\_CVPR\_2019/html/Zhou\_On\_the\_Continuity\_of\_Rotation\_Representations\_in\_Neural\_Networks\_CVPR\_2019\_paper.html},
  doi          = {10.1109/CVPR.2019.00589},
  bibsource    = {dblp computer science bibliography, https://dblp.org}
}

@inproceedings{flowmatching,
  author       = {Yaron Lipman and
                  Ricky T. Q. Chen and
                  Heli Ben{-}Hamu and
                  Maximilian Nickel and
                  Matthew Le},
  title        = {Flow Matching for Generative Modeling},
  booktitle    = {The Eleventh International Conference on Learning Representations,
                  {ICLR} 2023, Kigali, Rwanda, May 1-5, 2023},
  publisher    = {OpenReview.net},
  year         = {2023},
  url          = {https://openreview.net/forum?id=PqvMRDCJT9t},
  bibsource    = {dblp computer science bibliography, https://dblp.org}
}

@inproceedings{transformer,
  author       = {Ashish Vaswani and
                  Noam Shazeer and
                  Niki Parmar and
                  Jakob Uszkoreit and
                  Llion Jones and
                  Aidan N. Gomez and
                  Lukasz Kaiser and
                  Illia Polosukhin},
  editor       = {Isabelle Guyon and
                  Ulrike von Luxburg and
                  Samy Bengio and
                  Hanna M. Wallach and
                  Rob Fergus and
                  S. V. N. Vishwanathan and
                  Roman Garnett},
  title        = {Attention is All you Need},
  booktitle    = {Advances in Neural Information Processing Systems 30: Annual Conference
                  on Neural Information Processing Systems 2017, December 4-9, 2017,
                  Long Beach, CA, {USA}},
  pages        = {5998--6008},
  year         = {2017},
  url          = {https://proceedings.neurips.cc/paper/2017/hash/3f5ee243547dee91fbd053c1c4a845aa-Abstract.html},
  bibsource    = {dblp computer science bibliography, https://dblp.org}
}

@inproceedings{ddpm,
  author       = {Jonathan Ho and
                  Ajay Jain and
                  Pieter Abbeel},
  editor       = {Hugo Larochelle and
                  Marc'Aurelio Ranzato and
                  Raia Hadsell and
                  Maria{-}Florina Balcan and
                  Hsuan{-}Tien Lin},
  title        = {Denoising Diffusion Probabilistic Models},
  booktitle    = {Advances in Neural Information Processing Systems 33: Annual Conference
                  on Neural Information Processing Systems 2020, NeurIPS 2020, December
                  6-12, 2020, virtual},
  year         = {2020},
  url          = {https://proceedings.neurips.cc/paper/2020/hash/4c5bcfec8584af0d967f1ab10179ca4b-Abstract.html},
  bibsource    = {dblp computer science bibliography, https://dblp.org}
}

@inproceedings{scannet,
  author       = {Angela Dai and
                  Angel X. Chang and
                  Manolis Savva and
                  Maciej Halber and
                  Thomas A. Funkhouser and
                  Matthias Nie{\ss}ner},
  title        = {ScanNet: Richly-Annotated 3D Reconstructions of Indoor Scenes},
  booktitle    = {2017 {IEEE} Conference on Computer Vision and Pattern Recognition,
                  {CVPR} 2017, Honolulu, HI, USA, July 21-26, 2017},
  pages        = {2432--2443},
  publisher    = {{IEEE} Computer Society},
  year         = {2017},
  url          = {https://doi.org/10.1109/CVPR.2017.261},
  doi          = {10.1109/CVPR.2017.261},
  bibsource    = {dblp computer science bibliography, https://dblp.org}
}

@inproceedings{scannetpp,
  author       = {Chandan Yeshwanth and
                  Yueh{-}Cheng Liu and
                  Matthias Nie{\ss}ner and
                  Angela Dai},
  title        = {ScanNet++: {A} High-Fidelity Dataset of 3D Indoor Scenes},
  booktitle    = {{IEEE/CVF} International Conference on Computer Vision, {ICCV} 2023,
                  Paris, France, October 1-6, 2023},
  pages        = {12--22},
  publisher    = {{IEEE}},
  year         = {2023},
  url          = {https://doi.org/10.1109/ICCV51070.2023.00008},
  doi          = {10.1109/ICCV51070.2023.00008},
  bibsource    = {dblp computer science bibliography, https://dblp.org}
}

@inproceedings{arkitscenes,
  author       = {Afshin Dehghan and
                  Gilad Baruch and
                  Zhuoyuan Chen and
                  Yuri Feigin and
                  Peter Fu and
                  Thomas Gebauer and
                  Daniel Kurz and
                  Tal Dimry and
                  Brandon Joffe and
                  Arik Schwartz and
                  Elad Shulman},
  editor       = {Joaquin Vanschoren and
                  Sai{-}Kit Yeung},
  title        = {ARKitScenes: {A} Diverse Real-World Dataset For 3D Indoor Scene Understanding
                  Using Mobile {RGB-D} Data},
  booktitle    = {Proceedings of the Neural Information Processing Systems Track on
                  Datasets and Benchmarks 1, NeurIPS Datasets and Benchmarks 2021, December
                  2021, virtual},
  year         = {2021},
  url          = {https://datasets-benchmarks-proceedings.neurips.cc/paper/2021/hash/66f041e16a60928b05a7e228a89c3799-Abstract-round1.html},
  bibsource    = {dblp computer science bibliography, https://dblp.org}
}

@inproceedings{habitat,
  author       = {Manolis Savva and
                  Jitendra Malik and
                  Devi Parikh and
                  Dhruv Batra and
                  Abhishek Kadian and
                  Oleksandr Maksymets and
                  Yili Zhao and
                  Erik Wijmans and
                  Bhavana Jain and
                  Julian Straub and
                  Jia Liu and
                  Vladlen Koltun},
  title        = {Habitat: {A} Platform for Embodied {AI} Research},
  booktitle    = {2019 {IEEE/CVF} International Conference on Computer Vision, {ICCV}
                  2019, Seoul, Korea (South), October 27 - November 2, 2019},
  pages        = {9338--9346},
  publisher    = {{IEEE}},
  year         = {2019},
  url          = {https://doi.org/10.1109/ICCV.2019.00943},
  doi          = {10.1109/ICCV.2019.00943},
  bibsource    = {dblp computer science bibliography, https://dblp.org}
}

@inproceedings{7scenes,
  author       = {Jamie Shotton and
                  Ben Glocker and
                  Christopher Zach and
                  Shahram Izadi and
                  Antonio Criminisi and
                  Andrew W. Fitzgibbon},
  title        = {Scene Coordinate Regression Forests for Camera Relocalization in {RGB-D}
                  Images},
  booktitle    = {2013 {IEEE} Conference on Computer Vision and Pattern Recognition,
                  Portland, OR, USA, June 23-28, 2013},
  pages        = {2930--2937},
  publisher    = {{IEEE} Computer Society},
  year         = {2013},
  url          = {https://doi.org/10.1109/CVPR.2013.377},
  doi          = {10.1109/CVPR.2013.377},
  bibsource    = {dblp computer science bibliography, https://dblp.org}
}

@article{ransac,
  author       = {Martin A. Fischler and
                  Robert C. Bolles},
  title        = {Random Sample Consensus: {A} Paradigm for Model Fitting with Applications
                  to Image Analysis and Automated Cartography},
  journal      = {Commun. {ACM}},
  volume       = {24},
  number       = {6},
  pages        = {381--395},
  year         = {1981},
  url          = {https://doi.org/10.1145/358669.358692},
  doi          = {10.1145/358669.358692},
  bibsource    = {dblp computer science bibliography, https://dblp.org}
}

@inproceedings{dit,
  author       = {William Peebles and
                  Saining Xie},
  title        = {Scalable Diffusion Models with Transformers},
  booktitle    = {{IEEE/CVF} International Conference on Computer Vision, {ICCV} 2023,
                  Paris, France, October 1-6, 2023},
  pages        = {4172--4182},
  publisher    = {{IEEE}},
  year         = {2023},
  url          = {https://doi.org/10.1109/ICCV51070.2023.00387},
  doi          = {10.1109/ICCV51070.2023.00387},
  bibsource    = {dblp computer science bibliography, https://dblp.org}
}

@inproceedings{liu2022rectified,
  author       = {Xingchao Liu and
                  Chengyue Gong and
                  Qiang Liu},
  title        = {Flow Straight and Fast: Learning to Generate and Transfer Data with
                  Rectified Flow},
  booktitle    = {The Eleventh International Conference on Learning Representations,
                  {ICLR} 2023, Kigali, Rwanda, May 1-5, 2023},
  publisher    = {OpenReview.net},
  year         = {2023},
  url          = {https://openreview.net/forum?id=XVjTT1nw5z},
  bibsource    = {dblp computer science bibliography, https://dblp.org}
}

@inproceedings{vit,
  author       = {Alexey Dosovitskiy and
                  Lucas Beyer and
                  Alexander Kolesnikov and
                  Dirk Weissenborn and
                  Xiaohua Zhai and
                  Thomas Unterthiner and
                  Mostafa Dehghani and
                  Matthias Minderer and
                  Georg Heigold and
                  Sylvain Gelly and
                  Jakob Uszkoreit and
                  Neil Houlsby},
  title        = {An Image is Worth 16x16 Words: Transformers for Image Recognition
                  at Scale},
  booktitle    = {9th International Conference on Learning Representations, {ICLR} 2021,
                  Virtual Event, Austria, May 3-7, 2021},
  publisher    = {OpenReview.net},
  year         = {2021},
  url          = {https://openreview.net/forum?id=YicbFdNTTy},
  bibsource    = {dblp computer science bibliography, https://dblp.org}
}

@article{nister5pt,
  author       = {David Nist{\'{e}}r},
  title        = {An Efficient Solution to the Five-Point Relative Pose Problem},
  journal      = {{IEEE} Trans. Pattern Anal. Mach. Intell.},
  volume       = {26},
  number       = {6},
  pages        = {756--777},
  year         = {2004},
  url          = {https://doi.org/10.1109/TPAMI.2004.17},
  doi          = {10.1109/TPAMI.2004.17},
  bibsource    = {dblp computer science bibliography, https://dblp.org}
}

@article{streamvggt,
  author       = {Dong Zhuo and
                  Wenzhao Zheng and
                  Jiahe Guo and
                  Yuqi Wu and
                  Jie Zhou and
                  Jiwen Lu},
  title        = {Streaming 4D Visual Geometry Transformer},
  journal      = {CoRR},
  volume       = {abs/2507.11539},
  year         = {2025},
  url          = {https://doi.org/10.48550/arXiv.2507.11539},
  doi          = {10.48550/ARXIV.2507.11539},
  eprinttype   = {arXiv},
  eprint       = {2507.11539},
  bibsource    = {dblp computer science bibliography, https://dblp.org}
}

@article{infinitevggt,
  author       = {Shuai Yuan and
                  Yantai Yang and
                  Xiaotian Yang and
                  Xupeng Zhang and
                  Zhonghao Zhao and
                  Lingming Zhang and
                  Zhipeng Zhang},
  title        = {InfiniteVGGT: Visual Geometry Grounded Transformer for Endless Streams},
  journal      = {CoRR},
  volume       = {abs/2601.02281},
  year         = {2026},
  url          = {https://doi.org/10.48550/arXiv.2601.02281},
  doi          = {10.48550/ARXIV.2601.02281},
  eprinttype   = {arXiv},
  eprint       = {2601.02281},
  bibsource    = {dblp computer science bibliography, https://dblp.org}
}

@article{stream3r,
  author       = {Yushi Lan and
                  Yihang Luo and
                  Fangzhou Hong and
                  Shangchen Zhou and
                  Honghua Chen and
                  Zhaoyang Lyu and
                  Shuai Yang and
                  Bo Dai and
                  Chen Change Loy and
                  Xingang Pan},
  title        = {STream3R: Scalable Sequential 3D Reconstruction with Causal Transformer},
  journal      = {CoRR},
  volume       = {abs/2508.10893},
  year         = {2025},
  url          = {https://doi.org/10.48550/arXiv.2508.10893},
  doi          = {10.48550/ARXIV.2508.10893},
  eprinttype   = {arXiv},
  eprint       = {2508.10893},
  bibsource    = {dblp computer science bibliography, https://dblp.org}
}

@article{gca,
  title={Geometric Context Transformer for Streaming 3D Reconstruction},
  author={Chen, Lin-Zhuo and Gao, Jian and Chen, Yihang and Cheng, Ka Leong and Sun, Yipengjing and Hu, Liangxiao and Xue, Nan and Zhu, Xing and Shen, Yujun and Yao, Yao and others},
  journal={arXiv preprint arXiv:2604.14141},
  year={2026}
}

@article{planarrecon,
  author       = {Yiming Xie and
                  Matheus Gadelha and
                  Fengting Yang and
                  Xiaowei Zhou and
                  Huaizu Jiang},
  title        = {PlanarRecon: Real-time 3D Plane Detection and Reconstruction from
                  Posed Monocular Videos},
  journal      = {CoRR},
  volume       = {abs/2206.07710},
  year         = {2022},
  url          = {https://doi.org/10.48550/arXiv.2206.07710},
  doi          = {10.48550/ARXIV.2206.07710},
  eprinttype   = {arXiv},
  eprint       = {2206.07710},
  bibsource    = {dblp computer science bibliography, https://dblp.org}
}

@inproceedings{yu2019single,
  author       = {Zehao Yu and
                  Jia Zheng and
                  Dongze Lian and
                  Zihan Zhou and
                  Shenghua Gao},
  title        = {Single-Image Piece-Wise Planar 3D Reconstruction via Associative Embedding},
  booktitle    = {{IEEE} Conference on Computer Vision and Pattern Recognition, {CVPR}
                  2019, Long Beach, CA, USA, June 16-20, 2019},
  pages        = {1029--1037},
  publisher    = {Computer Vision Foundation / {IEEE}},
  year         = {2019},
  url          = {http://openaccess.thecvf.com/content\_CVPR\_2019/html/Yu\_Single-Image\_Piece-Wise\_Planar\_3D\_Reconstruction\_via\_Associative\_Embedding\_CVPR\_2019\_paper.html},
  doi          = {10.1109/CVPR.2019.00112},
  bibsource    = {dblp computer science bibliography, https://dblp.org}
}

@inproceedings{planercnn,
  author       = {Chen Liu and
                  Kihwan Kim and
                  Jinwei Gu and
                  Yasutaka Furukawa and
                  Jan Kautz},
  title        = {PlaneRCNN: 3D Plane Detection and Reconstruction From a Single Image},
  booktitle    = {{IEEE} Conference on Computer Vision and Pattern Recognition, {CVPR}
                  2019, Long Beach, CA, USA, June 16-20, 2019},
  pages        = {4450--4459},
  publisher    = {Computer Vision Foundation / {IEEE}},
  year         = {2019},
  url          = {http://openaccess.thecvf.com/content\_CVPR\_2019/html/Liu\_PlaneRCNN\_3D\_Plane\_Detection\_and\_Reconstruction\_From\_a\_Single\_Image\_CVPR\_2019\_paper.html},
  doi          = {10.1109/CVPR.2019.00458},
  bibsource    = {dblp computer science bibliography, https://dblp.org}
}

@inproceedings{planerectr,
  author       = {Jingjia Shi and
                  Shuaifeng Zhi and
                  Kai Xu},
  title        = {PlaneRecTR: Unified Query Learning for 3D Plane Recovery from a Single
                  View},
  booktitle    = {{IEEE/CVF} International Conference on Computer Vision, {ICCV} 2023,
                  Paris, France, October 1-6, 2023},
  pages        = {9343--9352},
  publisher    = {{IEEE}},
  year         = {2023},
  url          = {https://doi.org/10.1109/ICCV51070.2023.00860},
  doi          = {10.1109/ICCV51070.2023.00860},
  bibsource    = {dblp computer science bibliography, https://dblp.org}
}

@inproceedings{liu2025towards,
  author       = {Jiachen Liu and
                  Rui Yu and
                  Sili Chen and
                  Sharon X. Huang and
                  Hengkai Guo},
  title        = {Towards In-the-wild 3D Plane Reconstruction from a Single Image},
  booktitle    = {{IEEE/CVF} Conference on Computer Vision and Pattern Recognition,
                  {CVPR} 2025, Nashville, TN, USA, June 11-15, 2025},
  pages        = {27027--27037},
  publisher    = {Computer Vision Foundation / {IEEE}},
  year         = {2025},
  url          = {https://openaccess.thecvf.com/content/CVPR2025/html/Liu\_Towards\_In-the-wild\_3D\_Plane\_Reconstruction\_from\_a\_Single\_Image\_CVPR\_2025\_paper.html},
  doi          = {10.1109/CVPR52734.2025.02517},
  bibsource    = {dblp computer science bibliography, https://dblp.org}
}

@inproceedings{zhao2024monoplane,
  author       = {Wang Zhao and
                  Jiachen Liu and
                  Sheng Zhang and
                  Yishu Li and
                  Sili Chen and
                  Sharon X. Huang and
                  Yong{-}Jin Liu and
                  Hengkai Guo},
  title        = {MonoPlane: Exploiting Monocular Geometric Cues for Generalizable 3D
                  Plane Reconstruction},
  booktitle    = {{IEEE/RSJ} International Conference on Intelligent Robots and Systems,
                  {IROS} 2024, Abu Dhabi, United Arab Emirates, October 14-18, 2024},
  pages        = {8481--8488},
  publisher    = {{IEEE}},
  year         = {2024},
  url          = {https://doi.org/10.1109/IROS58592.2024.10802672},
  doi          = {10.1109/IROS58592.2024.10802672},
  bibsource    = {dblp computer science bibliography, https://dblp.org}
}

@inproceedings{planenet,
  author       = {Chen Liu and
                  Jimei Yang and
                  Duygu Ceylan and
                  Ersin Yumer and
                  Yasutaka Furukawa},
  title        = {PlaneNet: Piece-Wise Planar Reconstruction From a Single {RGB} Image},
  booktitle    = {2018 {IEEE} Conference on Computer Vision and Pattern Recognition,
                  {CVPR} 2018, Salt Lake City, UT, USA, June 18-22, 2018},
  pages        = {2579--2588},
  publisher    = {Computer Vision Foundation / {IEEE} Computer Society},
  year         = {2018},
  url          = {http://openaccess.thecvf.com/content\_cvpr\_2018/html/Liu\_PlaneNet\_Piece-Wise\_Planar\_CVPR\_2018\_paper.html},
  doi          = {10.1109/CVPR.2018.00273},
  bibsource    = {dblp computer science bibliography, https://dblp.org}
}

@inproceedings{chang2017matterport3dlearningrgbddata,
  author       = {Angel X. Chang and
                  Angela Dai and
                  Thomas A. Funkhouser and
                  Maciej Halber and
                  Matthias Nie{\ss}ner and
                  Manolis Savva and
                  Shuran Song and
                  Andy Zeng and
                  Yinda Zhang},
  title        = {Matterport3D: Learning from {RGB-D} Data in Indoor Environments},
  booktitle    = {2017 International Conference on 3D Vision, 3DV 2017, Qingdao, China,
                  October 10-12, 2017},
  pages        = {667--676},
  publisher    = {{IEEE} Computer Society},
  year         = {2017},
  url          = {https://doi.org/10.1109/3DV.2017.00081},
  doi          = {10.1109/3DV.2017.00081},
  bibsource    = {dblp computer science bibliography, https://dblp.org}
}

@article{metric3d,
  author       = {Mu Hu and
                  Wei Yin and
                  Chi Zhang and
                  Zhipeng Cai and
                  Xiaoxiao Long and
                  Hao Chen and
                  Kaixuan Wang and
                  Gang Yu and
                  Chunhua Shen and
                  Shaojie Shen},
  title        = {Metric3D v2: {A} Versatile Monocular Geometric Foundation Model for
                  Zero-Shot Metric Depth and Surface Normal Estimation},
  journal      = {{IEEE} Trans. Pattern Anal. Mach. Intell.},
  volume       = {46},
  number       = {12},
  pages        = {10579--10596},
  year         = {2024},
  url          = {https://doi.org/10.1109/TPAMI.2024.3444912},
  doi          = {10.1109/TPAMI.2024.3444912},
  bibsource    = {dblp computer science bibliography, https://dblp.org}
}

@inproceedings{dsac,
  author       = {Eric Brachmann and
                  Alexander Krull and
                  Sebastian Nowozin and
                  Jamie Shotton and
                  Frank Michel and
                  Stefan Gumhold and
                  Carsten Rother},
  title        = {{DSAC} - Differentiable {RANSAC} for Camera Localization},
  booktitle    = {2017 {IEEE} Conference on Computer Vision and Pattern Recognition,
                  {CVPR} 2017, Honolulu, HI, USA, July 21-26, 2017},
  pages        = {2492--2500},
  publisher    = {{IEEE} Computer Society},
  year         = {2017},
  url          = {https://doi.org/10.1109/CVPR.2017.267},
  doi          = {10.1109/CVPR.2017.267},
  bibsource    = {dblp computer science bibliography, https://dblp.org}
}

@inproceedings{dsacpp,
  author       = {Eric Brachmann and
                  Carsten Rother},
  title        = {Learning Less Is More - 6D Camera Localization via 3D Surface Regression},
  booktitle    = {2018 {IEEE} Conference on Computer Vision and Pattern Recognition,
                  {CVPR} 2018, Salt Lake City, UT, USA, June 18-22, 2018},
  pages        = {4654--4662},
  publisher    = {Computer Vision Foundation / {IEEE} Computer Society},
  year         = {2018},
  url          = {http://openaccess.thecvf.com/content\_cvpr\_2018/html/Brachmann\_Learning\_Less\_Is\_CVPR\_2018\_paper.html},
  doi          = {10.1109/CVPR.2018.00489},
  bibsource    = {dblp computer science bibliography, https://dblp.org}
}

@article{dsacstar,
  author       = {Eric Brachmann and
                  Carsten Rother},
  title        = {Visual Camera Re-Localization From {RGB} and {RGB-D} Images Using
                  {DSAC}},
  journal      = {{IEEE} Trans. Pattern Anal. Mach. Intell.},
  volume       = {44},
  number       = {9},
  pages        = {5847--5865},
  year         = {2022},
  url          = {https://doi.org/10.1109/TPAMI.2021.3070754},
  doi          = {10.1109/TPAMI.2021.3070754},
  bibsource    = {dblp computer science bibliography, https://dblp.org}
}

@inproceedings{sparseplane,
  author       = {Linyi Jin and
                  Shengyi Qian and
                  Andrew Owens and
                  David F. Fouhey},
  title        = {Planar Surface Reconstruction from Sparse Views},
  booktitle    = {2021 {IEEE/CVF} International Conference on Computer Vision, {ICCV}
                  2021, Montreal, QC, Canada, October 10-17, 2021},
  pages        = {12971--12980},
  publisher    = {{IEEE}},
  year         = {2021},
  url          = {https://doi.org/10.1109/ICCV48922.2021.01275},
  doi          = {10.1109/ICCV48922.2021.01275},
  bibsource    = {dblp computer science bibliography, https://dblp.org}
}

@inproceedings{loshchilov2017decoupled,
  author       = {Ilya Loshchilov and
                  Frank Hutter},
  title        = {Decoupled Weight Decay Regularization},
  booktitle    = {7th International Conference on Learning Representations, {ICLR} 2019,
                  New Orleans, LA, USA, May 6-9, 2019},
  publisher    = {OpenReview.net},
  year         = {2019},
  url          = {https://openreview.net/forum?id=Bkg6RiCqY7},
  bibsource    = {dblp computer science bibliography, https://dblp.org}
}

@inproceedings{diffusionsfm,
  author       = {Qitao Zhao and
                  Amy Lin and
                  Jeff Tan and
                  Jason Y. Zhang and
                  Deva Ramanan and
                  Shubham Tulsiani},
  title        = {DiffusionSfM: Predicting Structure and Motion via Ray Origin and Endpoint
                  Diffusion},
  booktitle    = {{IEEE/CVF} Conference on Computer Vision and Pattern Recognition,
                  {CVPR} 2025, Nashville, TN, USA, June 11-15, 2025},
  pages        = {6317--6326},
  publisher    = {Computer Vision Foundation / {IEEE}},
  year         = {2025},
  url          = {https://openaccess.thecvf.com/content/CVPR2025/html/Zhao\_DiffusionSfM\_Predicting\_Structure\_and\_Motion\_via\_Ray\_Origin\_and\_Endpoint\_CVPR\_2025\_paper.html},
  doi          = {10.1109/CVPR52734.2025.00592},
  bibsource    = {dblp computer science bibliography, https://dblp.org}
}

@article{rotaveraging,
  author       = {Richard I. Hartley and
                  Jochen Trumpf and
                  Yuchao Dai and
                  Hongdong Li},
  title        = {Rotation Averaging},
  journal      = {Int. J. Comput. Vis.},
  volume       = {103},
  number       = {3},
  pages        = {267--305},
  year         = {2013},
  url          = {https://doi.org/10.1007/s11263-012-0601-0},
  doi          = {10.1007/S11263-012-0601-0},
  bibsource    = {dblp computer science bibliography, https://dblp.org}
}

@inproceedings{rotaveraging_iccv,
  author       = {Avishek Chatterjee and
                  Venu Madhav Govindu},
  title        = {Efficient and Robust Large-Scale Rotation Averaging},
  booktitle    = {{IEEE} International Conference on Computer Vision, {ICCV} 2013, Sydney,
                  Australia, December 1-8, 2013},
  pages        = {521--528},
  publisher    = {{IEEE} Computer Society},
  year         = {2013},
  url          = {https://doi.org/10.1109/ICCV.2013.70},
  doi          = {10.1109/ICCV.2013.70},
  bibsource    = {dblp computer science bibliography, https://dblp.org}
}

@article{DBLP:journals/ijrr/TaoMZWFF26,
  author       = {Yifu Tao and
                  Miguel {\'{A}}ngel Mu{\~{n}}oz{-}Ba{\~{n}}{\'{o}}n and
                  Lintong Zhang and
                  Jiahao Wang and
                  Lanke Frank Tarimo Fu and
                  Maurice Fallon},
  title        = {The Oxford Spires Dataset: Benchmarking large-scale LiDAR-visual localisation,
                  reconstruction and radiance field methods},
  journal      = {Int. J. Robotics Res.},
  volume       = {45},
  number       = {6},
  pages        = {839--857},
  year         = {2026},
  url          = {https://doi.org/10.1177/02783649251369905},
  doi          = {10.1177/02783649251369905},
  bibsource    = {dblp computer science bibliography, https://dblp.org}
}

@inproceedings{DBLP:conf/iccv/0002JXX0L023,
  author       = {Yixuan Li and
                  Lihan Jiang and
                  Linning Xu and
                  Yuanbo Xiangli and
                  Zhenzhi Wang and
                  Dahua Lin and
                  Bo Dai},
  title        = {MatrixCity: {A} Large-scale City Dataset for City-scale Neural Rendering
                  and Beyond},
  booktitle    = {{IEEE/CVF} International Conference on Computer Vision, {ICCV} 2023,
                  Paris, France, October 1-6, 2023},
  pages        = {3182--3192},
  publisher    = {{IEEE}},
  year         = {2023},
  url          = {https://doi.org/10.1109/ICCV51070.2023.00297},
  doi          = {10.1109/ICCV51070.2023.00297},
  bibsource    = {dblp computer science bibliography, https://dblp.org}
}

@inproceedings{DBLP:conf/cvpr/YanZRLD22,
  author       = {Qi Yan and
                  Jianhao Zheng and
                  Simon Reding and
                  Shanci Li and
                  Iordan Doytchinov},
  title        = {CrossLoc: Scalable Aerial Localization Assisted by Multimodal Synthetic
                  Data},
  booktitle    = {{IEEE/CVF} Conference on Computer Vision and Pattern Recognition,
                  {CVPR} 2022, New Orleans, LA, USA, June 18-24, 2022},
  pages        = {17337--17347},
  publisher    = {{IEEE}},
  year         = {2022},
  url          = {https://doi.org/10.1109/CVPR52688.2022.01684},
  doi          = {10.1109/CVPR52688.2022.01684},
  bibsource    = {dblp computer science bibliography, https://dblp.org}
}

@inproceedings{DBLP:conf/cvpr/WangXDX00Y25,
  author       = {Ruicheng Wang and
                  Sicheng Xu and
                  Cassie Dai and
                  Jianfeng Xiang and
                  Yu Deng and
                  Xin Tong and
                  Jiaolong Yang},
  title        = {MoGe: Unlocking Accurate Monocular Geometry Estimation for Open-Domain
                  Images with Optimal Training Supervision},
  booktitle    = {{IEEE/CVF} Conference on Computer Vision and Pattern Recognition,
                  {CVPR} 2025, Nashville, TN, USA, June 11-15, 2025},
  pages        = {5261--5271},
  publisher    = {Computer Vision Foundation / {IEEE}},
  year         = {2025},
  url          = {https://openaccess.thecvf.com/content/CVPR2025/html/Wang\_MoGe\_Unlocking\_Accurate\_Monocular\_Geometry\_Estimation\_for\_Open-Domain\_Images\_with\_CVPR\_2025\_paper.html},
  doi          = {10.1109/CVPR52734.2025.00496},
  bibsource    = {dblp computer science bibliography, https://dblp.org}
}
\bibliographystyle{icml2025}

\clearpage
\appendix
\startcontents[appendix]
\section*{Appendix Contents}
\printcontents[appendix]{}{1}{\setcounter{tocdepth}{2}}
\clearpage
\section*{Technical Appendices and Supplementary Material}

\section{Geometric and Algorithmic Foundations}
\label{app:geometry}

This section formalizes three pieces of background that the main text builds on: the plane-warping residuals used as conditioning (Sec.~\ref{app:geo-residual}), the continuous $6$-D rotation representation carried by the pose latent (Sec.~\ref{app:geo-rot6d}), and the flow-matching parameterization under which a single Euler step at inference is sufficient (Sec.~\ref{app:denoising}).

\subsection{Plane-Warping Residuals as Conditioning Signal}
\label{app:geo-residual}

\paragraph{Vanishing at the true pose; monotonic away from it.}
Let $\bm{e}_{n}(R)\!:=\!\bm{n}_{1}-R\bm{n}_{2}$ and $e_{d}(R,\bm{t})\!:=\!d_{1}-d_{2}+(R\bm{n}_{2})^{\!\top}\bm{t}$ denote the (vector / signed) raw residuals whose norm and absolute value are the $\Delta\bm{n}$ and $\Delta d$ of Eq.~\eqref{eq:residual-prop}. By construction, $\bm{e}_{n}(R_{\text{gt}})=\bm{0}$ and $e_{d}(R_{\text{gt}},\bm{t}_{\text{gt}})=0$ identically. Under a joint first-order perturbation $R = \exp([\delta\bm{r}]_{\times})R_{\text{gt}}$ and $\bm{t} = \bm{t}_{\text{gt}} + \delta\bm{t}$, these raw residuals expand to
\begin{equation}
\bm{e}_{n}(R) \;\approx\; -[\delta\bm{r}]_{\times}\,R_{\text{gt}}\bm{n}_{2}\,,
\label{eq:delta-n-expansion}
\end{equation}
\begin{equation}
e_{d}(R,\bm{t}) \;\approx\; \big(R_{\text{gt}}\bm{n}_{2}\big)^{\!\top}\delta\bm{t} \;+\; \big([\delta\bm{r}]_{\times}R_{\text{gt}}\bm{n}_{2}\big)^{\!\top}\bm{t}_{\text{gt}}\,.
\label{eq:delta-d-expansion}
\end{equation}
$\Delta\bm{n}=\|\bm{e}_{n}\|_{2}$ is therefore linear in the rotation error along the angle between $\delta\bm{r}$ and $R_{\text{gt}}\bm{n}_{2}$; $\Delta d=|e_{d}|$ contains a translation-only probe along the direction $R_{\text{gt}}\bm{n}_{2}$ \emph{plus} a rotation-coupled term whose strength scales with $\|\bm{t}_{\text{gt}}\|$. A set of $M$ correspondences with diverse normals therefore exposes the full $\delta\bm{r}$/$\delta\bm{t}$ error along $M$ different directions, effectively a dense per-direction error map.

\paragraph{Estimator-agnostic by construction.}
Eqs.~\eqref{eq:delta-n-expansion}--\eqref{eq:delta-d-expansion} read the residuals off paired plane primitives and a candidate pose alone, with no dependency on how that pose was produced. The same conditioning signal is therefore presented to the refiner regardless of upstream---feed-forward network, matcher-plus-solver, or identity initialization.

\begin{figure}[!htbp]
\centering
\vspace{-4pt}
\begin{minipage}[c]{0.025\linewidth}
  \centering
  \raisebox{15pt}{\rotatebox{90}{\footnotesize matched plane pairs}}
\end{minipage}%
\begin{minipage}[c]{0.46\linewidth}
  \centering
  \begin{minipage}[t]{\linewidth}
    \centering
    \includegraphics[width=\linewidth]{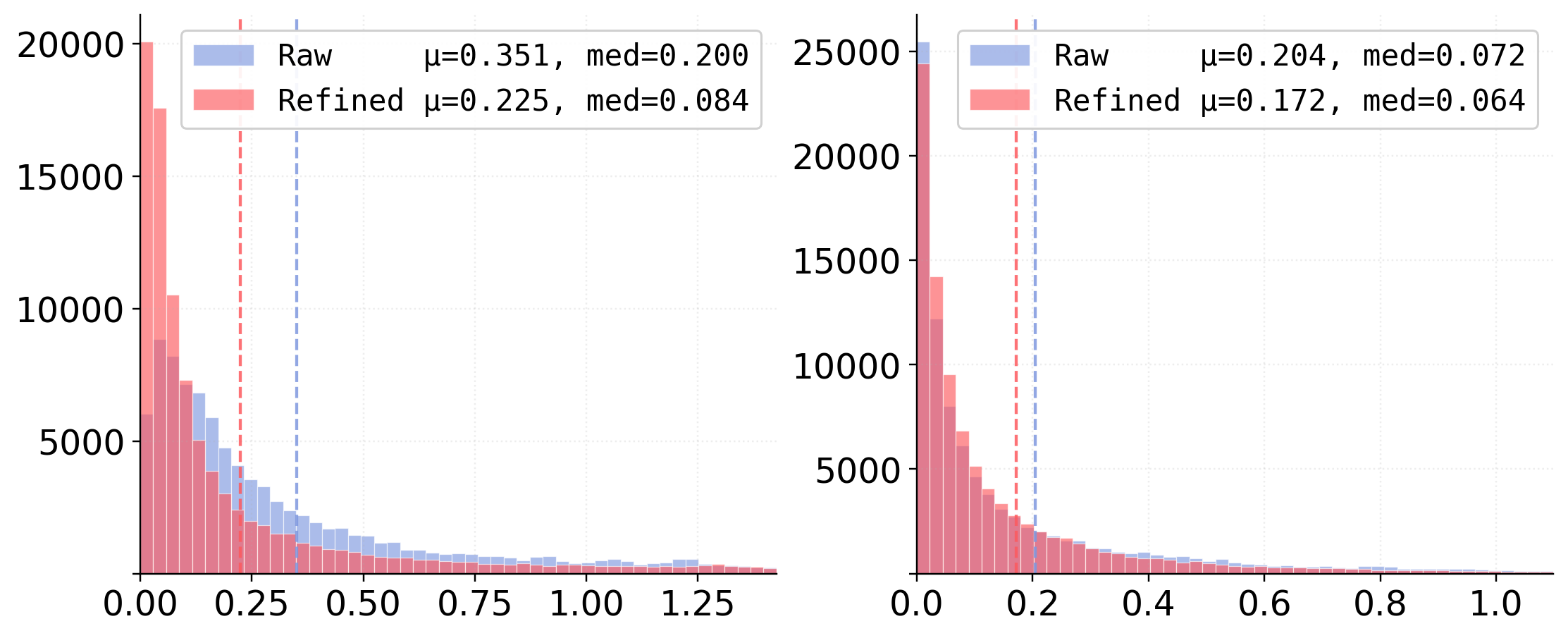}\\[-1pt]
    {\small (a) ScanNet1500, upstream = \textsc{LoFTR}}
  \end{minipage}
\end{minipage}\hfill
\begin{minipage}[c]{0.025\linewidth}
  \centering
  \raisebox{15pt}{\rotatebox{90}{\footnotesize matched plane pairs}}
\end{minipage}%
\begin{minipage}[c]{0.46\linewidth}
  \centering
  \begin{minipage}[t]{\linewidth}
    \centering
    \includegraphics[width=\linewidth]{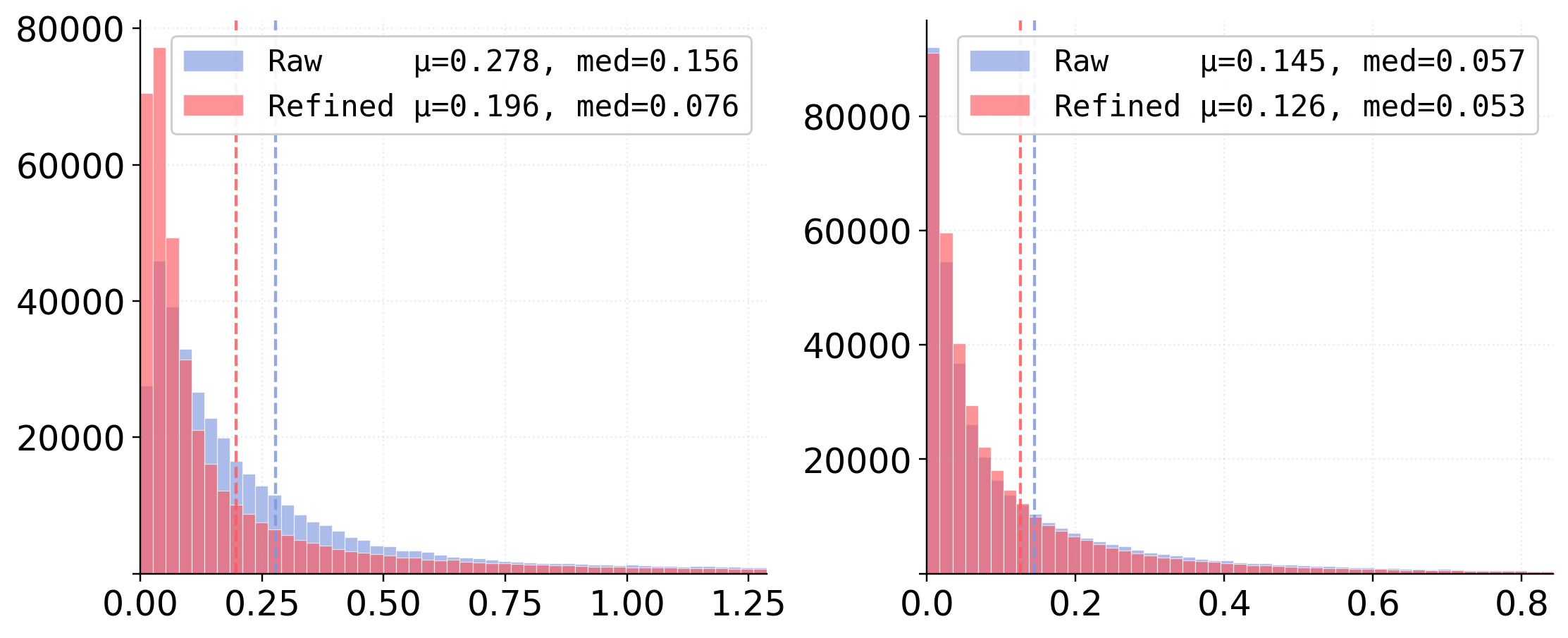}\\[-1pt]
    {\small (b) 7Scenes, upstream = \textsc{LoFTR}}
  \end{minipage}
\end{minipage}
\caption{\textbf{Refinement collapses the plane-warping residuals.} (a) ScanNet1500 and (b) 7Scenes, \textsc{LoFTR} upstream. Each sub-figure: $\|\Delta\bm{n}\|_{2}$ (left) and $|\Delta d|$ (right, scene-normalized) over all matched plane pairs; \textcolor{blue}{\textbf{blue}} = raw, \textcolor{red}{\textbf{red}} = after \textsc{PosePlaner}; dashed lines mark means.}
\label{fig:warp-residuals}
\vspace{-6pt}
\end{figure}

\paragraph{Empirical verification: the refiner implicitly learns the warp.}
In contrast to \textsc{NopeSAC}~\citep{nopesac}, which explicitly warps view-1's plane parameters by the initial pose to feed the network a pre-aligned representation, \textsc{PosePlaner} provides only the raw per-view primitives $(\bm{n}^1_m, \bar d^1_m, \bm{n}^2_m, \bar d^2_m)$ in each camera's own local frame; the warping residuals $(\Delta\bm{n},\Delta d)$ enter only as detached geometric weights $\omega_m$ (Eq.~\eqref{eq:omega}) and the auxiliary $\mathcal{L}_{\text{geo}}$ (Eq.~\eqref{eq:loss}), never as inputs. This setup encourages the network to learn the warping algebra from the geometric supervision rather than rely on input-space normalization. Fig.~\ref{fig:warp-residuals} suggests the network does pick up this implicit warp: across all matched plane pairs, both histograms collapse markedly toward zero after refinement---median $\|\Delta\bm{n}\|_{2}$ drops from $0.200$ to $0.084$ and median $|\Delta d|$ from $0.072$ to $0.064$---even though neither residual is fed in, and the effect transfers across ScanNet1500 (in-distribution) and 7Scenes (zero-shot).

\subsection{Rotation as a Continuous 6-D Representation}
\label{app:geo-rot6d}

The rotation part of the pose residual $\bm{z}_{0}$ in Eq.~\eqref{eq:z0} is encoded in the 6-D representation of~\citet{rot6d}, not as a unit quaternion. Under a flow-matching objective this choice is forced, not cosmetic: quaternions carry a topological discontinuity that interacts destructively with continuous probability paths.

\paragraph{Formal definition of a continuous representation.}
Following~\citet{rot6d}, a \emph{representation} of $\SO(3)$ with domain $X\subseteq\mathbb{R}^{d}$ is a pair of maps
\begin{equation}
f:\SO(3)\to X,\qquad g:X\to\SO(3),\qquad g\circ f=\mathrm{id}_{\SO(3)},
\label{eq:rep-def}
\end{equation}
and the representation is \emph{continuous} when both $f$ and $g$ are continuous under the Euclidean topology of $\mathbb{R}^{d}$. Neural networks are continuous under this topology, so a discontinuous $g$ forces the network to reproduce a discontinuity somewhere inside its training support.

\paragraph{The $d\leq 4$ obstruction.}
\citet{rot6d} prove that no continuous representation of $\SO(3)$ exists for $d\leq 4$. Concretely, the discontinuity always lives in the encoder $f$, since the polynomial decoders typically used for these representations are themselves continuous:
\begin{itemize}[leftmargin=1.2em,itemsep=1pt]
  \item \textbf{Euler angles} ($d{=}3$) have a gimbal-lock singularity where $f$ is not unique and $g$ is not one-to-one.
  \item \textbf{Axis-angle} ($d{=}3$) maps $\SO(3)$ to a closed ball of radius $\pi$, with antipodal points on the boundary $\|\bm{r}\|=\pi$ identified to the same rotation; the encoder $f$ must therefore choose one of the two boundary representatives, creating a sign-flip discontinuity at $\|\bm{r}\|=\pi$.
  \item \textbf{Unit quaternions} ($d{=}4$) form a \emph{double cover} of $\SO(3)$: every $R$ maps to an antipodal pair $\pm\bm{q}$ on $S^{3}$. The decoder $g:\mathbb{R}^{4}\!\to\!\SO(3)$ is the standard quaternion-to-matrix polynomial, which is continuous everywhere; the obstruction lies in the encoder $f:\SO(3)\!\to\! S^{3}$, which must select one of $\pm\bm{q}$ globally and therefore admits no continuous global section---arbitrarily close rotations either side of the chosen branch are encoded to far-apart quaternions of opposite sign.
\end{itemize}
$\SO(3)$ is compact and topologically non-trivial, so no homeomorphism to $\mathbb{R}^{d}$ exists for any finite $d$; a continuous \emph{embedding} is the best one can hope for, and \citet{rot6d} show $d=5$ is the minimum ($d=6$ is the simplest practical construction).

\paragraph{The 6-D continuous representation.}
The encoder drops the third column of $R$,
\begin{equation}
f\bigl([\,\bm{r}_{1}\mid\bm{r}_{2}\mid\bm{r}_{3}\,]\bigr) \;=\; [\bm{r}_{1};\,\bm{r}_{2}]\in\mathbb{R}^{6}.
\label{eq:rot6d-encode}
\end{equation}
The decoder $g:\mathbb{R}^{6}\to\SO(3)$ is Gram--Schmidt orthonormalization on the two column vectors: given $(\bm{a}_{1},\bm{a}_{2})\in\mathbb{R}^{3}\times\mathbb{R}^{3}$ linearly independent,
\begin{equation}
\bm{b}_{1}=N(\bm{a}_{1}),\quad
\bm{b}_{2}=N\!\bigl(\bm{a}_{2}-(\bm{b}_{1}\!\cdot\!\bm{a}_{2})\,\bm{b}_{1}\bigr),\quad
\bm{b}_{3}=\bm{b}_{1}\times\bm{b}_{2},
\label{eq:rot6d-decode}
\end{equation}
with $g(\bm{a}_{1},\bm{a}_{2})=[\bm{b}_{1}\mid\bm{b}_{2}\mid\bm{b}_{3}]$ and $N(\bm{v})=\bm{v}/\|\bm{v}\|_{2}$. Both $f$ and $g$ are continuous on the full-measure open subset of $\mathbb{R}^{6}$ where $\bm{a}_{1},\bm{a}_{2}$ are linearly independent, and $g\circ f=\mathrm{id}_{\SO(3)}$ by construction.

\paragraph{Why continuity is critical under flow matching.}
The flow-matching path of Eq.~\eqref{eq:app-fm-path} is a straight line in the ambient Euclidean space, not on $S^{3}$. The encoder $f$ enters only at the endpoint $\bm{z}_{0}=f(R_{\text{gt}}R_{\text{init}}^{\!\top})$, but precisely because $f$ is discontinuous along the antipodal branch cut, two arbitrarily close ground-truth rotations on either side of the cut are encoded to quaternions of opposite sign $\pm\bm{q}$, producing two FM target velocities $\bm{v}^{\star}=\bm{\epsilon}-(\pm\bm{q})$ that disagree by $2\bm{q}$ for the \emph{same} input geometric condition. The network is then forced to average two conflicting targets near the cut, biasing the learned velocity toward the conditional mean of the bimodal label distribution rather than either branch. With rot6d, the encoder $f$ is continuous on the open dense subset where $R$ has linearly independent first two columns, so $\bm{z}_{0}$ varies continuously with $R$ and $\bm{v}^{\star}$ is single-valued at every $(\bm{z}_{t},t)$. The one-step Euler guarantee established in App.~\ref{app:denoising} therefore holds end-to-end only when the encoder is continuous.

\subsection{Flow Matching as a Linear Denoising Process}
\label{app:denoising}

This subsection reviews the flow-matching objective used by the refiner and then explains why a \emph{single} Euler step at inference is sufficient, via an analogy to Newton's method on linear systems.

\paragraph{Flow-matching preliminaries.}
\label{app:denoising-fm}

Flow matching~\citep{flowmatching,liu2022rectified} defines a continuous probability path $p_{t}(\bm{z})$ between a tractable prior $p_{1}=\mathcal{N}(0,\bm{I})$ and the data distribution $p_{0}$ through a \emph{straight-line} interpolation,
\begin{equation}
\bm{z}_{t} \;=\; (1-t)\,\bm{z}_{0}+t\,\bm{\epsilon},\qquad \bm{\epsilon}\sim\mathcal{N}(0,\bm{I}),\qquad t\in[0,1].
\label{eq:app-fm-path}
\end{equation}
Differentiating with respect to $t$ gives the instantaneous target velocity
\begin{equation}
\bm{v}^{\star}(\bm{z}_{t},t) \;=\; \frac{d\bm{z}_{t}}{dt} \;=\; \bm{\epsilon}-\bm{z}_{0},
\label{eq:app-fm-velocity}
\end{equation}
which is \emph{constant along the line} connecting $\bm{z}_{0}$ and $\bm{\epsilon}$. The training objective regresses a learned velocity field $\bm{v}_{\theta}(\bm{z}_{t},t)$ to this target,
\begin{equation}
\mathcal{L}_{\mathrm{FM}} \;=\; \mathbb{E}_{t,\bm{z}_{0},\bm{\epsilon}}\!\left[\big\|\bm{v}_{\theta}(\bm{z}_{t},t)-\bm{v}^{\star}\big\|_{1}\right].
\end{equation}
At inference, $\bm{z}_{0}$ is recovered by integrating the learned ODE from $t=1$ to $t=0$,
\begin{equation}
\bm{z}_{0} \;=\; \bm{z}_{1} - \int_{0}^{1}\bm{v}_{\theta}(\bm{z}_{t},t)\,dt.
\label{eq:app-fm-inference}
\end{equation}

\paragraph{Comparison with DDPM.}
Standard DDPM~\citep{ddpm} regresses either $\bm{\epsilon}$ or $\bm{z}_{0}$ under a variance-preserving schedule whose probability path is curved (a geodesic on a Gaussian manifold, not a straight line). Integrating a curved path requires many steps to stay close to it, so DDPM samplers typically use 50--1000 steps. Flow matching replaces the curve with a line, so the integrand in Eq.~\eqref{eq:app-fm-inference} becomes constant in $t$ and sampling accuracy decouples from step count.

\paragraph{Why one Euler step suffices: conditional-mean regression at $t=1$.}
\label{app:denoising-newton}
The most striking empirical property of our refiner is that a \emph{single} Euler step at inference (Eq.~\eqref{eq:inf}) matches a many-step DDPM sampler. The clean justification is not that the learned ODE is constant along paths---it is not, since the network regresses the \emph{marginal} velocity $\bm{v}_{\theta}(\bm{z}_{t},t)\!=\!\mathbb{E}[\bm{\epsilon}-\bm{z}_{0}\mid \bm{z}_{t},t]$ rather than the sample-specific $\bm{v}^{\star}$---but rather that, at the single time $t\!=\!1$ we use, this marginal velocity reduces to a conditional-mean estimator of $\bm{z}_{0}$. Concretely, since $\bm{z}_{1}=\bm{\epsilon}\sim\mathcal{N}(0,\bm{I})$ is independent of $\bm{z}_{0}$ under the FM coupling,
\begin{equation}
\bm{v}_{\theta}(\bm{\epsilon},1) \;=\; \mathbb{E}\!\bigl[\bm{\epsilon}-\bm{z}_{0}\,\big|\,\bm{z}_{1}=\bm{\epsilon}\bigr] \;=\; \bm{\epsilon} - \mathbb{E}[\bm{z}_{0}\mid c],
\end{equation}
where $c$ collects the conditioning tokens (geometry $\{\hat{\bm{g}}_{m}\}$ and pose $\hat{\bm{c}}$). A single forward Euler step from $\bm{z}_{1}$ therefore yields
\begin{equation}
\bm{z}_{1} - \bm{v}_{\theta}(\bm{z}_{1},1) \;=\; \bm{\epsilon} - \bigl(\bm{\epsilon} - \mathbb{E}[\bm{z}_{0}\mid c]\bigr) \;=\; \mathbb{E}[\bm{z}_{0}\mid c].
\end{equation}
That is, single-step FM inference is mathematically equivalent to direct conditional-mean regression of $\bm{z}_{0}$ given the geometric conditioning. Additional Euler steps between $t\!=\!1$ and $t\!=\!0$ only matter when the marginal field is non-affine in $t$ along the integration trajectory; they cannot improve the $t\!=\!1$ readout, which is already an unbiased estimator of $\mathbb{E}[\bm{z}_{0}\mid c]$.

\paragraph{Newton-style intuition.}
The structural reason this conditional-mean readout is sharp in our setting is the same one that makes Newton's method exact on linear systems: a tractable forward map admits a closed-form inverse, so iteration is only useful when the map is non-affine. Here the FM forward process $\bm{z}_{t}=(1-t)\bm{z}_{0}+t\bm{\epsilon}$ is affine in $t$, and the conditioning $c$ collapses the residual distribution $p(\bm{z}_{0}\mid c)$ to a low-variance, near-unimodal cluster (the pose-residual is small once a sensible $T_{\text{init}}$ has been provided). Three properties of our problem keep this conditional-mean estimator low-variance:
\begin{itemize}[leftmargin=1.2em,itemsep=1pt]
  \item The latent is $9$-dimensional, far lower than image-scale diffusion, so the marginal velocity is an easy function to fit.
  \item The pose-residual is concentrated near a fixed point in this $9$-D space by construction, via scene normalization and the residual parameterization of Eq.~\eqref{eq:z0}, so $p(\bm{z}_{0}\mid c)$ is sharply peaked.
  \item The FM path is continuous end to end (App.~\ref{app:geo-rot6d}), so the conditional mean varies smoothly with $c$.
\end{itemize}
A single Euler step at $t\!=\!1$ is therefore the natural inference choice for pose-residual denoising under our flow-matching probability path.

\section{Model and Training Details}
\label{app:impl}

\subsection{Plane Head Training Details}
\label{app:plane-training}

This section gives the plane-head representation, the per-view local-frame convention, and the training objective omitted from Sec.~\ref{sec:planehead}.

\paragraph{Representation.} The plane head produces a \emph{patch-level} planar description of each view: every decoder token emits one plane $(\bm{n},d)$, giving one normal-offset pair per patch. Following \textsc{Plana3r}~\citep{plana3r}, each token regresses not $(\bm{n},d)$ directly but a compact tuple $(d^{i}_{\pi},\bm{r}^{i}_{\pi},\bm{q}^{i,j}_{\pi})$: the center depth along the token's viewing ray in source view $I^{i}$, two elliptical radii defining the primitive's disk footprint, and a unit quaternion expressed in the frame of a reference camera $I^{j}$. From this tuple,
\begin{equation}
\label{eq:plane-recover}
\bm{n}_{\pi} \;=\; R(\bm{q}^{i,j}_{\pi})\,[0,0,1]^{\!\top},
\qquad
d_{\pi} \;=\; -\bm{n}_{\pi}^{\!\top}\bm{P}_{\pi},
\end{equation}
with $\bm{P}_{\pi}$ the primitive's $3$-D center in $I^{j}$, so $\bm{n}_{\pi}$ and $d_{\pi}$ live in $I^{j}$. Our sole departure from \textsc{Plana3r} is the choice of $j$: the original fixes $j{=}1$ (a single global frame, which forces view-$2$ primitives to be transformed through the predicted pose before supervision), whereas we set $j{=}i$, so every primitive lives in its \emph{own source camera's frame}. Three consequences follow. \textbf{(i)} Plane prediction decouples from pose and scales directly to $n\geq 2$ views. \textbf{(ii)} The plane-head loss decomposes per view: each view's primitives are compared only against its own ground truth, so no pose enters the plane-head gradient. Training still uses image pairs, and we keep \textsc{Plana3r}'s paired sampling, siamese encoder, and cross-attention decoders to preserve metric-scale consistency within each pair. \textbf{(iii)} The warping identity~\eqref{eq:warp-prop} becomes \emph{observable}: the two views produce genuinely distinct $(\bm{n}^{1},d^{1})$ and $(\bm{n}^{2},d^{2})$, so the residuals $(\Delta\bm{n},\Delta d)$ of Eq.~\eqref{eq:residual-prop} turn into non-trivial conditioning signals that carry the pose error into Sec.~\ref{sec:refiner}. All remaining components (ViT encoder, two cross-attention decoders, the hierarchical primitive prediction architecture (HPPA) with low $\tfrac{H}{16}\!\times\!\tfrac{W}{16}$ and high $\tfrac{H}{8}\!\times\!\tfrac{W}{8}$ resolutions selectively fused, and the \textsc{PlanarSplatting} rasterizer~\citep{planarsplatting}) are inherited from \textsc{Plana3r} unchanged.

\paragraph{Training objective and refiner input.} We supervise the head with the \emph{plane-only} losses of \textsc{Plana3r}, applied per view in its own local frame. Following~\citep[Eq.~3]{plana3r}, the patch loss on each HPPA resolution $*\!\in\!\{\text{low},\text{high}\}$ is
\begin{equation}
\label{eq:plane-patch}
\mathcal{L}^{\text{patch}}_{*} \;=\; \alpha_{1}\big\|1-(\bm{N}^{\text{patch}}_{*})^{\!\top}\bm{N}^{\text{r, gt}}_{*}\big\|_{1} + \alpha_{1}\big\|\bm{N}^{\text{patch}}_{*}-\bm{N}^{\text{r, gt}}_{*}\big\|_{1} + \alpha_{2}\big\|\bm{D}^{\text{patch}}_{*}-\bm{D}^{\text{r, gt}}_{*}\big\|_{1},
\end{equation}
where $(\bm{N}^{\text{r, gt}}_{*},\bm{D}^{\text{r, gt}}_{*})$ are the source-view ground-truth normal and depth maps resized to the HPPA grid. The rendering loss $\mathcal{L}^{\text{render}}_{*}$~\citep[Eq.~4]{plana3r} has the identical functional form, substituting $\bm{N}^{\text{patch}}_{*}\!\to\!\bm{N}^{\text{render}}_{*}$ and $\bm{D}^{\text{patch}}_{*}\!\to\!\bm{D}^{\text{render}}_{*}$ (the full-resolution \textsc{PlanarSplatting} renderings) against the original-resolution $(\bm{N}^{\text{gt}},\bm{D}^{\text{gt}})$ with weights $(\beta_{1},\beta_{2})$ and index $*\!\in\!\{\text{low},\text{high},\text{selected}\}$. The plane-head objective sums the two losses over the two views,
\begin{equation}
\label{eq:plane-loss}
\mathcal{L}_{\text{plane}} \;=\; \sum_{i=1,2}\big[\mathcal{L}^{\text{patch},(i)}+\mathcal{L}^{\text{render},(i)}\big].
\end{equation}
We drop the pose loss of the original \textsc{Plana3r}~\citep[Eq.~5]{plana3r}: under our local-frame convention it is both unnecessary (pose is not part of the head's output) and inconsistent with the per-view decomposition above. At inference, the refiner consumes only the low-resolution primitives $\{(\bm{n}^{i}_{m},d^{i}_{m})\}_{m=1}^{M}$ from the native $\tfrac{H}{16}\!\times\!\tfrac{W}{16}$ HPPA grid (without the deconvolution upsampler), yielding $M$ plane correspondences per pair; this keeps $M$ small enough for full self-attention in the refiner and retains the resolution at which the planar assumption is most reliable.

\paragraph{Training data.} Both the plane head and the refiner are trained on the same four-indoor-scene mixture used by \textsc{Plana3r}~\citep{plana3r}, summarized in Tab.~\ref{tab:training-data}: $\sim\!3.94$M two-view pairs from \textsc{ScanNetV2}~\citep{scannet}, \textsc{ScanNet++}~\citep{scannetpp}, \textsc{ARKitScenes}~\citep{arkitscenes}, and \textsc{Habitat}~\citep{habitat}. Pseudo ground-truth normal maps are generated by \textsc{Metric3Dv2}~\citep{metric3d}; depth maps are taken directly from the datasets.

\begin{table}[t]
\centering
\small
\caption{\textbf{Training data and sample sizes.} Same mixture used by \textsc{Plana3r}~\citep{plana3r}.}
\label{tab:training-data}
\begin{tabular}{lcr}
\toprule
Dataset & Type & \# Pairs \\
\midrule
\textsc{ScanNetV2}~\citep{scannet}        & Indoor / Real      & $610$K \\
\textsc{ScanNet++}~\citep{scannetpp}      & Indoor / Real      & $810$K \\
\textsc{ARKitScenes}~\citep{arkitscenes}  & Indoor / Real      & $2{,}400$K \\
\textsc{Habitat}~\citep{habitat}          & Indoor / Synthetic & $120$K \\
\midrule
\textbf{Total}                             &                    & $\bm{\sim\!3{,}940}$\textbf{K} \\
\bottomrule
\end{tabular}
\end{table}

\begin{figure*}[!t]
\centering
\vspace{-4pt}
\begin{minipage}[c]{0.32\linewidth}
  \centering
  \includegraphics[width=\linewidth]{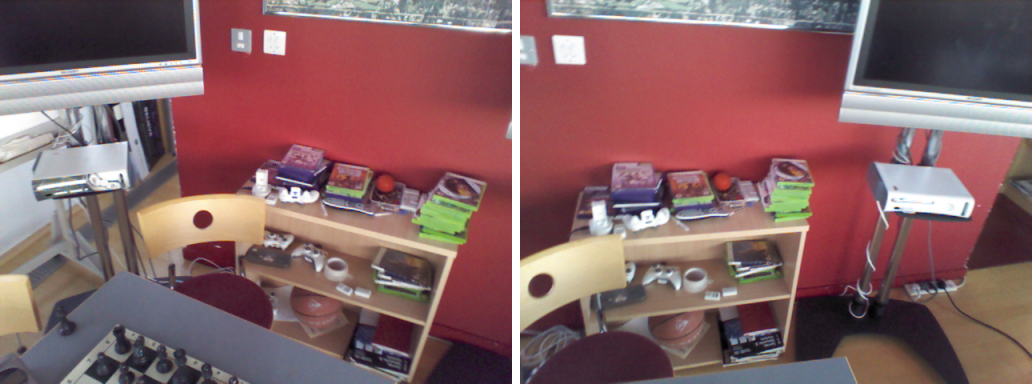}\\[2pt]
  \includegraphics[width=\linewidth]{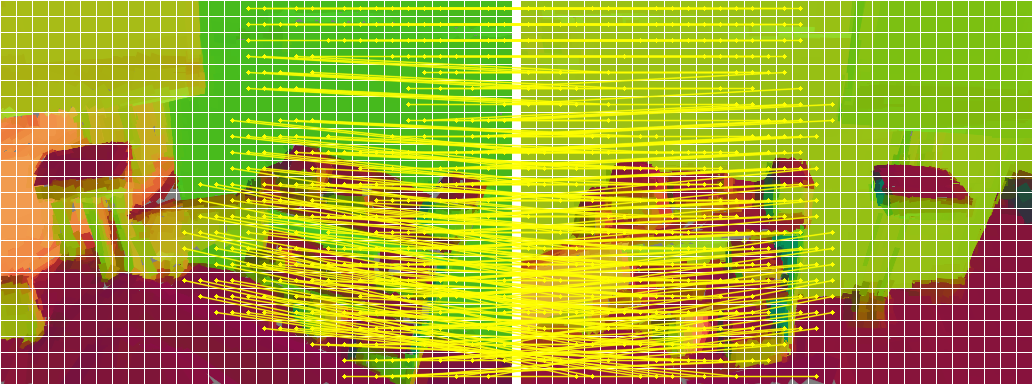}\\[2pt]
  {\small (a) \textsc{chess}, \texttt{seq-01}}
\end{minipage}\hfill
\begin{minipage}[c]{0.32\linewidth}
  \centering
  \includegraphics[width=\linewidth]{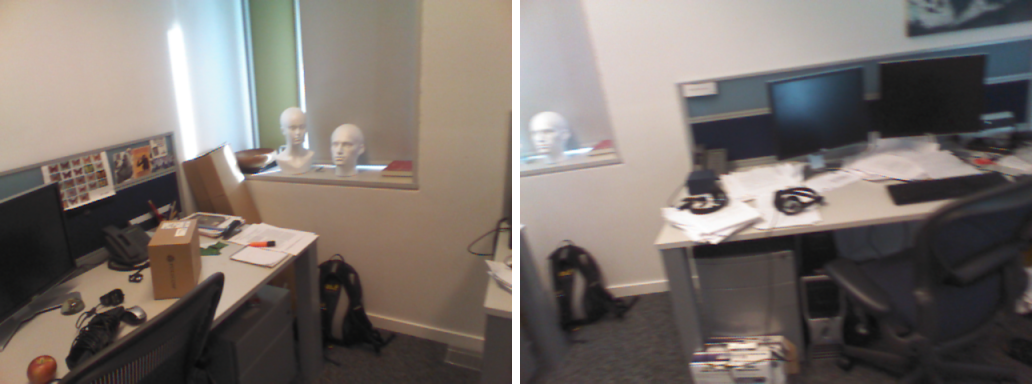}\\[2pt]
  \includegraphics[width=\linewidth]{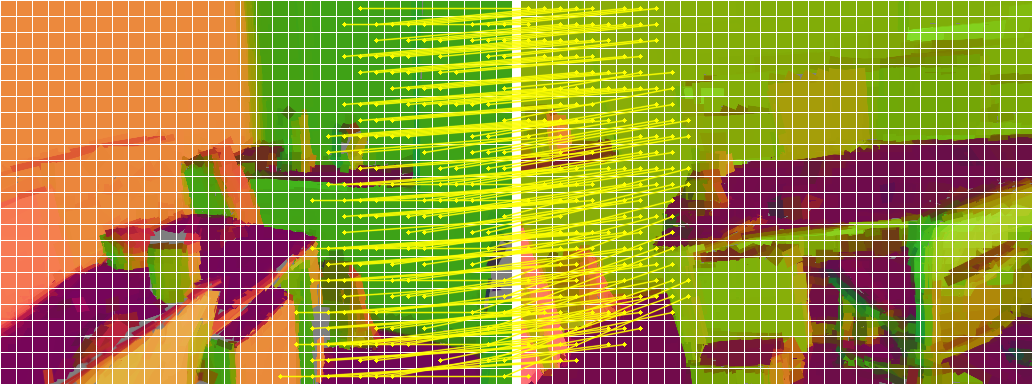}\\[2pt]
  {\small (b) \textsc{office}, \texttt{seq-01}}
\end{minipage}\hfill
\begin{minipage}[c]{0.32\linewidth}
  \centering
  \includegraphics[width=\linewidth]{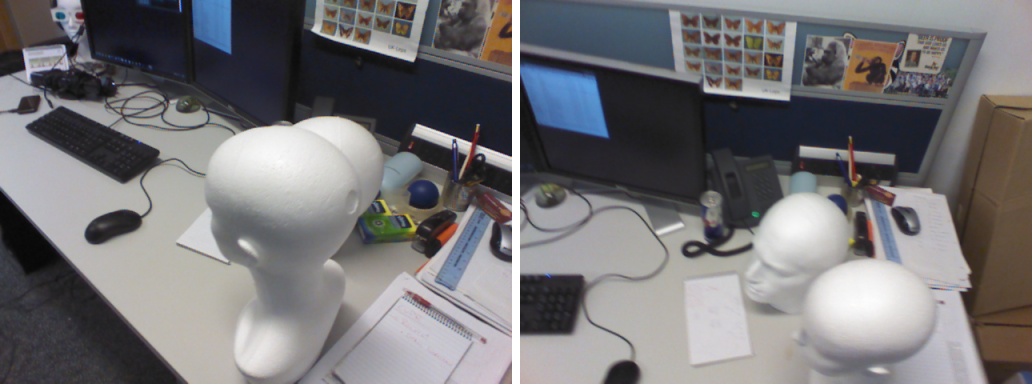}\\[2pt]
  \includegraphics[width=\linewidth]{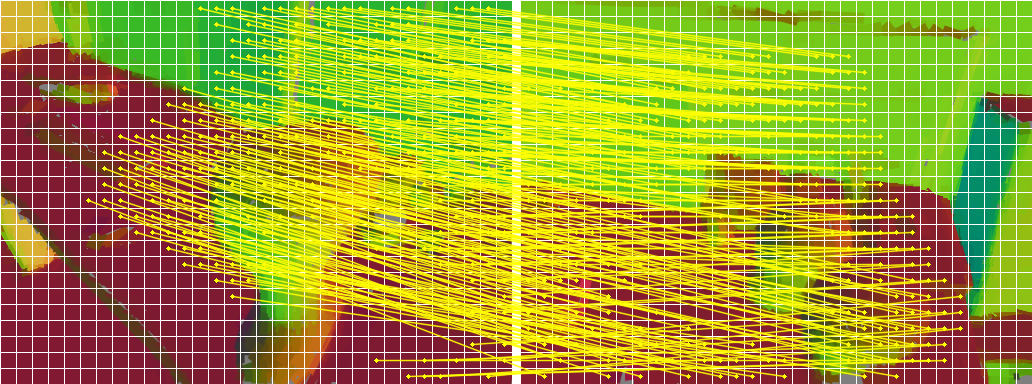}\\[2pt]
  {\small (c) \textsc{stairs}, \texttt{seq-01}}
\end{minipage}
\caption{\textbf{Predicted plane primitives and matched patch correspondences.} Three \textsc{7Scenes} pairs (columns). \textbf{Top}: input RGB (view 1 $|$ view 2). \textbf{Bottom}: \textsc{PlanarSplatting}-rasterized normal from our plane head with the $24\!\times\!32$ patch grid (white); yellow lines connect \textsc{MASt3R}'s matched patch centers, which form $\{\bm{g}_m\}_{m=1}^{M}$ in Eq.~\eqref{eq:geotoken}.}
\label{fig:plane-normals}
\vspace{-6pt}
\end{figure*}

\paragraph{Visualization.} Fig.~\ref{fig:plane-normals} visualizes both the plane head's per-pixel rasterized normal and the \textsc{MASt3R}~\citep{mast3r} patch correspondences entering the refiner as $\{\bm{g}_m\}_{m=1}^{M}$ (Eq.~\eqref{eq:geotoken}) on three \textsc{7Scenes} pairs. Color-uniform tiles correspond to single plane primitives; the yellow corr lines pair patches that, after rasterization, look like the same physical plane viewed from two cameras.

\subsection{Refiner Architecture Details}
\label{app:refiner-arch}

This section expands the multi-query denoiser introduced in Sec.~\ref{sec:refiner} with the per-query construction, attention depth, output head, and the precomputable-context optimization that the single-step inference relies on.

\paragraph{Per-query latent-time and geometry bias.} The latent-time component of Eq.~\eqref{eq:qshared} uses $\phi_{z}\!:\!\mathbb{R}^{9}\!\to\!\mathbb{R}^{d}$ and $\phi_{t}\!:\!\mathbb{R}^{d_{t}}\!\to\!\mathbb{R}^{d}$ as small MLPs, with $\mathrm{SinEmb}(\cdot)$ a sinusoidal time embedding. The per-match geometry bias $\bm{q}_{m}=\bm{q}_{0}+W_{b}\,\hat{\bm{g}}_{m}$, with learned $W_{b}\!\in\!\mathbb{R}^{d\times d}$, grounds query $m$ in correspondence $m$.

\paragraph{Cross-attention stack and output head.} The $M$ queries pass through $L_{\text{den}}{=}4$ pre-norm cross-attention blocks attending to the context $\hat{\bm{C}}$, followed by a $\mathrm{LayerNorm}{+}\mathrm{Linear}(d,9)$ head that emits the per-match velocity predictions $\bm{v}_{m}(\bm{z}_{t},t)$.

\paragraph{Precomputable context.} The time step $t$ modulates only the queries (no AdaLN-Zero on $\hat{\bm{C}}$), so $\hat{\bm{C}}$ is computed once per pair and reused across all queries and Euler steps---a natural fit for the single-step inference of Sec.~\ref{sec:refiner}.

\subsection{Training and Inference Details}
\label{app:training-recipe}

\paragraph{Implementation.} The plane head adopts the training recipe of \textsc{Plana3r}~\citep{plana3r} adapted to the per-view local-frame convention of Sec.~\ref{sec:planehead}: the ViT encoder and the two cross-attention decoders are initialized from \textsc{DUSt3R}'s $512$-DPT pre-trained weights, optimized with AdamW~\citep{loshchilov2017decoupled} (learning rate $1\!\times\!10^{-4}\!\to\!1\!\times\!10^{-6}$) under the plane-related losses of Eq.~\eqref{eq:plane-loss} at input resolution $512\!\times\!384$, with a one-epoch warm-up under the patch loss alone followed by ten epochs jointly optimizing patch and rendering losses; the gradient threshold for selectively fusing the low- and high-resolution primitives is $g_{\text{th}}\!=\!0.5$. With the plane head frozen, the diffusion refiner and the score head are then trained on our four-dataset mixture (Tab.~\ref{tab:training-data}) with Eq.~\eqref{eq:loss} for $20$ epochs at per-GPU batch size $12$, using AdamW with the same $1\!\times\!10^{-4}\!\to\!1\!\times\!10^{-6}$ schedule, $L_{\text{cond}}\!=\!L_{\text{den}}\!=\!4$ Transformer blocks, and $M$ query tokens per pair. At inference, a single Euler step (Eq.~\eqref{eq:inf}) yields the refined rotation $R^{\star}$; translation is recovered by the adaptively regularized Levenberg--Marquardt solver of Eq.~\eqref{eq:trans-refine}. Total compute is $128$ GPU-days on NVIDIA A800 GPUs.

The remainder of this subsection details the inference-time solvers omitted from Sec.~\ref{sec:refiner}.
\label{app:trans-solver}

\paragraph{Scene-range normalization.} The per-pair scalar $\sigma$ that non-dimensionalises $d$ in Eq.~\eqref{eq:sest} is the mean absolute plane offset across all $S$ patches per view,
\begin{equation}
\sigma \;=\; \tfrac{1}{2S}\!\Big(\textstyle\sum_{p=1}^{S}|d^{1}_{p}|+\sum_{p=1}^{S}|d^{2}_{p}|\Big),
\label{eq:sigma-def}
\end{equation}
which makes $\bar d^{v}_{m}=d^{v}_{m}/\sigma$ approximately $O(1)$ across pairs.

\paragraph{Two-view translation solver (Eq.~\eqref{eq:trans-refine}).} Locking $s=\|\bm{t}_{\text{init}}\|$, the two data residuals are
\begin{align}
\Delta d_{m}(\hat{\bm{t}}) &\;=\; d^{1}_{m}-d^{2}_{m}+(R^{\star}\bm{n}^{2}_{m})^{\!\top}(s\hat{\bm{t}}),\\
\Delta\bm{P}_{m}(\hat{\bm{t}}) &\;=\; \bm{P}^{1}_{m}-R^{\star}\bm{P}^{2}_{m}-s\hat{\bm{t}},
\end{align}
where $\bm{P}^{v}_{m}=z^{v}_{m}K_{v}^{-1}[u_{m};1]$ is the back-projection of matched pixel $u_{m}$ through the dense depth $z^{v}_{m}$ rasterized by \textsc{PlanarSplatting} from the per-patch plane primitives. Both $\rho_{d},\rho_{p}$ are Huber losses with threshold $0.1$ and weights $w_{d}\!=\!w_{p}\!=\!1$. The init-anchored Tikhonov weight $\lambda_{\text{reg}}$ adapts to the init's quality: with $r^{(0)}_{m}=(R^{\star}\bm{n}^{2}_{m})^{\!\top}\bm{t}_{\text{init}}-(d^{2}_{m}-d^{1}_{m})$ the $\Delta d$ residual at $\hat{\bm{t}}_{\text{init}}$, we set $\sigma_{\text{init}}=1.4826\,\mathrm{MAD}(r^{(0)})$ on its $20$--$80\%$ IQR-trimmed core and
\begin{equation}
\lambda_{\text{reg}}=\mathrm{clip}\!\big(\lambda_{0}(\sigma_{\text{ref}}/\sigma_{\text{init}})^{\alpha},\;\lambda_{\min},\,\lambda_{\max}\big),\quad (\lambda_{0},\sigma_{\text{ref}},\alpha,\lambda_{\min},\lambda_{\max})=(200,\,0.5,\,4,\,1,\,10^{6}),
\end{equation}
so a small $\sigma_{\text{init}}$ (init already consistent with the predicted planes) inflates $\lambda_{\text{reg}}$ and the solver returns $\approx\hat{\bm{t}}_{\text{init}}$, while a large $\sigma_{\text{init}}$ lets the data terms drive the solve. We optimize via \texttt{scipy.optimize.least\_squares} (TRF, Huber $f_{\text{scale}}=0.1$) over $\hat{\bm{t}}\in S^{2}$, $\le\!15$ Jacobian evaluations.

\paragraph{Multi-view aggregation ($n$-view).} For streaming $n$-view inputs we follow the \textsc{Glomap}~\citep{glomap} global-SfM pipeline---rotation averaging, multi-view track construction, point-based global positioning, and bundle adjustment---over the refinement-window edge set $\mathcal{E}$. The diffusion refiner is invoked only for the per-edge \emph{rotation} $R^{\star}_{ij}$; the per-edge translation $\bm{t}^{\text{raw}}_{ij}$ is taken directly from the upstream estimator and used unchanged downstream.

\textit{Rotation averaging.} Per-frame absolute rotations are recovered from the pairwise $\{R^{\star}_{ij}\}$ by robust least-squares~\citep{rotaveraging,rotaveraging_iccv},
\begin{equation}
\{R_{i}\}^{\star}=\argmin_{\{R_{i}\}}\sum_{(i,j)\in\mathcal{E}}\rho\!\left(d(R_{j}^{\!\top}R^{\star}_{ij}R_{i},\,\bm{I})\right),
\label{eq:ra}
\end{equation}
with $\rho$ a Huber loss and $d$ chordal distance.

\textit{Tracks and translation warm-start.} Multi-view tracks $\{\tau_{k}\}$ are built by union-find over pixel-cell-quantized correspondences from the matcher; the upstream's raw pairwise translations $\{\bm{t}^{\text{raw}}_{ij}\}$ together with $\{R_{i}\}^{\star}$ are composed into an absolute c2w trajectory and used as warm-start camera centers for the positioner below.

\textit{Global positioning.} With $\{R_{i}\}$ fixed and tracks $\{\tau_{k}\}$ attached with bearings $\bm{v}_{ki}=R_{i}^{\!\top}K_{i}^{-1}[u_{ki};1]$, \textsc{Glomap}'s point-based mode~\citep{glomap} jointly optimizes camera centers, $3$-D point positions, and per-observation scales,
\begin{equation}
\{\bm{c}_{i},\bm{X}_{k},s_{ki}\}^{\star}=\argmin\sum_{k}\sum_{(k,i)\in\tau_{k}}\rho\!\left(\|\bm{v}_{ki}-s_{ki}(\bm{X}_{k}-\bm{c}_{i})\|_{2}\right),\quad s_{ki}\geq 0.
\label{eq:positioning}
\end{equation}

\textit{Bundle adjustment.} A Huber-robust BA~\citep{glomap} over $(\bm{c}_{i},\bm{X}_{k})$ with pinhole reprojection error refines centers and $3$-D points; we run \textsc{Glomap} with \texttt{fix\_rotations=True} so that $\{R_{i}\}^{\star}$ from rotation averaging remain frozen throughout positioning and BA, and the entire downstream pipeline only ever consumes the refined rotations. The two-view LM translation solver of Eq.~\eqref{eq:trans-refine} is bypassed entirely at $n$-view scale, since the dense edge set $\mathcal{E}$ already over-constrains every $\bm{c}_{i}$ through the RA + GP + BA pipeline; the same plane-conditioned refiner---whose only role at $n$-view is to produce $\{R^{\star}_{ij}\}$ per edge---therefore serves both regimes without retraining.

\section{Additional Evaluation Setup and Results}
\label{app:eval-extra}

\subsection{Pair and Sequence Construction}
\label{app:eval-data}

This subsection gives the per-sequence quantitative breakdown deferred from Sec.~\ref{sec:experiment} for the two-view and streaming evaluation splits.

\paragraph{7Scenes two-view pair list.} We use all $46$ sequences across the seven scenes (chess, fire, heads, office, pumpkin, redkitchen, stairs) without any train/test restriction. Pairs are formed as $(i,\,i{+}50)$ at stride $10$ within each sequence, producing $95$ pairs per $1000$-frame sequence and $45$ per $500$-frame \emph{stairs} sequence, totalling $40\!\times\!95+6\!\times\!45=4070$ pairs. The exact pair list is released with our code.

\paragraph{Matterport3D two-view pair list.} For an additional zero-shot test, we use the $7996$-pair test set of \textsc{SparsePlanes}~\citep{sparseplane} on Matterport3D~\citep{chang2017matterport3dlearningrgbddata}. Results in Tab.~\ref{tab:matterport-twoview} mirror the pattern from ScanNet1500 and 7Scenes: \textsc{PosePlaner} delivers a large gain on the weaker \textsc{LoFTR} upstream (R@$5$: $10.7\!\to\!20.2$, R@$20$: $30.6\!\to\!52.8$) and a smaller but consistent gain on the stronger \textsc{SuperGlue} upstream (R@$5$: $38.3\!\to\!40.8$, R@$20$: $66.7\!\to\!70.6$).

\begin{table}[t]
\centering
\footnotesize
\caption{\textbf{Two-view rotation refinement on Matterport3D.} Evaluated zero-shot on the $7996$-pair test set of \textsc{SparsePlanes}~\citep{sparseplane}. Rotation AUC@$\tau^\circ$ for two upstream matchers in two configurations: \emph{init} and +\textsc{PosePlaner}. The \emph{Valid} column counts pairs with a valid upstream pose; AUC is computed only on this subset. Higher is better; best per block in \textbf{bold}.}
\label{tab:matterport-twoview}
\begin{tabular}{l|cccc|cccc}
\toprule
\multirow{2}{*}{Method} & \multicolumn{4}{c|}{\textsc{LoFTR}~\citep{loftr}} & \multicolumn{4}{c}{\textsc{SuperGlue}~\citep{superglue}} \\
                        & R@5 & R@10 & R@20 & Valid & R@5 & R@10 & R@20 & Valid \\
\midrule
\emph{init}              & 10.7 & 18.8 & 30.6 & $4385/7996$ & 38.3 & 53.5 & 66.7 & $6853/7996$ \\
+\,\textsc{PosePlaner}        & \textbf{20.2} & \textbf{36.0} & \textbf{52.8} &           & \textbf{40.8} & \textbf{57.2} & \textbf{70.6} &           \\
\bottomrule
\end{tabular}
\end{table}

\paragraph{ScanNet streaming.} The $100$ trajectories of the ScanNet1500 test set (scenes $0707$--$0806$) range in raw length from $163$ to $4168$ frames; we sample at frame gap $2$ to obtain the streaming sequences. For each input-length threshold $N\!\in\!\{200,500,2000\}$ we retain trajectories with $\geq\!2N$ raw frames (equivalently $\geq\!N$ sampled frames)---$99$, $86$, and $7$ respectively---and let the streaming baseline ingest the first $N$ sampled frames while \textsc{PosePlaner} refines the last $100$.

\paragraph{7Scenes streaming.} We form $N$-frame trajectories at frame gap $2$ from the same $46$ sequences as the two-view setup. For $N\!\in\!\{200,500\}$ we take the first $N$ sampled frames of every sequence with $\geq\!2N$ raw frames---$46$ trajectories at $N{=}200$ (every sequence qualifies) and $40$ at $N{=}500$ (the six $500$-frame \emph{stairs} sequences fall below the $1000$-frame raw threshold). For $N{=}2000$ no single sequence is long enough, so we concatenate sequences within each scene end-to-end and take the first $2000$ sampled frames, yielding $5$ scene-level trajectories (\emph{chess, fire, office, pumpkin, redkitchen}; \emph{heads} and \emph{stairs} have insufficient total raw frames after gap-$2$ subsampling).

\paragraph{Outdoor multi-view sequences.} We preserve the source frame order and take the first $N\in\{200,500,2000\}$ frames of each sufficiently long trajectory. Oxford Spires~\citep{DBLP:journals/ijrr/TaoMZWFF26} provides $14$ trajectories at each length, and MatrixCity~\citep{DBLP:conf/iccv/0002JXX0L023} provides $10$ (five aerial and five street-level). The four CrossLoc~\citep{DBLP:conf/cvpr/YanZRLD22} flight segments contain $1176$, $841$, $623$, and $373$ frames, giving four trajectories at $N=200$, three at $N=500$, and none at $N=2000$. CUT3R and TTT3R use the same inputs. We evaluate relative-pose pairs among the final $100$ frames and average per-trajectory AUCs without weighting. Pairs with undefined translation direction are excluded from translation AUC in both branches. Both branches share the reconstruction backend over all $N$ frames; \textsc{PosePlaner} supplies refined pairwise rotations. Table~\ref{tab:crossloc-nview} reports the additional tests on CrossLoc drone sequences at $N=200$ and $N=500$.

\begin{table}[!htbp]
\centering
\small
\setlength{\tabcolsep}{4pt}
\caption{\textbf{Zero-shot multi-view refinement on CrossLoc drone sequences.} R@5/R@20 and t@5/t@20: rotation/translation-direction AUC at $5^\circ$/$20^\circ$ (\%, $\uparrow$). Better results within each pair are \textbf{bold}.}
\label{tab:crossloc-nview}
\begin{tabular}{l|cccc|cccc}
\toprule
\multirow{2}{*}{Method} & \multicolumn{4}{c|}{$N=200$} & \multicolumn{4}{c}{$N=500$} \\
 & R@5 & R@20 & t@5 & t@20 & R@5 & R@20 & t@5 & t@20 \\
\midrule
\textsc{CUT3R} & \textbf{20.5} & 55.2 & 22.6 & 64.6 & 11.7 & 42.7 & 5.1 & 45.0 \\
\textsc{CUT3R}\,+\,\textsc{PosePlaner} & 20.3 & \textbf{60.9} & \textbf{24.6} & \textbf{68.4} & \textbf{12.1} & \textbf{44.5} & \textbf{5.7} & \textbf{48.1} \\
\textsc{TTT3R} & 26.0 & 69.6 & \textbf{33.3} & \textbf{73.5} & 20.5 & 70.9 & 24.8 & 73.6 \\
\textsc{TTT3R}\,+\,\textsc{PosePlaner} & \textbf{26.2} & \textbf{70.0} & 29.6 & 65.9 & \textbf{25.1} & \textbf{73.0} & \textbf{28.2} & \textbf{75.0} \\
\bottomrule
\end{tabular}
\end{table}

\subsection{External Match-Source Sensitivity}
\label{app:match-ablation}

\textsc{PosePlaner}'s geometry tokens $\{\bm{g}_m\}$ come from patch correspondences produced by an external matcher (Sec.~\ref{sec:planehead}); this matcher is in principle independent of the upstream pose source. To verify that the refiner's gain is not tied to a particular matcher, we fix the upstream pose source to either \textsc{SuperGlue} or \textsc{LoFTR} and sweep four matchers (\textsc{SuperGlue}, \textsc{LoFTR}, \textsc{ROMA}, \textsc{MASt3R}) as the source of patch correspondences fed into the diffusion refiner. Tab.~\ref{tab:match-ablation} reports the resulting two-view rotation AUC on ScanNet1500. \textsc{PosePlaner} improves over the init pose for the vast majority of (upstream, matcher) combinations, and stronger matchers (\textsc{ROMA}, \textsc{MASt3R}) yield clearly larger gains; the qualitative trend is preserved across all four matchers, confirming that the refiner is robust to the choice of correspondence source.

\begin{table}[t]
\centering
\footnotesize
\caption{\textbf{Sensitivity to the external match source.} Two-view rotation AUC on ScanNet1500 with the upstream pose fixed to \textsc{SuperGlue} or \textsc{LoFTR}, while the matcher used to produce patch correspondences for \textsc{PosePlaner}'s conditioning is varied across four options. The \emph{init} row reports each upstream's raw pose accuracy from Tab.~\ref{tab:twoview}. Higher is better.}
\label{tab:match-ablation}
\begin{tabular}{l|ccc|ccc}
\toprule
\multirow{2}{*}{Conditioning match} & \multicolumn{3}{c|}{Upstream: \textsc{SuperGlue}} & \multicolumn{3}{c}{Upstream: \textsc{LoFTR}} \\
                                                 & R@5  & R@10  & R@20  & R@5  & R@10  & R@20 \\
\midrule
\emph{init} (no refinement)                      & 43.9 & 62.4  & 76.4    & 11.5 & 25.0  & 44.0   \\
\midrule
+\,\textsc{PosePlaner}, \textsc{SuperGlue}~\citep{superglue} match & 45.1   & 64.1    & 78.0    & 17.8   & 38.8    & 59.5   \\
+\,\textsc{PosePlaner}, \textsc{LoFTR}~\citep{loftr} match         & 43.5   & 62.6    & 76.8    & 14.5   & 31.1    & 50.5   \\
+\,\textsc{PosePlaner}, \textsc{ROMA}~\citep{roma} match           & \textbf{49.8}   & \textbf{69.6}    & \textbf{82.5}    & \textbf{32.9}   &\textbf{56.3}    & \textbf{72.8}   \\
+\,\textsc{PosePlaner}, \textsc{MASt3R}~\citep{mast3r} match       & 48.9   & 68.9    & 82.1    & 29.4   & 52.9    & 70.6   \\
\bottomrule
\end{tabular}
\end{table}

\subsection{Additional Streaming Trajectories}
\label{app:streaming-extra}

\paragraph{Sliding-window protocol.} Each trajectory is $200$ frames long. We slide a $100$-frame window with $50$-frame overlap, so the trajectory contains three refinement windows ($[0,100)$, $[50,150)$, $[100,200)$); within each window, \textsc{PosePlaner} treats every pair as the two-view problem of Sec.~\ref{sec:exp-twoview-new} and the per-pair refined poses are aggregated by rotation averaging~\citep{rotaveraging,rotaveraging_iccv} plus a fix-rotation \textsc{Glomap}-style positioner~\citep{glomap}. The same protocol is used for the main paper's Fig.~\ref{fig:streaming-curves}.

\begin{figure*}[!t]
\centering
\vspace{-4pt}
%
\begin{minipage}[c]{0.258\linewidth}
  \centering
  \begin{tabular}[b]{@{}c@{\hspace{1pt}}c@{}}
    \parbox[b][0.95in][c]{0.13in}{\centering\rotatebox{90}{\small Scene0725\_00}} &
    \includegraphics[height=0.95in]{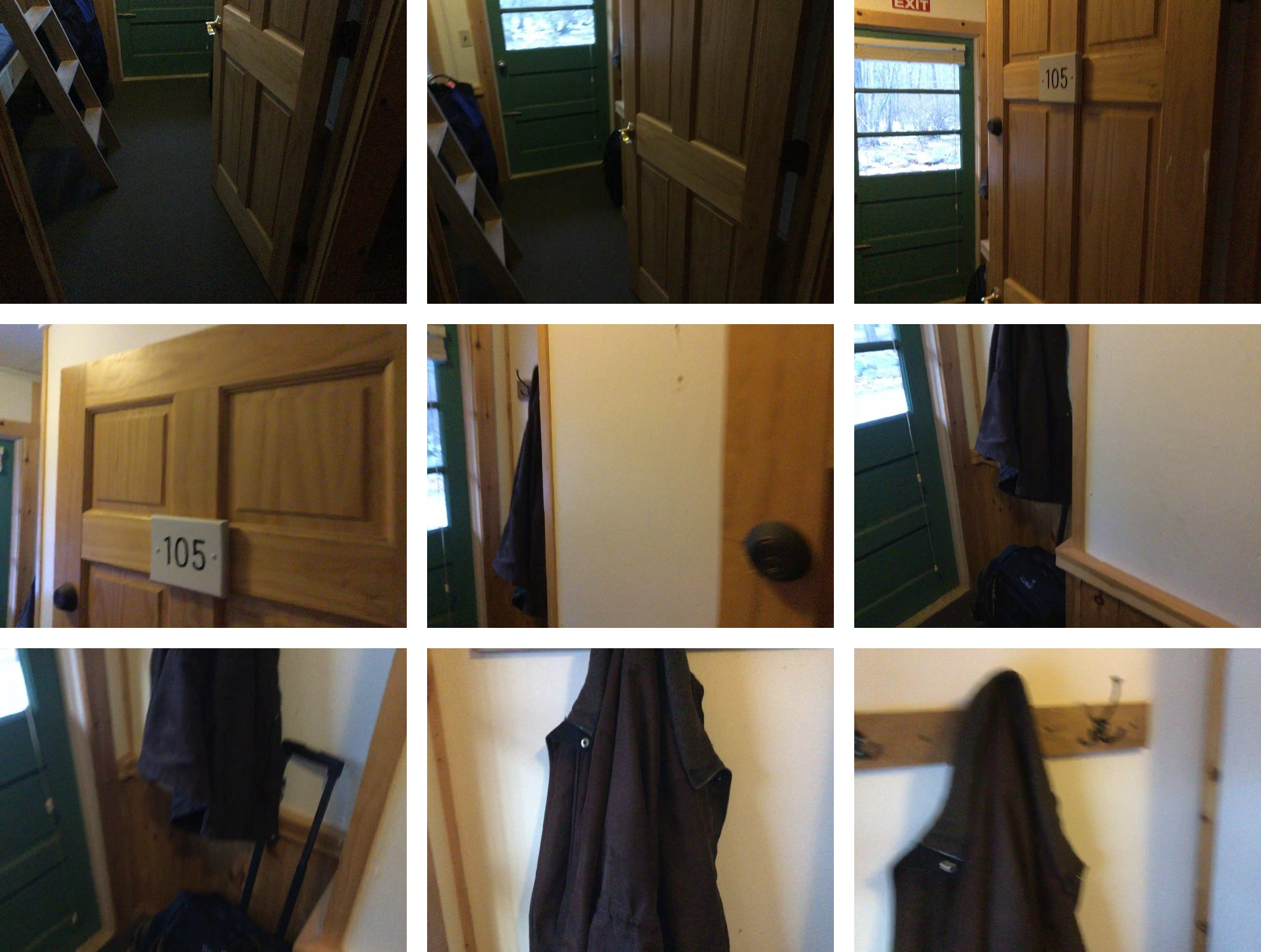}
  \end{tabular}
\end{minipage}%
\begin{minipage}[c]{0.245\linewidth}
  \centering
  \includegraphics[width=\linewidth,height=1.35in,keepaspectratio]{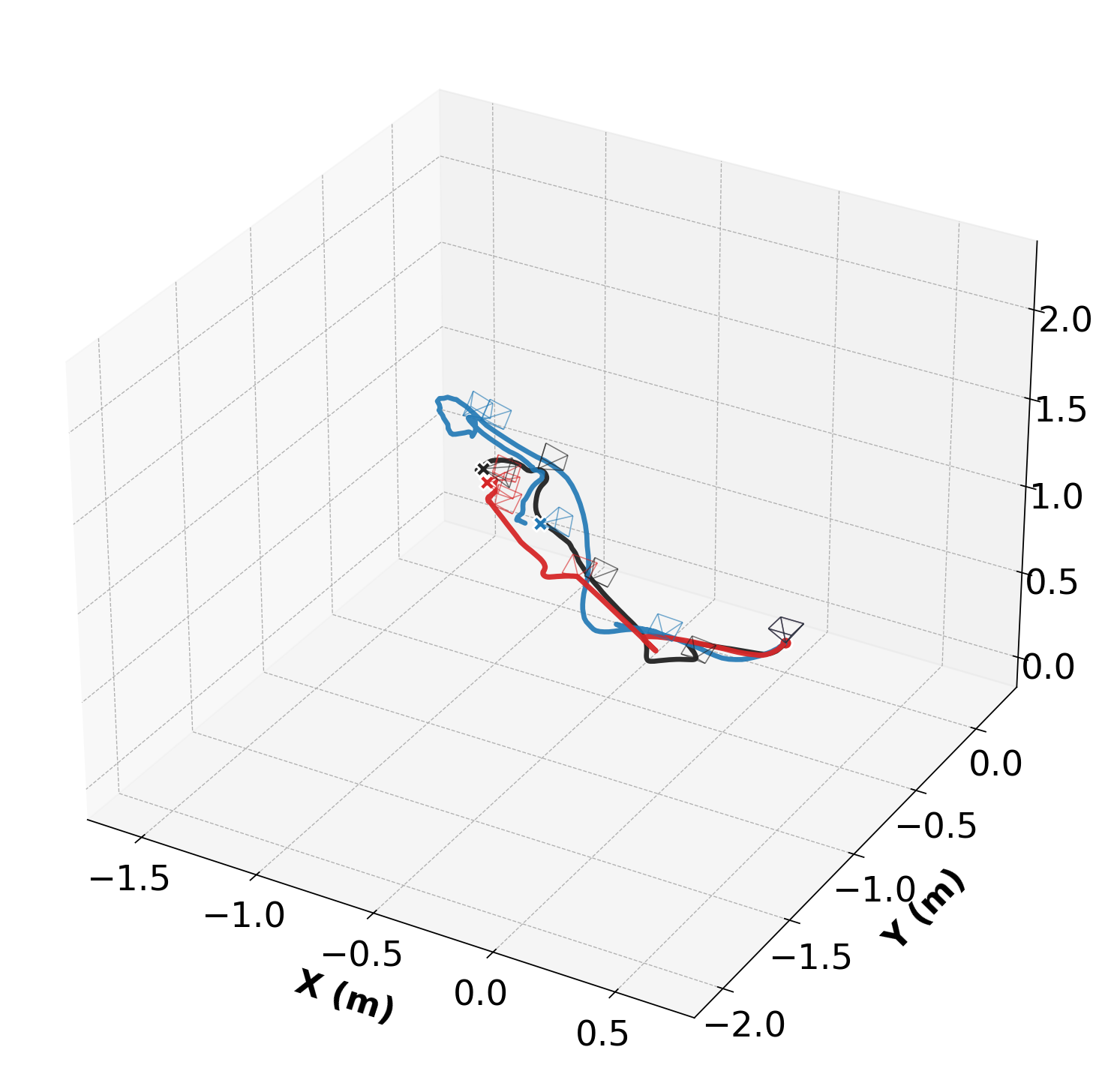}
\end{minipage}%
\begin{minipage}[c]{0.245\linewidth}
  \centering
  \includegraphics[width=\linewidth,height=1.35in,keepaspectratio]{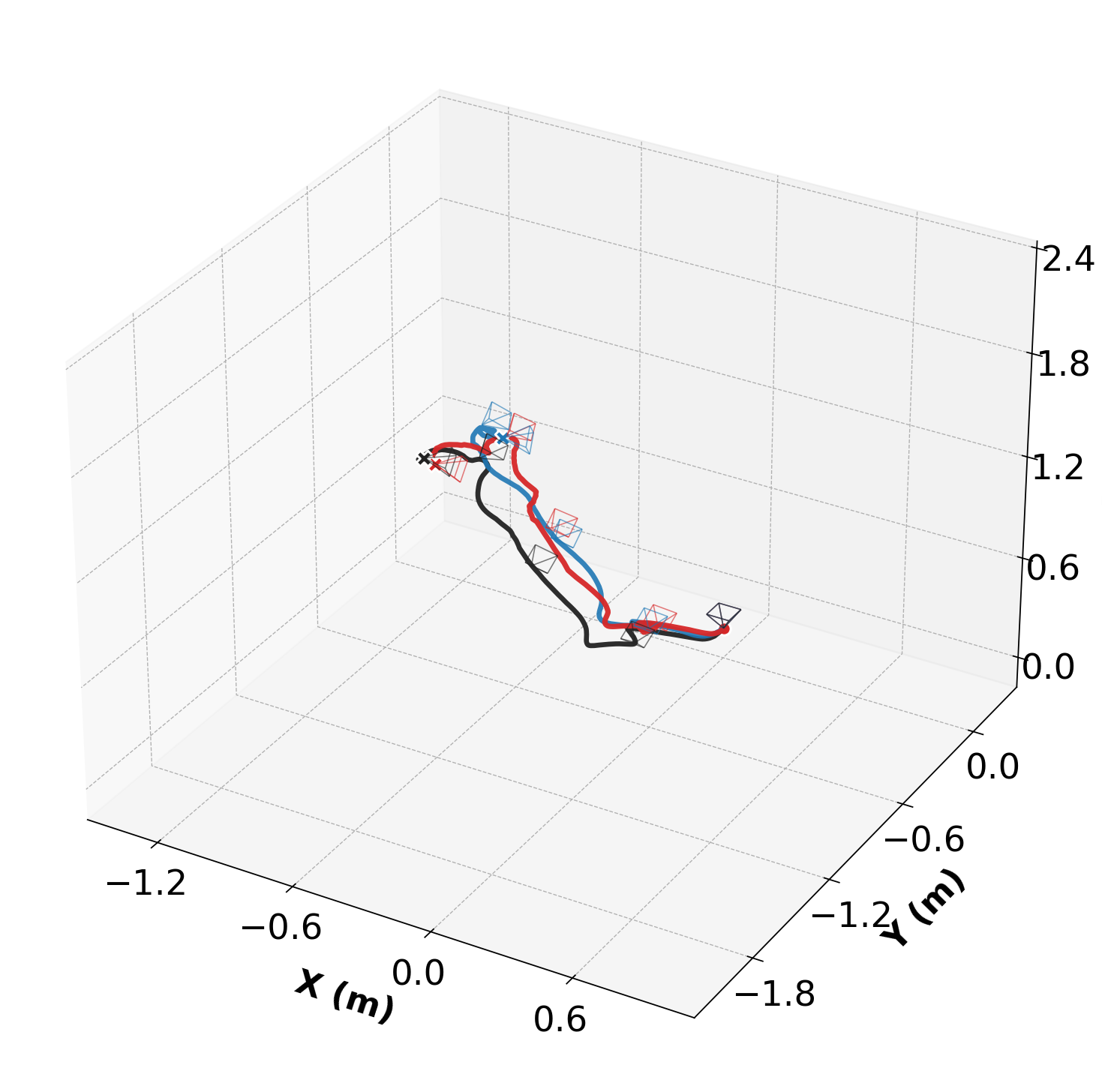}
\end{minipage}%
\begin{minipage}[c]{0.245\linewidth}
  \centering
  \includegraphics[width=\linewidth,height=1.35in,keepaspectratio]{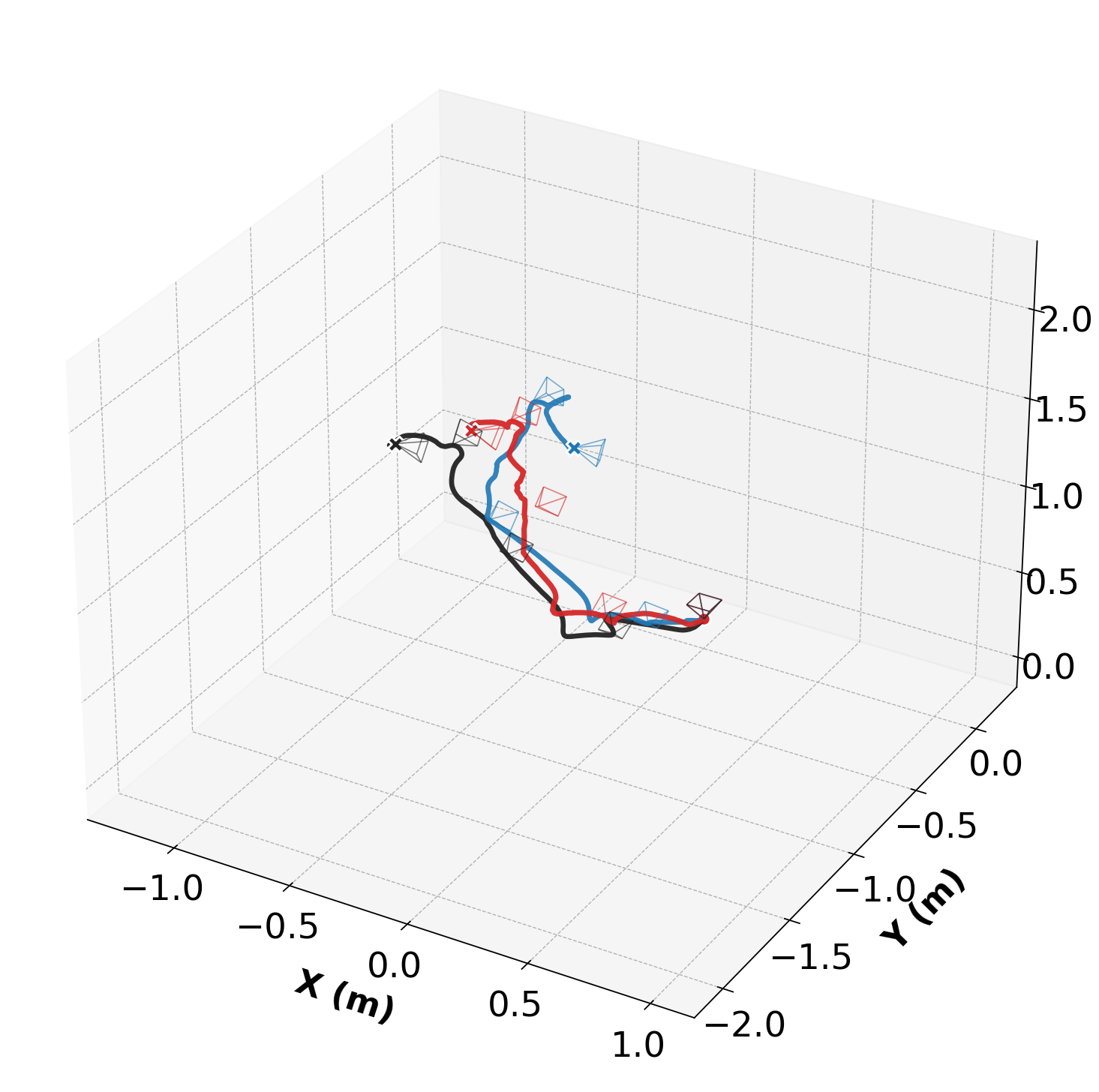}
\end{minipage}\\[1pt]
\begin{minipage}[c]{0.258\linewidth}
  \centering
  \begin{tabular}[b]{@{}c@{\hspace{1pt}}c@{}}
    \parbox[b][0.95in][c]{0.13in}{\centering\rotatebox{90}{\small Scene0733\_00}} &
    \includegraphics[height=0.95in]{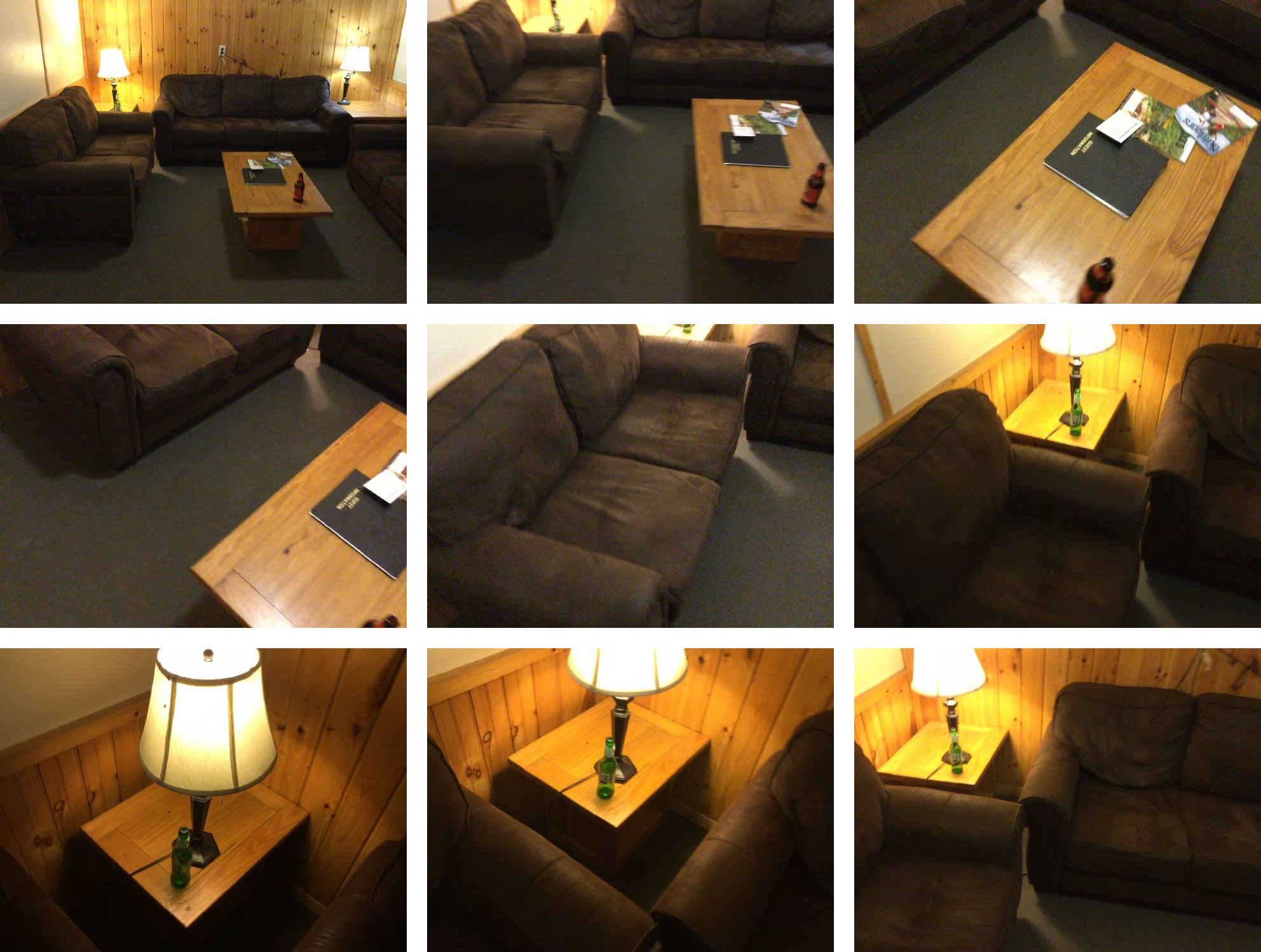}
  \end{tabular}
\end{minipage}%
\begin{minipage}[c]{0.245\linewidth}
  \centering
  \includegraphics[width=\linewidth,height=1.35in,keepaspectratio]{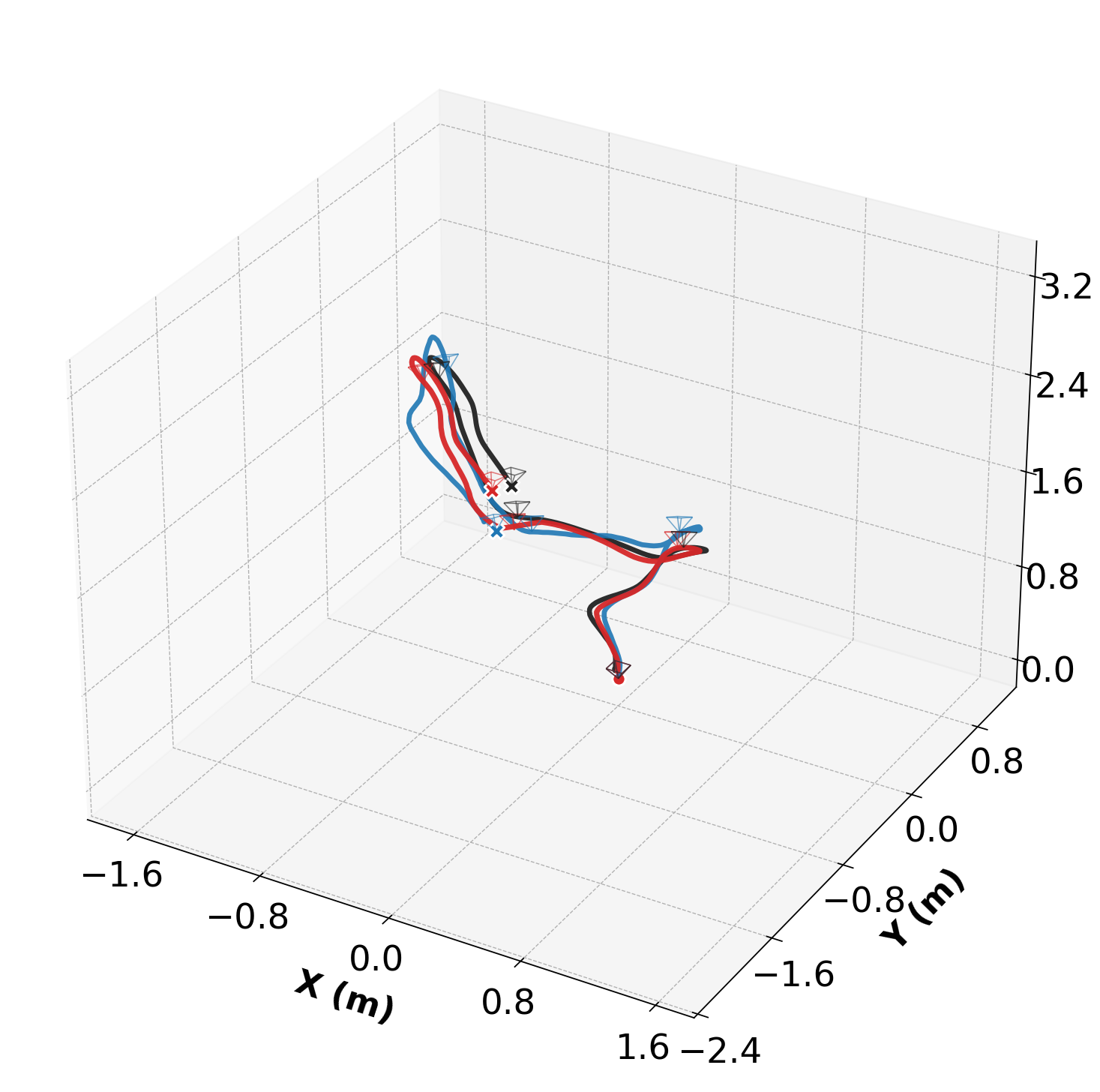}
\end{minipage}%
\begin{minipage}[c]{0.245\linewidth}
  \centering
  \includegraphics[width=\linewidth,height=1.35in,keepaspectratio]{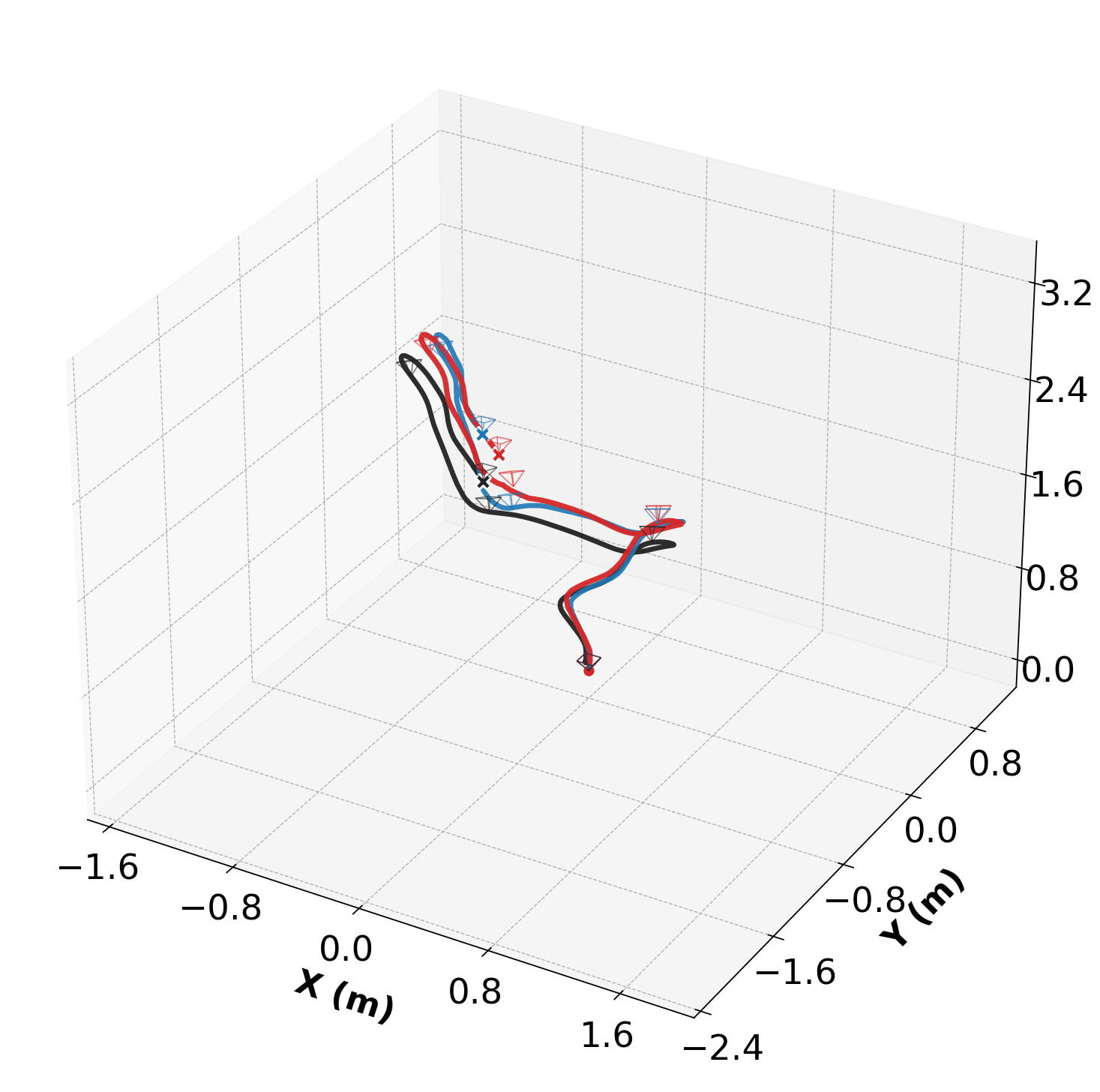}
\end{minipage}%
\begin{minipage}[c]{0.245\linewidth}
  \centering
  \includegraphics[width=\linewidth,height=1.35in,keepaspectratio]{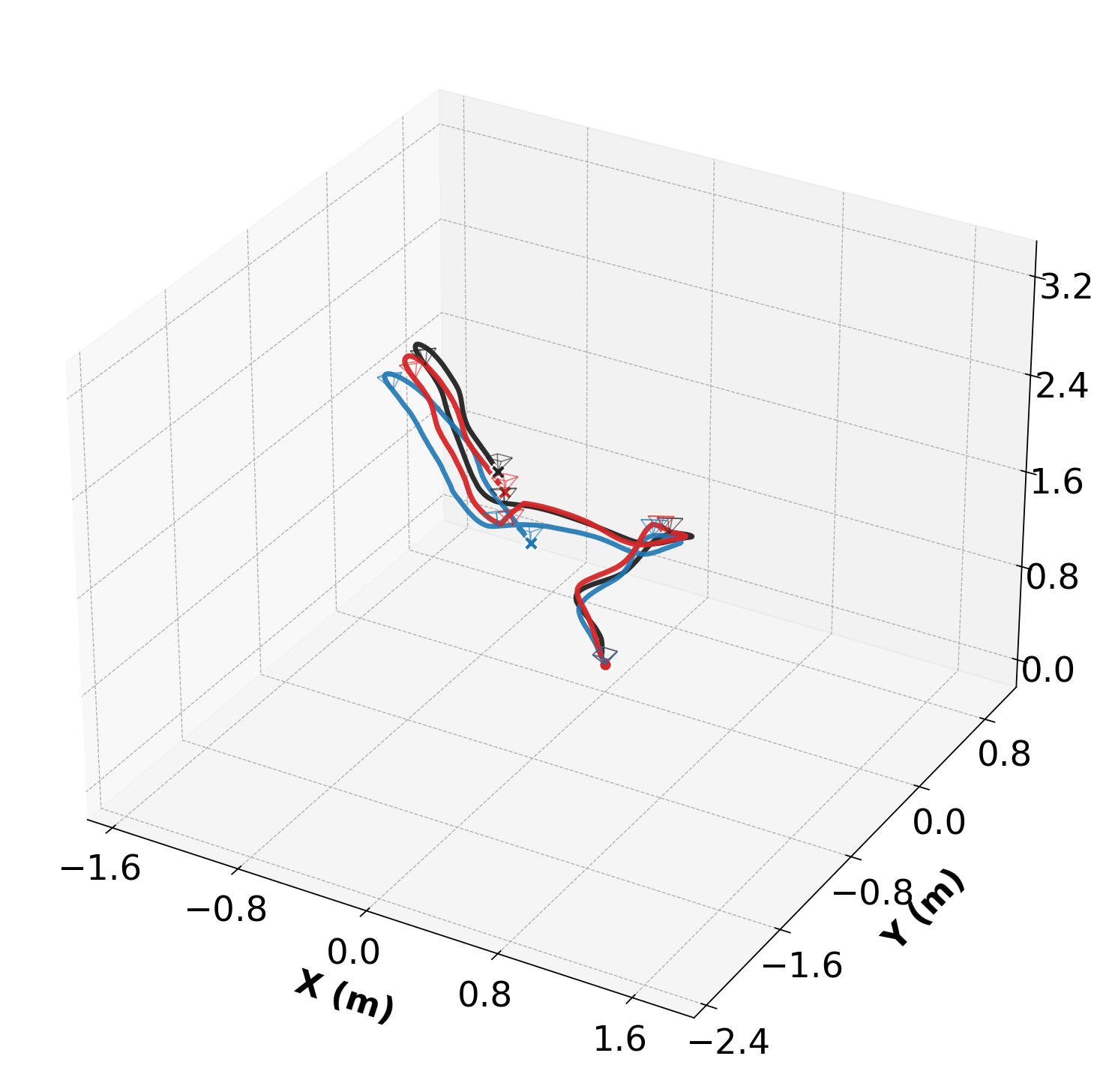}
\end{minipage}\\[1pt]
\begin{minipage}[c]{0.258\linewidth}
  \centering
  \begin{tabular}[b]{@{}c@{\hspace{1pt}}c@{}}
    \parbox[b][0.95in][c]{0.13in}{\centering\rotatebox{90}{\small Scene0806\_00}} &
    \includegraphics[height=0.95in]{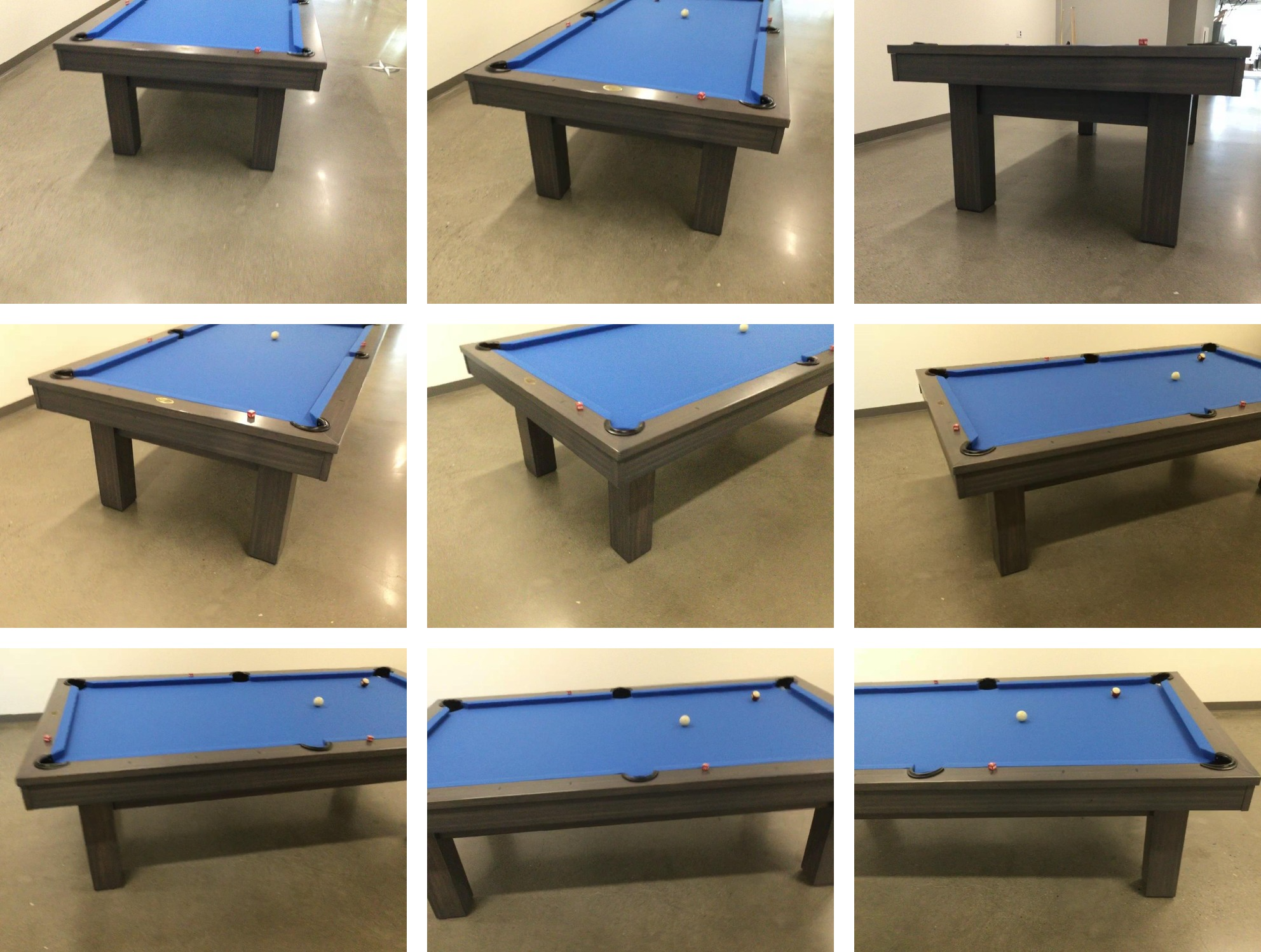}
  \end{tabular}\\[-1pt]
  {\small \phantom{\textsc{CUT3R} $\pm$\,ours}}%
\end{minipage}%
\begin{minipage}[c]{0.245\linewidth}
  \centering
  \includegraphics[width=\linewidth,height=1.35in,keepaspectratio]{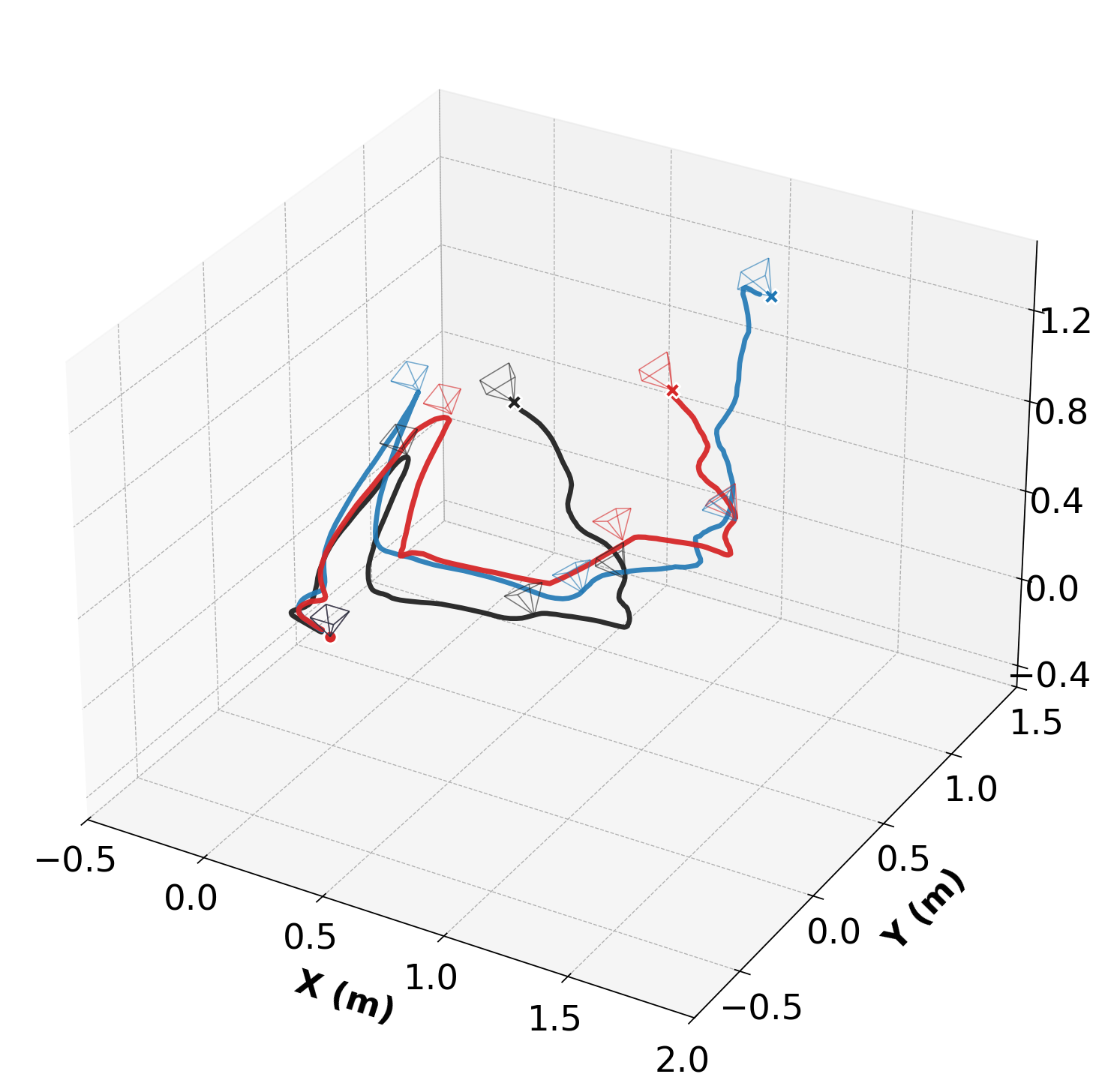}\\[-1pt]
  {\small \textsc{CUT3R} $\pm$\,ours}
\end{minipage}%
\begin{minipage}[c]{0.245\linewidth}
  \centering
  \includegraphics[width=\linewidth,height=1.35in,keepaspectratio]{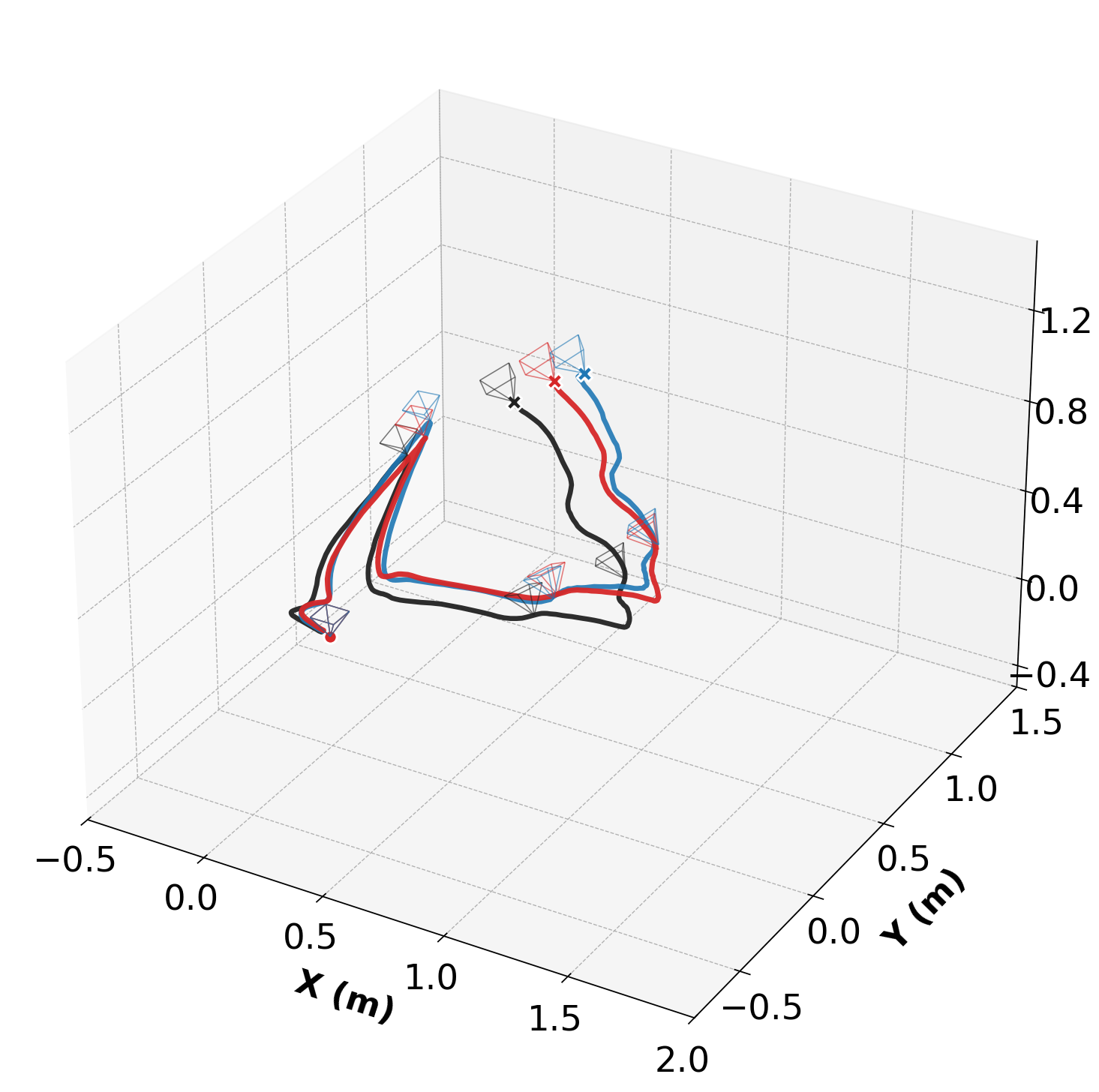}\\[-1pt]
  {\small \textsc{TTT3R} $\pm$\,ours}
\end{minipage}%
\begin{minipage}[c]{0.245\linewidth}
  \centering
  \includegraphics[width=\linewidth,height=1.35in,keepaspectratio]{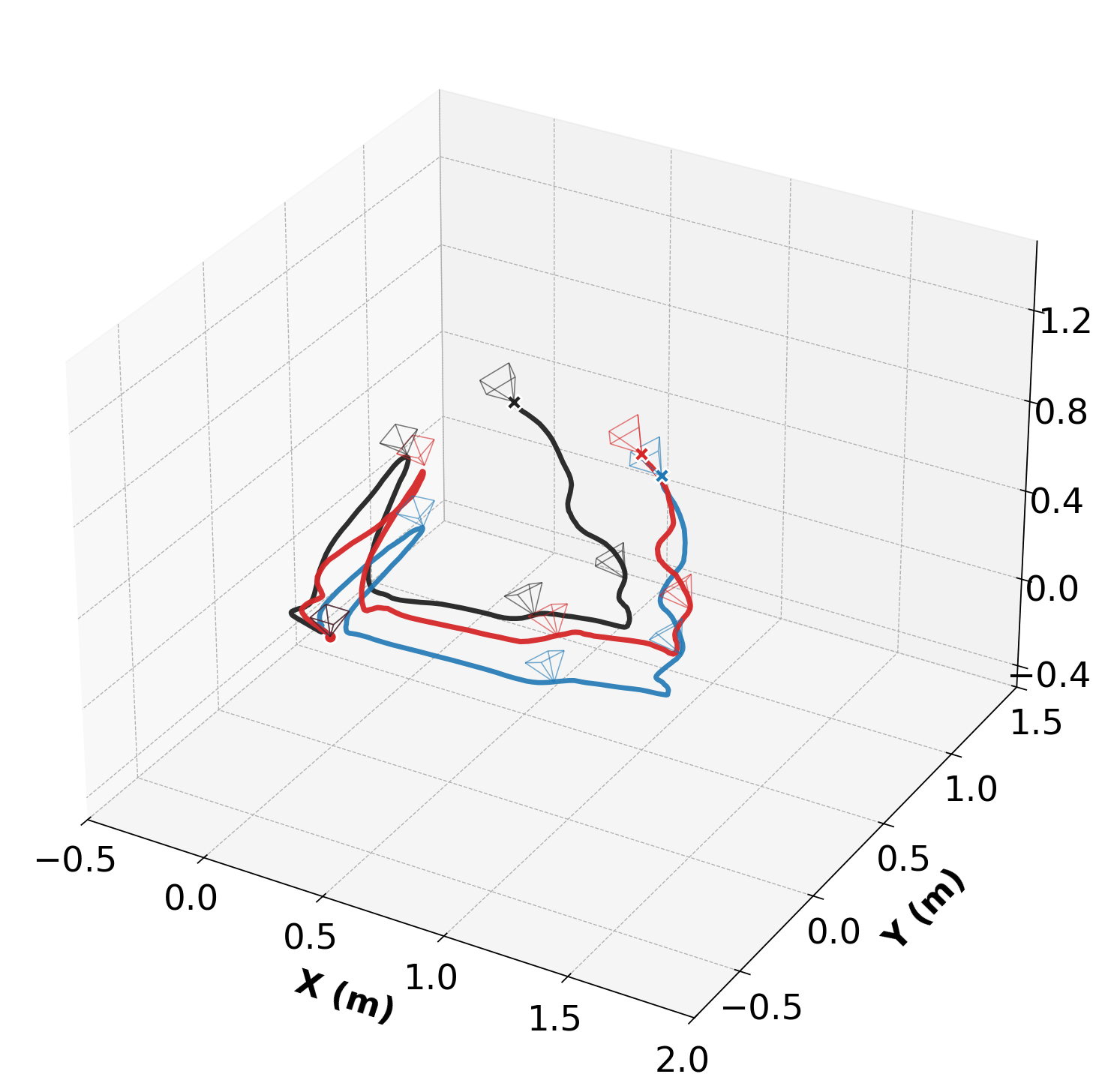}\\[-1pt]
  {\small \textsc{WinT3R} $\pm$\,ours}
\end{minipage}
\caption{\textbf{Additional streaming trajectories on ScanNet.} Each row is one $200$-frame sequence refined by \textsc{PosePlaner} in the same $100$-frame sliding-window scheme as Fig.~\ref{fig:streaming-curves} ($50$-frame overlap). Right three columns plot the estimated camera trajectory under \textsc{CUT3R}, \textsc{TTT3R}, and \textsc{WinT3R}; \textbf{black} is ground truth, \textcolor{blue}{\textbf{blue}} is the raw baseline, and \textcolor{red}{\textbf{red}} is after \textsc{PosePlaner} refinement.}
\label{fig:streaming-extra}
\vspace{-6pt}
\end{figure*}

\paragraph{Additional trajectories.} Fig.~\ref{fig:streaming-extra} extends the streaming results of Sec.~\ref{sec:exp-nview-new} with three additional sequences spanning low-texture and cluttered indoor settings; each is refined on top of \textsc{CUT3R}~\citep{cut3r}, \textsc{TTT3R}~\citep{ttt3r}, and \textsc{WinT3R}~\citep{wint3r}. Across all three sequences the refined trajectory tracks the ground truth substantially closer than the raw baseline, confirming that the refiner's drift correction is consistent across upstream estimators rather than tuned to any one of them.

\subsection{Training Objective Comparison}
\label{app:objective-comparison}
We compare flow matching with deterministic Direct-$x_0$ regression and DDPM pred-$x_0$, using the same architecture, training data, and 20-epoch schedule. Direct-$x_0$ predicts the clean pose residual in one forward pass; DDPM uses 100 denoising steps, and flow matching uses one Euler step. All variants use MASt3R correspondences and the same geometric translation solver. Both alternatives improve the LoFTR initialization, but flow matching achieves higher rotation and translation-direction AUC at all reported thresholds (Tab.~\ref{tab:objective-comparison}). Each objective is trained once.

\begin{table}[!htbp]
\centering
\small
\setlength{\tabcolsep}{5pt}
\caption{\textbf{Training objective comparison on ScanNet1500.} LoFTR initialization, $1028$ valid pairs. R@$\theta$/t@$\theta$: rotation/translation-direction AUC at $\theta^\circ$ (\%, $\uparrow$). Best results are \textbf{bold}.}
\label{tab:objective-comparison}
\begin{tabular}{l|ccc|ccc}
\toprule
Method & R@5 & R@10 & R@20 & t@5 & t@10 & t@20 \\
\midrule
LoFTR init & 11.5 & 25.0 & 44.0 & 2.1 & 5.6 & 13.0 \\
Direct-$x_0$ & 14.8 & 30.3 & 49.2 & 2.3 & 6.6 & 14.5 \\
DDPM pred-$x_0$ & 16.7 & 33.2 & 51.8 & 2.2 & 6.5 & 14.6 \\
Flow matching & \textbf{29.4} & \textbf{52.9} & \textbf{70.6} & \textbf{3.5} & \textbf{9.0} & \textbf{18.4} \\
\bottomrule
\end{tabular}
\end{table}

\begin{samepage}
\subsection{Alternative Plane Source}
\label{app:alternative-planes}
We replace the learned plane predictions with planes fitted to frozen MoGe~\citep{DBLP:conf/cvpr/WangXDX00Y25} point maps using patchwise RANSAC/PCA. The refiner, initial poses, correspondences, and remaining modules are unchanged, with no retraining. Table~\ref{tab:alternative-planes} shows that MoGe-derived planes retain substantial gains over the LoFTR initialization, although our learned planes achieve higher rotation AUC at all three thresholds.

\end{samepage}

\begin{table}[!htbp]
\centering
\small
\setlength{\tabcolsep}{7pt}
\caption{\textbf{Alternative plane source on ScanNet1500.} Rotation AUC at $5^\circ/10^\circ/20^\circ$ (\%, $\uparrow$) on the same $1028$ valid LoFTR pairs. Best results are \textbf{bold}.}
\label{tab:alternative-planes}
\begin{tabular}{l|ccc}
\toprule
Plane source & R@5 & R@10 & R@20 \\
\midrule
LoFTR init (no refinement) & 11.5 & 25.0 & 44.0 \\
MoGe + RANSAC & 27.1 & 50.4 & 69.2 \\
Learned planes (ours) & \textbf{29.4} & \textbf{52.9} & \textbf{70.6} \\
\bottomrule
\end{tabular}
\end{table}

\section{Social Impacts}
\label{app:social-impact}

\paragraph{Positive social impact.} \textsc{PosePlaner} corrects relative camera pose at the geometric level and extends from two-view training to streaming $n$-view inference without retraining. This makes accurate pose estimation more accessible to downstream applications such as AR/VR, robotics, and indoor scene understanding.

\paragraph{Potential negative impact.} As with any technology that improves camera pose estimation, our method could in principle be used in privacy-sensitive contexts. We recommend deploying it within a broader pipeline that applies the appropriate consent and access controls.
\stopcontents[appendix]

\end{document}